\documentclass{article}

\usepackage[preprint]{neurips_2026}

\usepackage[utf8]{inputenc} 
\usepackage[T1]{fontenc}    
\usepackage{hyperref}       
\usepackage{url}            
\usepackage{booktabs}       
\usepackage{amsfonts}       
\usepackage{nicefrac}       
\usepackage{microtype}      
\usepackage{xcolor}         
\usepackage{tikz}
\usepackage{amsmath}
\usepackage{amssymb}
\usepackage{graphicx}
\usepackage{comment}
\usepackage{tabularx}
\usepackage{multirow}
\usepackage{float}
\usepackage{enumitem}
\usetikzlibrary{arrows.meta, positioning, fit, backgrounds, shapes.geometric}
\usepackage{subcaption}

\definecolor{PriorBg}    {RGB}{245,243,255}   
\definecolor{PriorBdr}   {RGB}{131,109,221}   
\definecolor{ModelFill}  {RGB}{224,217,252}
\definecolor{ModelBdr}   {RGB}{83,74,183}
\definecolor{NodeFill}   {RGB}{193,237,220}   
\definecolor{NodeBdr}    {RGB}{15,110,86}
\definecolor{GrayFill}   {RGB}{215,213,205}
\definecolor{GrayBdr}    {RGB}{95,94,90}
\definecolor{SOTABg}     {RGB}{255,245,242}
\definecolor{SOTABdr}    {RGB}{216,90,48}
\definecolor{SOTABar}    {RGB}{240,153,123}
\definecolor{OursBg}     {RGB}{241,250,245}
\definecolor{OursBdr}    {RGB}{15,110,86}
\definecolor{OursBar}    {RGB}{29,158,117}
\definecolor{ArrowCol}   {RGB}{100,100,95}
\definecolor{BoxBg}      {RGB}{250,250,248}
\definecolor{DarkText}   {RGB}{40,40,38}

\tikzset{
  every node/.style={font=\small, text=DarkText},
  panel/.style={
    draw, rounded corners=6pt, inner sep=6pt,
    fill=BoxBg
  },
  model box/.style={
    draw=GrayBdr, rounded corners=2pt,
    inner sep=5pt, line width=0.6pt
  },
  gene causal/.style={
    circle, draw=NodeBdr, fill=NodeFill,
    inner sep=0pt, minimum size=18pt, line width=0.6pt,
    font=\footnotesize\bfseries, text=NodeBdr
  },
  gene noise/.style={
    circle, draw=GrayBdr, fill=GrayFill,
    inner sep=0pt, minimum size=16pt, line width=0.6pt,
    font=\footnotesize\bfseries, text=GrayBdr
  },
  tf node/.style={
    circle, rounded corners=3pt, draw=ModelBdr, fill=ModelFill,
    inner sep=0pt, minimum size=18pt, line width=0.6pt,
    font=\footnotesize\bfseries, text=ModelBdr
  },
  bar sota/.style={
    draw=SOTABdr, fill=SOTABar, rounded corners=2pt,
    minimum height=10pt, line width=0.4pt
  },
  bar ours/.style={
    draw=OursBdr, fill=OursBar, rounded corners=2pt,
    minimum height=10pt, line width=0.4pt
  },
  bar neutral/.style={
    draw=GrayBdr, fill=GrayFill, rounded corners=2pt,
    minimum height=10pt, line width=0.4pt
  },
  graph edge/.style={
    -Stealth, ArrowCol, line width=0.6pt, shorten >=2pt, shorten <=2pt
  },
  panel arrow/.style={
    -Stealth, ArrowCol, line width=0.8pt
  },
  annot box/.style={
    draw=ArrowCol!40, fill=white, rounded corners=4pt,
    inner sep=4pt, line width=0.4pt, align=center,
    font=\scriptsize, text=DarkText
  },
}

\title{Causal Integrated Gradients: \\ Scalable Feature Attribution Along Causal Paths}

\author{%
  Joran Michiels$^{1,2,3\ast\dagger}$ \quad Anis Ismail$^{1,2\ast}$ \quad Maarten De Vos$^{3,4}$ \quad Alejandro Sifrim$^{1,2}$ \\[4pt]
  $^1$ Laboratory of Multi-omics Integrative Bioinformatics, Department of Human Genetics, \\
  KU Leuven, Leuven, Belgium \\
  $^2$ KU Leuven Institute for Single Cell Omics (LISCO), KU Leuven, Leuven, Belgium \\
  $^3$ Department of Electrical Engineering, KU Leuven, Leuven, Belgium \\
  $^4$ Department of Development and Regeneration, KU Leuven, Leuven, Belgium \\[4pt]
  {\small $^\ast$ Equal contribution \quad $^\dagger$ Corresponding author: \texttt{joran.michiels@kuleuven.be}}
}

\newcommand{\motivationfigscale}{1}

\newcommand{\chainrulehsep}{1.7cm}
\newcommand{\chainrulevsep}{1.0cm}

\begin{document}

\maketitle

\begin{abstract}
Existing feature attribution methods that account for feature dependencies do not scale to high dimensions. We propose causal integrated gradients (CIG) and causal expected gradients (CEG), which extend integrated gradients to incorporate prior causal structure among input features while retaining its computational efficiency and, under conditions on the causal graph, its axiomatic guarantees. In many settings, some causal structure between features is already known—for example, that a transcription factor causally regulates a target gene's expression. Yet existing gradient-based methods cannot exploit this structure, instead crediting a feature's downstream effects over its true upstream cause, such as ranking the target gene above the transcription factor that regulates it. Our key idea is to embed this causal structure by routing the integration path along the causal graph. The method thereby decomposes each feature's attribution into a direct effect and effects mediated through its causal descendants. We evaluate CIG in three settings of increasing realism. First, on a synthetic benchmark with a known causal model, CIG's attributions best match the ground-truth causal effect from Pearl's do-calculus, matching the accuracy of causal Shapley values at a fraction of the computational cost. Second, on simulated single-cell gene expression data with a known gene regulatory network (Dyngen), CIG most accurately recovers the transcription factors driving the simulation. Third, on real interferon-$\beta$ stimulation data, where only an approximate causal graph is available, CIG better prioritizes the transcription factors driving the measured response than DeepSHAP and integrated gradients. Together, these results show that incorporating causal structure into gradient-based attributions recovers upstream drivers that existing methods miss, without sacrificing scalability.
Code is available at \url{https://github.com/sifrimlab/CausalIG}.
\end{abstract}

\section{Introduction}

As machine learning models grow increasingly complex, particularly deep neural networks trained on large-scale datasets, the need for interpretable explanations of their predictions becomes critical. Local post-hoc explanation methods have emerged as a popular approach for understanding the contribution of individual input features to model predictions. However, most existing methods assume that input features are independent, which is often unrealistic in practice.

Feature dependencies are common in real-world applications—in gene expression, for instance, genes exhibit complex inter-dependencies reflecting underlying biological processes. Ignoring them can produce misleading attributions that fail to uncover the true data-generating mechanism \citep{aas2021explaining, chen2020true}, with significant consequences for biological discovery.

Some approaches have been proposed to incorporate feature dependencies into attribution methods. Methods based on Shapley values, such as conditional SHAP \citep{aas2021explaining} and causal Shapley values \citep{heskes2020causal}, can account for dependencies but rely on sampling-based computations that become prohibitively expensive in high-dimensional settings. Conversely, gradient-based methods like integrated gradients (IG) \citep{sundararajan2017axiomatic} and expected gradients \citep{erion2021improving} offer computational efficiency but fundamentally assume feature independence.

\begin{figure}[t]
\centering
\begin{tikzpicture}[node distance=8pt, scale=\motivationfigscale, transform shape]


\node[model box] (nnBox) at (28pt, -30pt) {NN};

\node[above=12pt of nnBox, font=\small] (cellState) {cell state};
\draw[-Stealth, ArrowCol, line width=0.6pt] (nnBox) -- (cellState);

\node[gene noise] (G1)  at (0pt,  -90pt)  {G\textsubscript{1}};
\node[gene noise] (G2)  at (28pt, -90pt)  {G\textsubscript{2}};
\node[gene noise] (G3)  at (56pt, -90pt)  {G\textsubscript{3}};
\node[tf node]    (TF1) at (14pt, -120pt) {TF\textsubscript{1}};
\node[tf node]    (TF2) at (42pt, -120pt) {TF\textsubscript{2}};

\draw[graph edge] (TF1) -- (G1);
\draw[graph edge] (TF1) -- (G2);
\draw[graph edge] (TF2) -- (G2);
\draw[graph edge] (TF2) -- (G3);

\draw[graph edge] (G1)  -- (nnBox);
\draw[graph edge] (G2)  -- (nnBox);
\draw[graph edge] (G3)  -- (nnBox);
\draw[graph edge] (TF1) -- (nnBox);
\draw[graph edge] (TF2) -- (nnBox);

\node[inner sep=0pt, minimum size=0pt] (modelBotPh)   at (28pt, -47pt)  {};
\node[inner sep=0pt, minimum size=0pt] (modelLeftPh)  at (-9pt,  -2pt)  {};
\node[inner sep=0pt, minimum size=0pt] (modelRightPh) at (65pt,  -2pt)  {};
\node[inner sep=0pt, minimum size=0pt] (causalTopPh)  at (28pt, -73pt)  {};
\node[inner sep=0pt, minimum size=0pt] (causalBotPh)  at (28pt, -137pt) {};

\node[font=\small\bfseries, text=DarkText,
      above=8pt of cellState, anchor=south] (geneReg) {Gene regulation};

\begin{scope}[on background layer]
  \node[draw=ArrowCol!50, fill=white, rounded corners=10pt, inner sep=33pt,
        fit=(nnBox)(cellState)(modelBotPh)(modelLeftPh)(modelRightPh)(G1)(G2)(G3)(TF1)(TF2)(causalTopPh)(causalBotPh)] (leftOuterBg) {};
  \node[draw=GrayBdr, fill=GrayFill!40, rounded corners=6pt, inner sep=8pt,
        fit=(nnBox)(cellState)(modelBotPh)(modelLeftPh)(modelRightPh)] (modelBg) {};
  \node[draw=PriorBdr, fill=PriorBg, rounded corners=6pt, inner sep=8pt,
        fit=(G1)(G2)(G3)(TF1)(TF2)(causalTopPh)(causalBotPh)] (causalBg) {};
\end{scope}

\node[font=\small\itshape, text=GrayBdr, rotate=90,
      left=5pt of modelBg, xshift=25pt] {Trained model};
\node[font=\small\itshape, text=PriorBdr, rotate=90,
      left=5pt of causalBg, xshift=15pt] {Causal graph};

\node[below=1pt of causalBg.south, font=\scriptsize,
      text=ModelBdr, anchor=north] (legTF) {TF: transcription factor};
\node[below=-5pt of legTF, font=\scriptsize, text=GrayBdr] {G: target gene};


\node[font=\small\bfseries, text=SOTABdr,
      right=3cm of geneReg] (sotaTitle) {Existing integrated gradients};

\node[bar sota, minimum width=50pt,
      below=9pt of sotaTitle.south west, anchor=north west, xshift=-3pt,
        label={[font=\scriptsize,xshift=2pt]right:0.81}]
  (bs1) {\scriptsize G\textsubscript{1}~};
\node[bar sota, minimum width=42pt, below=5pt of bs1.south west, anchor=west,
      label={[font=\scriptsize,xshift=2pt]right:0.71}]
  (bs2) {\scriptsize G\textsubscript{2}~};
\node[bar sota, minimum width=34pt, below=5pt of bs2.south west, anchor=west,
      label={[font=\scriptsize,xshift=2pt]right:0.56}]
  (bs3) {\scriptsize G\textsubscript{3}~};
\node[bar neutral, minimum width=14pt, below=5pt of bs3.south west, anchor=west,
      label={[font=\scriptsize,xshift=2pt]right:0.16}]
  (bstf1) {\scriptsize TF\textsubscript{1}~};
\node[bar neutral, minimum width=9pt, below=5pt of bstf1.south west, anchor=west,
      label={[font=\scriptsize,xshift=2pt]right:0.09}]
  (bstf2) {\scriptsize TF\textsubscript{2}~};

\node[annot box, right=55pt of bs1.west, yshift=-40pt, text width=2.2cm] (annotSota)
  {Selects correlated \\ target genes\dots};

\begin{scope}[on background layer]
  \node[draw=SOTABdr!60, fill=SOTABg, rounded corners=10pt, inner sep=10pt,
        fit=(sotaTitle)(bstf2)(annotSota)] (sotaPanel) {};
\end{scope}


\node[font=\small\bfseries, text=OursBdr,
      below=3.15cm of sotaTitle] (oursTitle) {Our causal implementation};

\node[bar ours, minimum width=54pt,
      below=114pt of sotaTitle.south west, anchor=north west, xshift=-3pt,
      label={[font=\scriptsize,xshift=2pt]right:0.91~\checkmark}]
  (bo1) {\scriptsize TF\textsubscript{1}~};
\node[bar ours, minimum width=46pt, below=5pt of bo1.south west, anchor=west,
      label={[font=\scriptsize,xshift=2pt]right:0.78~\checkmark}]
  (bo2) {\scriptsize TF\textsubscript{2}~};
\node[bar neutral, minimum width=16pt, below=5pt of bo2.south west, anchor=west,
      label={[font=\scriptsize,xshift=2pt]right:0.13}]
  (bog1) {\scriptsize G\textsubscript{1}~};
\node[bar neutral, minimum width=11pt, below=5pt of bog1.south west, anchor=west,
      label={[font=\scriptsize,xshift=2pt]right:0.08}]
  (bog2) {\scriptsize G\textsubscript{2}~};
\node[bar neutral, minimum width=7pt, below=5pt of bog2.south west, anchor=west,
      label={[font=\scriptsize,xshift=2pt]right:0.05}]
  (bog3) {\scriptsize G\textsubscript{3}~};

\node[annot box, right=55pt of bo1.west, yshift=-40pt,
      text width=2.2cm] (annotOurs)
  {Recovers true \\  causal drivers!};
\begin{scope}[on background layer]
  \node[draw=OursBdr!60, fill=OursBg, rounded corners=10pt, inner sep=10pt,
        fit=(oursTitle)(bog3)(annotOurs)] (oursPanel) {};
\end{scope}


\draw[panel arrow] (modelBg.east) to[out=0, in=180] (sotaPanel.west);

\draw[panel arrow] (modelBg.east) to[out=-30, in=150] (oursPanel.west);

\draw[-Stealth, PriorBdr, line width=0.8pt]
  (causalBg.east) to[out=0, in=180] (oursPanel.west);

\end{tikzpicture}
\caption{Simulated example inspired by attribution of gene regulation. Existing graph-agnostic methods assign the highest attribution scores to the directly correlated target genes, missing the true upstream regulators; our graph-aware method correctly pushes the transcription factors to the top.}
\label{fig:motivation}
\end{figure}
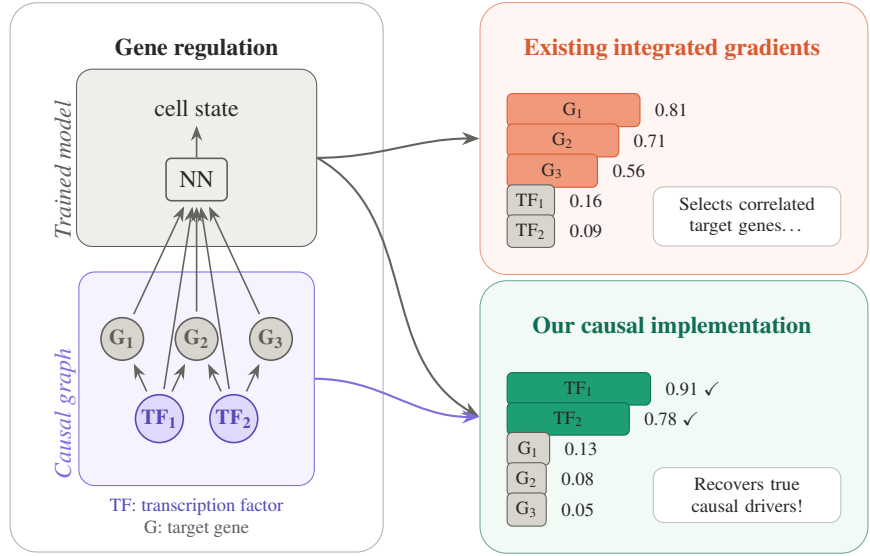

To address this gap, we propose causal integrated gradients (CIG) and causal expected gradients (CEG), principled extensions of integrated gradients that incorporate causal structure among input features. Like causal Shapley values \citep{heskes2020causal}, our methods require the structure of a causal graph over the inputs; unlike them, we need only a deterministic, off-the-shelf surrogate rather than a full conditional distribution to sample from. Conditional SHAP needs no causal graph but does need that conditional distribution, while standard integrated and expected gradients need neither. In return for this graph, our methods keep the efficiency of integrated gradients and, under conditions on the causal graph and surrogate, its axiomatic guarantees. They scale to high-dimensional settings, recovering upstream causal drivers that graph-agnostic methods miss, as in gene regulation (Figure~\ref{fig:motivation}).

The main contributions of this paper are as follows:
\begin{itemize}[topsep=2pt, itemsep=2pt, parsep=0pt]
    \item We propose causal integrated and expected gradients (CIG and CEG), extending integrated gradients to incorporate causal structure among input features. Both methods are applicable to any differentiable model for which causal dependencies among inputs are available.
    \item We evaluate the proposed methods on a synthetic benchmark with known causal effects and compare against the state-of-the-art (causal Shapley values \citep{heskes2020causal} and conditional SHAP \citep{aas2021explaining}), demonstrating improved causal attribution performance at substantially lower computational cost — reducing runtimes from hours to minutes.
    \item We demonstrate the real-world applicability of our method on two high-dimensional single-cell datasets: identifying transcription factors driving cell differentiation in simulated data, and prioritizing transcription factors that are most sensitive to interferon-$\beta$ stimulation. Across both tasks, our method outperforms existing approaches in recovering causal transcription factors, while scaling efficiently to high-dimensional RNA expression data.
\end{itemize}

\section{Background and related work}

A natural framework for gradient-based feature attribution is that of \emph{path methods}: let $F: \mathbb{R}^n \to \mathbb{R}$ be a differentiable model and $\gamma: [0,1] \to \mathbb{R}^n$ a differentiable path with $\gamma(0) = b$ (a baseline) and $\gamma(1) = \boldsymbol{x}$ (the input to explain). By the gradient theorem, $\int_{0}^{1} \nabla F(\gamma(\alpha)) \cdot \frac{d\gamma(\alpha)}{d\alpha} \, d\alpha = F(\boldsymbol{x}) - F(b)$; expanding the dot product gives the feature attribution $R_i$, interpreted as feature $\boldsymbol{x}_i$'s effect on the model output:
\begin{equation}
    \sum_{i=1}^{n} \int_{0}^{1} \frac{\partial F(\gamma(\alpha))}{\partial \gamma_i(\alpha)} \frac{d\gamma_i(\alpha)}{d\alpha} \, d\alpha = \sum_{i=1}^{n} R_i = F(\boldsymbol{x}) - F(b).
    \label{eq:attributions}
\end{equation}
Equation~\ref{eq:attributions} guarantees the \emph{completeness} property: attributions sum exactly to the difference in model output between the input and the baseline. Integrated gradients \citep{sundararajan2017axiomatic} and expected gradients \citep{erion2021improving} are both instances of this framework: integrated gradients uses a straight-line path $\gamma(\alpha) = b + \alpha(\boldsymbol{x} - b)$ from a fixed baseline, approximated in practice via a $K$-step Riemann sum; expected gradients instead averages over a distribution of baselines to remove the sensitivity to this single fixed choice, sampling the baseline and path parameter $\alpha \sim \mathcal{U}(0,1)$ jointly to form a Monte Carlo estimator. Both methods, however, implicitly assume that features are mutually independent along the path.

The incorporation of feature dependencies in local attribution methods has instead been most thoroughly explored in the context of Shapley values. Conditional SHAP \citep{aas2021explaining} and causal Shapley values \citep{heskes2020causal}
are prominent examples. Conditional SHAP accounts for statistical dependencies by replacing the marginal feature distribution with the conditional distribution when computing Shapley values, while causal Shapley values incorporate an explicit causal graph to distinguish direct effects from indirect effects mediated through other features. Another extension of Shapley values is the Owen value, which takes into account that features belong to groups \citep{owen_banzhaf_2019}. The partition explainer in the \texttt{shap} Python package \citep{lundberg2017unified} computes Owen values by reducing the number of coalitions on which the marginal contributions are computed, based on an initial partitioning of the features.
While incorporating dependencies is contested \citep{janzing2020feature, aas2021explaining}, it is preferable when the goal is to understand the true data-generating process \citep{chen2020true, michiels2024explaining} — a consideration especially consequential in biological discovery. To our knowledge, no prior work has extended this line of thinking to gradient-based attribution methods, which is the primary contribution of this paper.

Numerous modifications to the integrated gradients algorithm have been proposed, though none address feature dependencies. Most target either attribution noise \citep{kapishnikov2021guided}, the reliance on an arbitrary baseline \citep{erion2021improving}, or both \citep{yang2023local, zhuo2024ig}. Two works are particularly relevant to ours, as they also modify the integration path: \citet{kapishnikov2021guided} alters it to avoid regions of high partial derivatives, hypothesized to introduce noise, while \citet{zhuo2024ig} defines a path that follows the model's gradient. Our work differs fundamentally from both — rather than modifying the path for numerical reasons, we do so to encode causal structure among input features.

Feature attribution methods have been widely used to interpret deep learning models trained on RNA expression data and extract biologically meaningful insights \citep{yap_verifying_2021,janizek_pause_2023,tang_explainable_2023}. \citet{yap_verifying_2021}, for instance, applied expected gradients to models trained on the GTEx RNA-seq dataset (47 tissue types) to recover genes associated with tissue identity and function, while \citet{janizek_pause_2023} used explainability methods to assess whether unsupervised models capture meaningful gene relationships and pathways. However, most existing attribution methods ignore strong gene dependencies, potentially producing misleading attributions that fail to reflect the underlying biological structure, and their application to high-dimensional RNA expression data remains largely unexplored due to computational complexity.

\section{Causal integrated and expected gradients}

As shown in the background section, all path methods satisfy the same completeness property and differ only in their choice of path. This freedom in the path has been exploited before \citep{kapishnikov2021guided, zhuo2024ig}, but always for numerical reasons. We instead exploit it to encode causal structure by reparametrizing the path to reflect dependencies between input features.
Concretely, this requires three steps:
\begin{enumerate}[topsep=2pt, itemsep=2pt, parsep=0pt]
    \item Reparametrize the path function to incorporate dependencies between features, so that direct and indirect effects can be separated via the chain rule.
    \item Determine the gradients of the path function that appear in the resulting attribution formula.
    \item Model the feature dependencies needed to evaluate those gradients.
\end{enumerate}

\subsection{Incorporating dependencies in the path function}

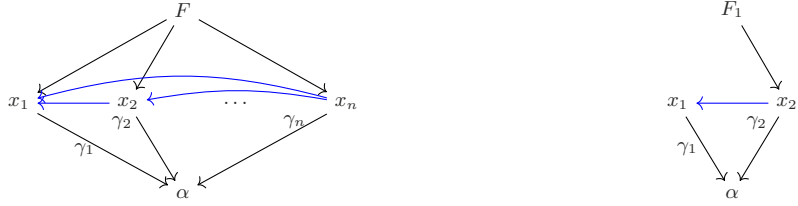
\begin{figure}[htbp]
    \centering
    \begin{subfigure}{0.58\textwidth}
        \centering
        \begin{tikzpicture}[node distance=\chainrulehsep, auto, scale=0.85, transform shape]
            \node (input1) {$x_1$};
            \node (input2) [right of=input1] {$x_2$};
            \node (dots)   [right of=input2] {$\cdots$};
            \node (inputn) [right of=dots]   {$x_n$};
            \node (alpha)  [below=\chainrulevsep of input2, xshift=0.5*\chainrulehsep] {$\alpha$};
            \node (output) [above=\chainrulevsep of input2, xshift=0.5*\chainrulehsep] {$F$};
            \draw[<-] (alpha) -- (input1) node[pos=0.5, left]  {$\gamma_1$};
            \draw[<-] (alpha) -- (input2) node[pos=0.9, left]  {$\gamma_2$};
            \draw[<-] (alpha) -- (inputn) node[pos=0.9, left] {$\gamma_n$};
            \draw[<-] (input1) -- (output);
            \draw[<-] (input2) -- (output);
            \draw[<-] (inputn) -- (output);
            \draw[<-, blue] (input1) -- (input2);
            \draw[<-, blue] (input2) to[bend left=10] (inputn);
            \draw[<-, blue] (input1) to[bend left=15] (inputn);
        \end{tikzpicture}
    \end{subfigure}
    \hfill
    \begin{subfigure}{0.38\textwidth}
        \centering
        \begin{tikzpicture}[node distance=\chainrulehsep, auto, scale=0.85, transform shape]
            \node (input1) {$x_1$};
            \node (input2) [right of=input1] {$x_2$};
            \node (alpha) [below=\chainrulevsep of input1, xshift=0.5*\chainrulehsep] {$\alpha$};
            \node (output) [above=\chainrulevsep of input1, xshift=0.5*\chainrulehsep] {$F_1$};
            \draw[<-] (alpha) -- (input1) node[pos=0.5, left] {$\gamma_1$};
            \draw[<-] (alpha) -- (input2) node[pos=0.9, left] {$\gamma_2$};
            \draw[<-] (input2) -- (output);
            \draw[<-, blue] (input1) -- (input2);
        \end{tikzpicture}
    \end{subfigure}
    \caption{Chain rule diagrams illustrating the reparametrization of path functions to incorporate causal dependencies. \emph{Left:} general $n$-feature case. \emph{Right:} two-feature example. Black edges are standard path method dependencies; blue edges are the causal dependencies introduced by our reparametrization.}
    \label{fig:chain_rule_diagrams}
\end{figure}

In standard path methods, each path variable $\gamma_i$ depends only on $\alpha$, meaning attributions reflect only the direct effect of each feature on the model output. To incorporate indirect effects, we allow $\gamma_i$ to depend on other path variables $\gamma_j$. This modification is best understood through the chain rule diagram, which for standard integrated gradients with $n$ features takes the form of the \textbf{black} edges in the left panel of Figure~\ref{fig:chain_rule_diagrams}, with $x_i = \gamma_i(\alpha)$.

Note that since the chain rule diagram is a directed acyclic graph (DAG), any dependency we add between input features must be directed — determining which feature causes the other — and must not introduce a cycle. We continue with the most general case of a fully connected DAG, with features ordered topologically: if $x_u$ depends on $x_v$, then $u > v$. Such dependencies are represented as \textcolor{blue}{blue} edges in the left panel of Figure~\ref{fig:chain_rule_diagrams}, and under this ordering the individual path functions become $x_i = \gamma_i(x_1, \dots, x_{i-1},\alpha)$: each feature's path depends on all features that causally precede it, plus $\alpha$.

With the modified path variables, computing the attributions in Equation~\ref{eq:attributions} requires an additional application of the chain rule. We illustrate this with a simple example: a function $F_1$ that depends only on $x_2$, where $x_2$ itself depends on $x_1$, as shown in the right panel of Figure~\ref{fig:chain_rule_diagrams}. Expanding the left-hand side of Equation~\ref{eq:attributions} for this example gives:
\begin{equation}
\begin{aligned}
    \int_{0}^{1} \frac{\partial F_1}{\partial \gamma_2} \frac{d\gamma_2}{d\alpha} \, d\alpha
    &= \int_{0}^{1} \frac{\partial F_1}{\partial \gamma_2} \frac{\partial \gamma_2}{\partial \alpha} \, d\alpha
    + \int_{0}^{1} \frac{\partial F_1}{\partial \gamma_2} \frac{\partial \gamma_2}{\partial \gamma_1} \frac{d\gamma_1}{d\alpha} \, d\alpha,
\end{aligned}
\label{eq:attributions_f1}
\end{equation}
where the first term is the direct attribution of $x_2$ and the second is the indirect attribution of $x_1$ through $x_2$. Generalizing to a fully connected DAG with $n$ features, the attribution of feature $i$ is:
\begin{equation}
    R_i = \int_{0}^{1} \frac{\partial F}{\partial \gamma_i} \frac{\partial \gamma_i}{\partial \alpha} \, d\alpha
    + \sum_{j > i} \int_{0}^{1} \frac{\partial F}{\partial \gamma_j} \frac{\partial \gamma_j}{\partial \gamma_i} \frac{d\gamma_i}{d\alpha} \, d\alpha = R_{i, \mathrm{direct}} + R_{i, \mathrm{indirect}},
    \label{eq:attributions_fully_connected_dag}
\end{equation}
where the first term captures the direct effect of $x_i$ and the second captures its indirect effect through all features that depend on it. This formulation applies to any DAG that represents model and input dependencies: missing edges simply correspond to zero terms ($\frac{\partial \gamma_j}{\partial \gamma_i} = 0$ if $x_j$ does not depend on $x_i$). Additionally, any DAG can be reordered to match the topological ordering (see Appendix~\ref{appendix:multiple_hop_implementation} for why this choice does not affect attributions). One subtlety is that the indirect terms use the full derivative $\frac{d\gamma_i}{d\alpha}$, meaning part of the indirect effect of $x_i$ may itself pass through features that $x_i$ depends on. In other words, this formulation attributes only indirect effects that are one hop long (also referred to as first-order). We provide an extension to multi-hop indirect effects in Appendix~\ref{appendix:multiple_hop_implementation}. Finally, to be able to compute the above integrals, we need to find a practical way to compute the gradients of the path function, which we will discuss in the next section.

\subsection{Determining the gradients of the path function}

To evaluate Equation~\ref{eq:attributions_fully_connected_dag} via a Riemann sum over $K$ steps, we require the following extra gradients at each point $\alpha^{(p)} = p/K$, $p = 1, \dots, K$, along the path:
\begin{equation}
\frac{\partial \gamma_i}{\partial \alpha}(\alpha^{(p)}), \quad \frac{\partial \gamma_j}{\partial \gamma_i}(\gamma_i^{(p)}),
\end{equation}
with $\gamma_i^{(p)} = b_i + \alpha^{(p)}(x_i - b_i)$, i.e.\ a straight-line path as in regular integrated gradients. To compute these, we need an explicit formulation of the path functions $\gamma_i$.

To this end, we interpret the model and the feature dependencies as a structural causal model (SCM) (as previously in \citealp{janzing2020feature,heskes2020causal}). If the arrow directions in the chain rule diagram of the left panel of Figure~\ref{fig:chain_rule_diagrams} are switched, the graph effectively represents a causal graph, and the path functions can represent the structural mechanism between parent and child nodes of the graph.
We will interpret $\alpha$ as residual information, which explains the part of the variation in the children that is not explained by their parents. As is common in causal inference, we assume this residual enters additively. Concretely, for a fully connected DAG with features ordered topologically, we model each feature as:
\begin{equation}
    x_i = \gamma_i(x_1, \dots, x_{i-1},\alpha) = c_i(x_1, \dots, x_{i-1}) + h_i(\alpha), \quad i = 1, \dots, n,
    \label{eq:scm_decomposition}
\end{equation}
where $c_i$ is the causal mechanism mapping the parents of $x_i$ to its expected value, and $h_i(\alpha)$ is an additive residual term. We approximate the causal mechanism $c_i$ with a surrogate model $s_i$ trained to predict $x_i$ from its causal parents, so that the residual $h_i^{(p)} \approx x_i^{(p)} - s_i(x_1^{(p)}, \dots, x_{i-1}^{(p)})$ captures the variation in $x_i$ not directly explained by its parents. The required gradients are then:
\begin{equation}
    \frac{\partial \gamma_j}{\partial \gamma_i}(\gamma_i^{(p)}) = \frac{\partial s_j}{\partial x_i}(x_i^{(p)}), \qquad
    \frac{\partial \gamma_i}{\partial \alpha}(\alpha^{(p)}) \approx \frac{h_i^{(p)} - h_i^{(p-1)}}{\Delta \alpha},
    \label{eq:finite_difference}
\end{equation}
where the last one is approximated via a backward finite difference, consistent with related work \citep{zhuo2024ig}. The quality of these gradients depends on the choice of surrogate model $s_i$, which we discuss in the next section.

\subsection{Modeling feature dependencies through a surrogate model} \label{sec:surrogate}
Any differentiable surrogate model can be employed to model the feature dependencies under a known causal structure. Here, we introduce a surrogate model that is flexible enough to model complex dependencies while being efficient and scalable to high-dimensional datasets. We propose a sparse masked surrogate $s$ that parameterizes parent-to-child relationships as a sum of edge-wise contributions, operating on a directed graph defined by an adjacency matrix $M \in \mathbb{R}^{n \times n}$, where $M_{ij}=1$ indicates that feature $i$ is a parent of feature $j$.

Each parent feature $x_i$ is first transformed using a per-feature basis expansion $\phi(x_i) \in \mathbb{R}^{Q}$ of size $Q$, to give the surrogate model more flexibility in learning nonlinear effects.
The basis consists of three components applied to standardized inputs: 1) polynomial terms, 2) truncated power functions with learnable knots, and 3) Fourier features (see Appendix~\ref{appendix:basis_expansion} for a detailed explanation of the basis expansion).
For every active edge $(i \rightarrow j)$ such that $M_{ij}=1$, the prediction for child node $j$ is obtained by summing contributions from all parent nodes:
\begin{equation}
s_j = \hat{x}_j = \sum_{i: M_{ij}=1} \langle w_{ij}, \phi(x_i) \rangle+\mu_j,
\end{equation}
where $\mu_j$ and $w_{ij} \in \mathbb{R}^{Q}$ are learnable per-node intercept and edge-specific coefficients, respectively.

Rather than materializing all possible $n^2$ parent-child interactions, the model stores only active graph edges. This sparse formulation reduces both memory usage and computation, making the model scalable to large sparse graphs. The surrogate is trained to reconstruct the set of predictable child nodes from their parent nodes by minimizing the mean squared reconstruction error, with early stopping on validation loss (full hyperparameters in Tables~\ref{tab:surrogate_training_details_dyngen} and \ref{tab:surrogate_training_details_ifnb}). The additive edge structure also makes the surrogate Jacobian needed for attribution analytical and sparse: for a child node $j$ and parent feature $x_i$, the derivative is
\begin{equation}
\frac{\partial s_j}{\partial x_i}=
\sum_{q=1}^{Q}
\frac{\partial \phi_q(x_i)}{\partial x_i}w_{ij,q}M_{ij},
\end{equation}
which is nonzero only for active edges.

\subsection{Extension to causal expected gradients} \label{sec:expected_gradients}
The above method can be extended to multiple baselines in the same way as expected gradients extends integrated gradients: sampling $\alpha$ and the baseline $b$ jointly still gives an unbiased single-sample estimate for either method. Standard expected gradients performs well with just one path point per baseline, since the straight-line path has a constant, closed-form derivative $d\gamma_i/d\alpha = x_i - b_i$, so a single sampled $\alpha$ already gives this derivative exactly. Our method's path derivative $\partial\gamma_i/\partial\alpha$ is instead estimated via a backward finite difference (Equation~\ref{eq:finite_difference}), since $h_i(\alpha)$ is generally nonlinear; our method therefore benefits from allocating more of a fixed evaluation budget to additional path points per baseline rather than to additional baselines, since both the finite-difference and Riemann-sum errors (Appendix~\ref{appendix:axiomatic_properties}) compound when paths are under-sampled.

\subsection{Axiomatic properties}

Since our method belongs to the family of path methods \citep{sundararajan2017axiomatic}, it inherits completeness and sensitivity(a) (for the direct term) unconditionally, and implementation invariance, symmetry, and linearity under precise preconditions on the causal graph and surrogate that we formalize in Appendix~\ref{appendix:axiomatic_properties}. The one exception is the dummy axiom: a feature absent from $F$ but causally upstream of a feature that does appear can receive non-zero indirect attribution, as shown in the example in the right of Figure~\ref{fig:chain_rule_diagrams}.

\section{Experiments} \label{sec:experiments}

We evaluate our method in three settings of increasing realism: a synthetic benchmark with known ground-truth causal effects, and two biological datasets where understanding causal relationships is crucial for scientific discovery. Experimental setup and costs are reported in Appendix~\ref{appendix:experimental_setup}.

\subsection{Comparison with causal attributions} \label{sec:causal_attributions_comparison}

Conditional SHAP \citep{aas2021explaining} and causal Shapley values \citep{heskes2020causal} are the most established approaches for dependency-aware feature attribution. To evaluate how well each method captures ground-truth causal effects, we compare their attributions against a reference causal effect derived from Pearl's do-calculus \citep{pearl2009causality}. Among the causal effects this framework gives rise to, the Conditional Average Treatment Effect (CATE) is the most appropriate for local explanations, as it measures the effect of intervening on a single feature while conditioning on the root nodes $\mathcal{R}$:
\begin{equation}
    \text{CATE}_i = \mathbb{E}[F \mid \text{do}(X_i = x_i), X_{\mathcal{R} \setminus \{i\}} = x_{\mathcal{R} \setminus \{i\}}] - \mathbb{E}[F \mid \text{do}(X_i = b_i), X_{\mathcal{R} \setminus \{i\}} = x_{\mathcal{R} \setminus \{i\}}].
\end{equation}
While several papers have analyzed the theoretical relationship between local attributions and Pearl's do-calculus \citep{heskes2020causal, janzing2020feature, ng2025causal}, to the best of our knowledge no prior work has empirically compared attribution methods against the CATE directly. Doing so gives us an exact ground truth for each feature's local causal effect, available only in this fully controlled setting. Since the CATE is an effect rather than an attribution (e.g.\ it need not satisfy completeness), we compare feature rankings via Spearman correlation rather than raw values; the two experiments that follow instead use biological data, where ground truth is global (Dyngen) or approximate (IFN-$\beta$).

\begin{figure}
    \centering
    \begin{subfigure}[c]{0.48\textwidth}
        \centering
        \includegraphics[width=\textwidth]{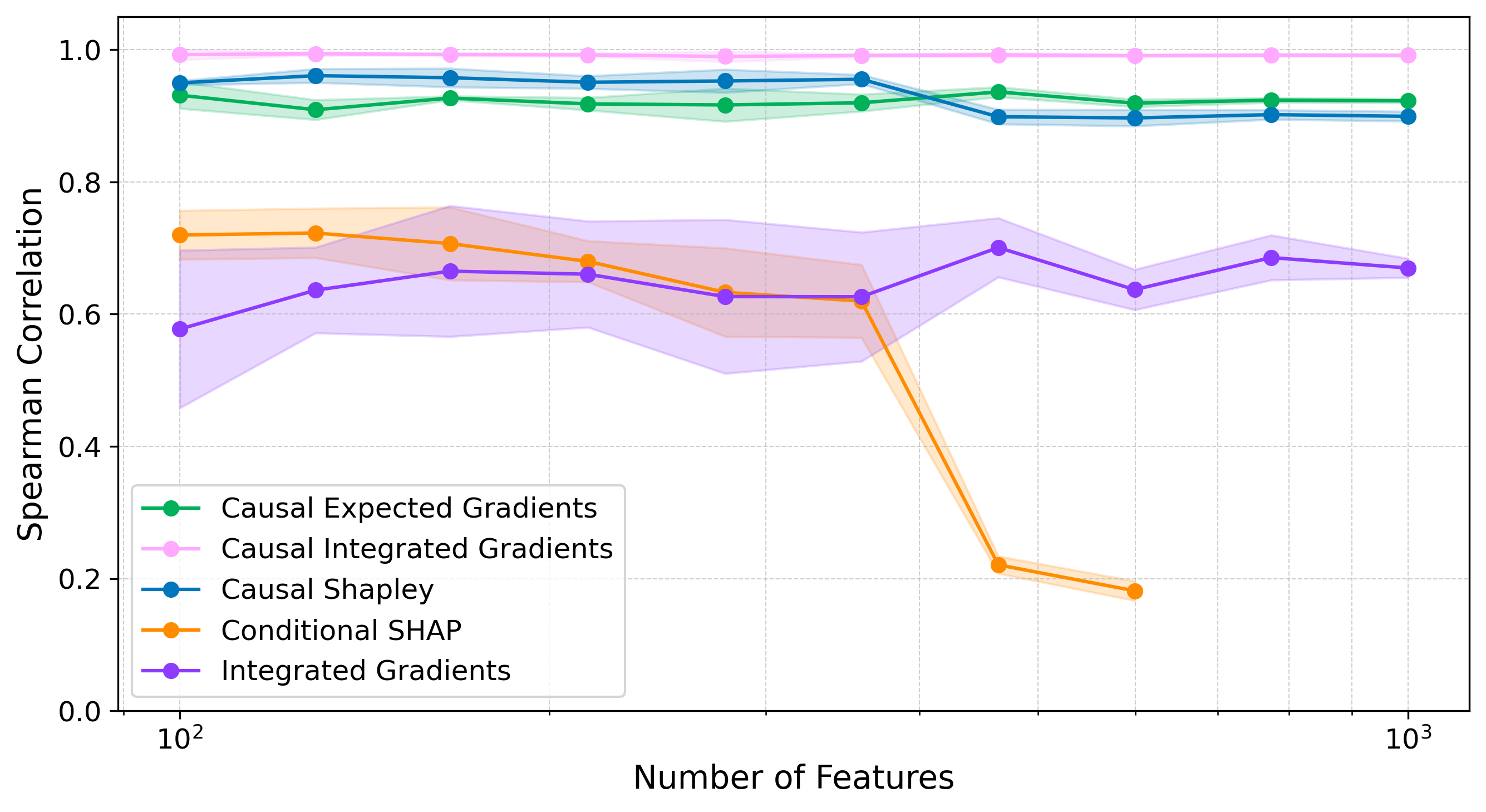}
    \end{subfigure}
    \hfill
    \begin{subfigure}[c]{0.48\textwidth}
        \centering
        \includegraphics[width=\textwidth]{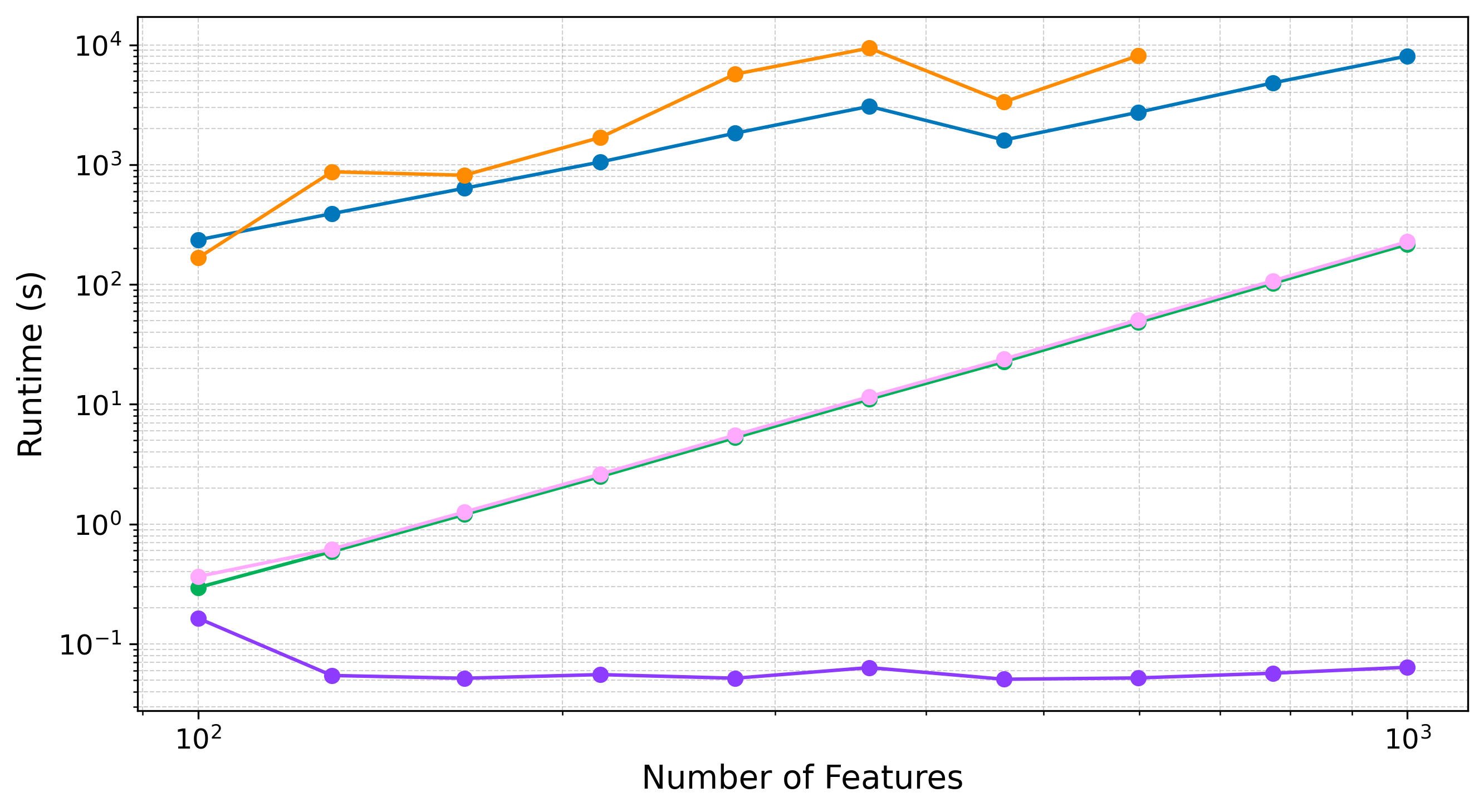}
    \end{subfigure}
    \caption{\emph{Left:} Spearman correlation with the CATE on root nodes only. \emph{Right:} wall-clock runtime (log--log scale) to explain all 100 samples; the dip for the Shapley-based methods reflects their reduced sampling budget at larger feature counts. Both as a function of the number of features. Shaded areas denote the standard deviation over 5 runs. Results on all nodes are shown in Figure~\ref{fig:causal_scaling_spearman_all_nodes}.}
    \label{fig:causal_scaling}
\end{figure}

To enable computation of both the CATE and the local attributions, we construct a synthetic benchmark with a known SCM: the target is a random linear combination of a subset of features, and a Lasso regression model $F$ is trained on the generated data, using the SCM's structural mechanisms directly for causal Shapley values and our own methods to isolate attribution quality from implementation noise (full details in Appendix~\ref{appendix:causal_attributions_extra}). Across an increasing number of features, we report the Spearman correlation between each method's attributions and the CATE, averaged over root nodes (left of Figure~\ref{fig:causal_scaling}) and over all nodes, alongside runtime (right panel of Figure~\ref{fig:causal_scaling}).
\paragraph{Results and discussion.} On root nodes (left panel of Figure~\ref{fig:causal_scaling}), CIG achieves near-perfect CATE correlation, with causal Shapley values trailing closely behind; conditional SHAP and integrated gradients fall considerably short, and conditional SHAP was excluded beyond moderate feature counts due to prohibitive runtime and consistently poor quality. Only causal methods shift attribution toward root nodes (see Appendix~\ref{appendix:root_node_intuition} for an intuitive example), which is needed to identify upstream drivers. Our methods achieve this at substantially lower computational cost than causal Shapley values (right panel of Figure~\ref{fig:causal_scaling}), whose exact computation scales exponentially in the number of features versus our worst-case quadratic scaling (full derivation, including a sparse-graph case, in Appendix~\ref{appendix:complexity}); standard integrated gradients remains the fastest (a single batched pass) but captures none of the causal structure. CEG underperforms CIG despite using the same computational budget: given a fixed number of gradient evaluations, CEG trades path steps for additional baselines, and as discussed in Section~\ref{sec:expected_gradients}, path steps are critical for accurately approximating the causal path integral, so reducing them degrades attribution quality. Finally, on all nodes (Figure~\ref{fig:causal_scaling_spearman_all_nodes}), CIG still performs better than all other methods.
\subsection{Causal gene recovery in a simulated biological dataset} \label{sec:dyngen_experiment}
We evaluate our method on a regression task using simulated single-cell gene expression data, predicting each cell's simulated differentiation time from its trajectory of activated and repressed transcription factors. The dataset is generated with Dyngen \citep{cannoodt_spearheading_2021} under a known gene regulatory network (GRN) with a backbone of transcription factors (TFs) that causally regulate downstream genes. Unlike the per-sample CATEs of the synthetic benchmark above, ground truth here is exact but global: the backbone TFs are the known drivers of the simulation, letting us test recovery at realistic single-cell scale and complexity. After filtering and normalization, a fully connected neural network is trained to predict simulated time.

The GRN defines both the feature ordering and surrogate structure for our method. We evaluate several CIG/CEG variants (linear or expanded-basis surrogate, first- or second-order terms), against attribution baselines from the \texttt{captum} and \texttt{shap} packages \citep{kokhlikyan2020captumunifiedgenericmodel,lundberg2017unified}, over five random seeds. Conditional SHAP and causal Shapley values were excluded due to their computational infeasibility at this scale. Performance is evaluated by backbone-TF recovery: area under the recall curve (AUC) (treating backbone TFs as positives), and discounted cumulative gain (DCG) over the top-ranked genes (relevance scores assigned according to each gene's role in the GRN). We also evaluate our method's ability to prioritize the causal regulators for each transition state. Dataset, model, surrogate, and evaluation details are given in Appendix~\ref{appendix:gene_recovery_extra_dyngen}.

\begin{figure}[h]
    \centering
    \includegraphics[width=0.9\linewidth]{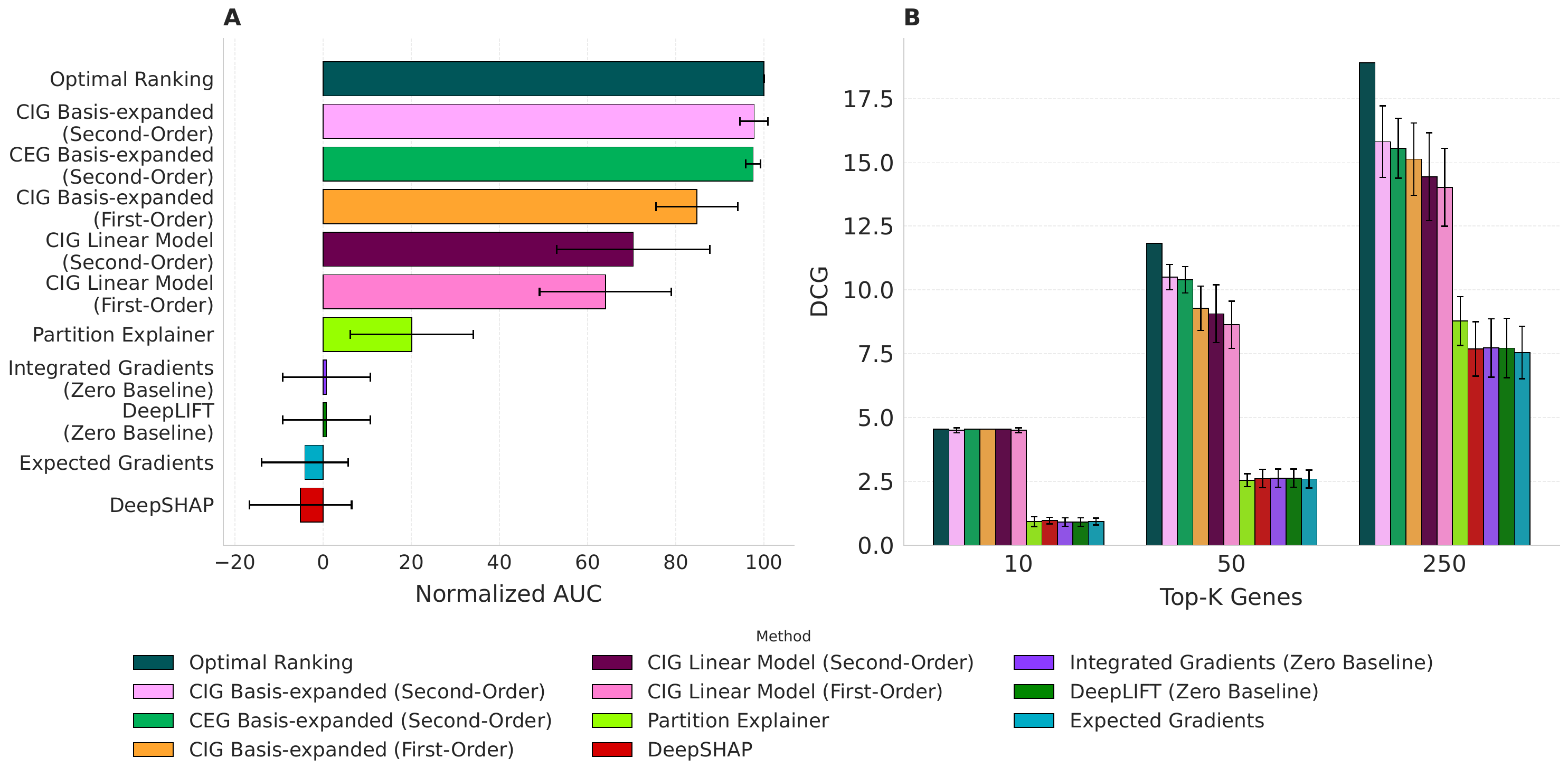}
    \caption{Evaluation of feature attribution methods on Dyngen. (A) Area under the recall curve for backbone transcription factors, normalized so that a random ranking scores 0 and the optimal ranking 100. (B) DCG over the top-$K$ ranked genes, with the DCG of the optimal ranking shown for reference; relevance scores follow Table~\ref{tab:gene_relevance_scores}. Error bars denote the standard deviation over 5 seeds.}
    \label{fig:simulation_main_results}
\end{figure}

\paragraph{Results and discussion.}
Our methods achieve the strongest recovery of backbone TFs (Figure~\ref{fig:simulation_main_results}A). The expanded-basis with second-order terms is the best overall, and SHAP's partition explainer is the best baseline. Both second-order terms and the expanded--basis improve performance over the linear and first-order variants. Similar trends are observed for the DCG scores (Figure~\ref{fig:simulation_main_results}B): Our method variants are near-optimal at a threshold of 10 genes, and the expanded-basis CIG remains the best overall at 50 and 250. Similar performance is shown in recovering causal regulators per transition state (Figure~\ref{fig:locality_test_CIG}), with CIG beating DeepSHAP and integrated gradients in every state.
\subsection{Perturbed gene discovery in the \texorpdfstring{IFN-$\beta$}{IFN-beta} stimulation dataset} \label{sec:ifn_beta_experiment}
We next evaluate our method on real single-cell RNA-sequencing data from \citet{kang_multiplexed_2018}, in which peripheral blood mononuclear cells were stimulated with interferon-$\beta$ (IFN-$\beta$) or left as controls. A fully connected neural network is trained to predict whether a cell was exposed to IFN-$\beta$ stimulation. Unlike the previous two experiments, no true causal graph is available here, so ground truth is only approximate: we rely on curated prior knowledge (OmniPath) and state-of-the-art regulatory network inference methods (SCENIC) to test whether our method still identifies the transcription factors driving the interferon response. We compare our method against DeepSHAP \citep{lundberg2017unified} and integrated gradients \citep{sundararajan2017axiomatic} (zero baseline). We derive a directed acyclic graph from OmniPath \citep{turei_integrated_2021}, used both for feature ordering and to train the surrogate. We compute attributions for all genes in held-out test cells across five seeds, summarized as the mean absolute attribution per gene, and evaluate rank by hop distance from the roots of both the OmniPath prior and a SCENIC-inferred regulatory network~\citep{aibar_scenic_2017}. SCENIC leverages motif- and TF-curated references to derive gene regulatory networks from the data while OmniPath is not data-specific. Dataset, model, graph construction, surrogate, and evaluation details are given in Appendix~\ref{appendix:gene_recovery_extra_ifnb}.

Next, we test whether the methods recover stimulation-responsive regulators: by
ranking genes per cell type by their attribution difference between stimulated and control cells and testing whether SCENIC-responsive TFs rank above the rest, using rank- and top-$k$-enrichment tests (one-sided Mann--Whitney $U$ test and hypergeometric test, respectively).
As an orthogonal readout, we evaluate recovery of per-cell-type pseudobulk differentially expressed genes (DEGs). For the three cell types most responsive to IFN-$\beta$ stimulation \citep{kang_multiplexed_2018} (CD14$^+$/FCGR3A$^+$ monocytes, dendritic cells), we also run over-representation analysis (ORA) \citep{fang_gseapy_2023} on each method's unique top-100 genes against MSigDB Hallmark gene sets \citep{liberzon_molecular_2015}.

\begin{figure}[h]
    \centering
    \includegraphics[width=\linewidth]{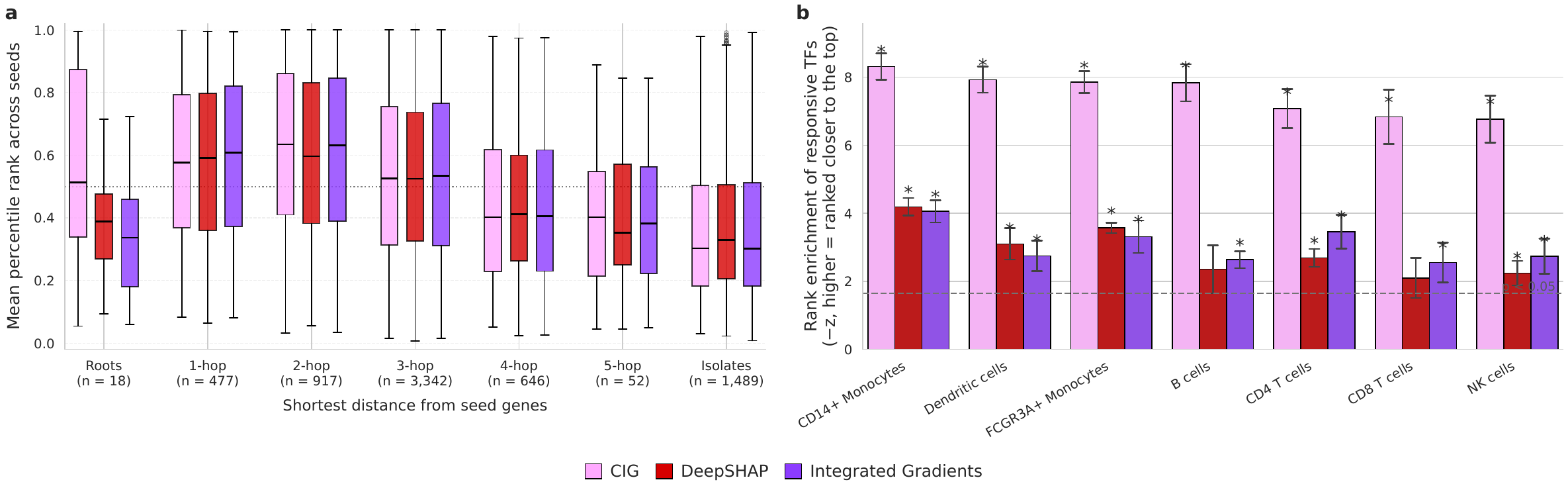}
    \caption{Recovery of stimulation-responsive regulators on the IFN-$\beta$ dataset.
    (A) Distribution of attribution ranks for genes grouped by hop distance from
    the roots of the SCENIC validation graph, median indicated by a horizontal
    line within each box. (B) $z$-score of the Mann--Whitney $U$ test statistic
    per method and cell type, averaged across 5 seeds; higher values indicate
    responsive TFs are ranked higher than other genes; error bars denote standard
    deviation across seeds; $\ast$: FDR $<0.05$ for every seed.}
    \label{fig:ifnb_combined}
\end{figure}

\paragraph{Results and discussion.}
The distribution of attribution ranks across hop-distance groups shows the CIG advantage is concentrated at the top of the regulatory hierarchy: on the OmniPath prior, CIG ranks root genes higher than DeepSHAP and integrated gradients, with a smaller advantage one hop downstream (Figure~\ref{fig:ifnb_hop_distance_omnipath}); the effect also holds on the SCENIC graph (Figure~\ref{fig:ifnb_combined}A), where DeepSHAP and integrated gradients rank roots below genes one to three hops downstream. Rankings are reproducible across seeds, with CIG the most stable of the three at SCENIC roots (Tables~\ref{tab:sd_median_omnipath} and \ref{tab:sd_median_scenic}).

Across cell types, CIG places stimulation-responsive TFs higher in its gene ranking than the remaining methods, with a rank-enrichment score two- to three-fold larger (Figure~\ref{fig:ifnb_combined}B). The top-$k$ analysis separates the methods further (Appendix~\ref{appendix:topk_enrichment}): CIG alone is enriched at every cutoff across cell types, while DeepSHAP and integrated gradients show no enrichment among the top 5--10 genes and recover only a fraction of CIG's enrichment from $k{=}25$ onward. The recovery of DEGs (Appendix~\ref{appendix:bulk_deg_recovery}, Table~\ref{tab:degs_cutoff10}) reveals that DeepSHAP and integrated gradients rank DEGs higher than CIG. CIG also shows the weakest correlation with the DE statistic across all cell types (Table~\ref{tab:degs_spearman}), likely because it prioritizes upstream regulators, which are rarely strong DEGs themselves. ORA results (Appendix~\ref{appendix:go_enrichment}) confirm that CIG uniquely picks out pathways with transcription factors that regulate the interferon-$\beta$ response, with significantly enriched pathways such as TNF-$\alpha$/NF-$\kappa$B, IL-6/JAK/STAT3, and interferon response. DeepSHAP, by contrast, enriches only for the interferon response, with the remaining pathways enriched by integrated gradients not reaching significance.

\section{Limitations and future work} \label{sec:limitations}
Our method requires prior knowledge of the directed causal structure among features, which for biological data can often be sourced from curated knowledge bases such as OmniPath \citep{turei_integrated_2021}. Since such a graph may be imperfect in practice, we test how much this assumption can be violated before performance suffers by removing, adding, and reversing edges: CIG degrades under genuine information loss (removal or reversal) but tolerates spurious additions almost entirely, so the graph need not be exactly correct to retain most of the benefit (Appendix~\ref{appendix:graph_corruption}). A separate concern is whether CIG's advantage merely reflects graph access, not correct use of it. We rule this out: CIG's attributions are instance-specific, tracking each cell's actual transition rather than the static graph alone (Appendix~\ref{appendix:naive_baseline_locality}). Like causal Shapley values \citep{heskes2020causal} and conditional SHAP \citep{aas2021explaining}, CIG is true to the data rather than to the model: it recovers causal drivers under the supplied structure and is not meant for debugging a model's idiosyncratic behavior. Extending the method to undirected or partially directed graphs is a natural direction for future work.

Our surrogate models each feature's mechanism as additive across parents (Section~\ref{sec:surrogate}) and so cannot capture non-additive parent interactions. This is a limitation of our chosen surrogate, not of the method, since any differentiable surrogate can be substituted. The one case a different surrogate cannot fix is an unmodeled confounder, whose effect is folded into the additive residual and attributed to $x_i$'s direct effect. In our experiments this biases attribution magnitudes for affected nodes (Appendix~\ref{appendix:residual_dependence}).

Our methods scale quadratically in the number of features in the worst case, slower than integrated gradients but substantially faster than Shapley-based alternatives; our synthetic and real-data experiments (100--1000 features; 2,320--6,941 genes) already exceed the scale at which existing dependency-aware baselines have been validated (\citet{heskes2020causal}: $M{=}7$ features; \citet{aas2021explaining}: 3--10 features, largest real-data example at 28). Finally, the current formulation concentrates attribution on root nodes, potentially underweighting relevant intermediate nodes; future work could explore distributing indirect effects more evenly across the causal path.

\section{Conclusion}
We introduced causal integrated gradients and causal expected gradients, which extend integrated gradients to incorporate known causal structure among input features. They apply to any differentiable model and, under conditions on the causal graph and surrogate, retain the axiomatic guarantees of integrated gradients. On a synthetic benchmark, CIG best matches ground-truth causal effects at a fraction of the cost of causal Shapley values; on simulated and real single-cell data, it best recovers the transcription factors driving the response while scaling to thousands of genes.

\subsection*{AI use statement}
We used generative AI tools (Claude, ChatGPT and GitHub Copilot) to assist with drafting and editing the paper text, writing proofs, formatting references, searching the literature, and general-purpose coding for data processing, plotting and experiment scripts; they were not used to implement the core CIG/CEG method. All mathematical claims and methodological decisions are the authors' own. The authors reviewed all AI-suggested text, verified all AI-assisted proofs, and tested all AI-assisted code before using it to produce reported results, and take full responsibility for the content of this work.

\subsection*{Ethics statement}
This work uses only public, de-identified data: the real perturbation dataset \citep{kang_multiplexed_2018} was collected and published by others, and no new human-subject data was collected for this study. The attributions produced by our method are intended to prioritize candidate causal genes for downstream experimental follow-up, not to be interpreted as definitive causal or clinical claims; any prioritized targets should be validated experimentally before drawing biological or clinical conclusions.

\subsection*{Reproducibility statement}
Our implementation of CIG/CEG and the code for the synthetic CATE benchmark (Section~\ref{sec:causal_attributions_comparison}) are available at \url{https://github.com/sifrimlab/CausalIG}. Full implementation and training details for the Dyngen and IFN-$\beta$ experiments — including preprocessing, hyperparameters, and dataset statistics — are given in Appendix~\ref{appendix:experimental_setup} and the corresponding experiment-specific appendices (Appendices~\ref{appendix:gene_recovery_extra_dyngen} and \ref{appendix:gene_recovery_extra_ifnb}); both the Dyngen simulator and the IFN-$\beta$ dataset \citep{kang_multiplexed_2018} are themselves publicly available. Complete proofs of all theoretical claims, including the axiomatic properties and complexity analysis, are provided in Appendices~\ref{appendix:axiomatic_properties} and \ref{appendix:complexity}.

\subsubsection*{Acknowledgments}
Anis Ismail, Joran Michiels, and Alejandro Sifrim are funded by the Leuven Future Fund (LISCO-BIOMED) and the Opening the Future Fund.
Maarten De Vos and Joran Michiels received funding from the Flemish Government (AI Research Program).
Maarten De Vos additionally received funding from the Flemish Government under the METHUSALEM grant METH/26/003 ``Methusalem-BioMedAI --- Explainable and generative AI for accelerating biomedical discoveries: from genome to function'', and from the FWO Research Project ``Artificial Intelligence (AI) for data-driven personalised medicine'' (G0C9623N).
The authors are affiliated with Leuven.AI --- KU Leuven institute for AI, B-3000, Leuven, Belgium.

We would like to thank Guido Gagliardi for his help with the initial theoretical methodology and Jelle Jacobs for the discussions on biological validity.

\bibliographystyle{plainnat}
\bibliography{refs}

\appendix
\counterwithin{figure}{section}
\counterwithin{table}{section}

\section{Implementation for longer dependency paths} \label{appendix:multiple_hop_implementation}
The topological ordering used to define the path is a bookkeeping device, not a modeling choice: any two valid topological orderings of the same DAG agree on every parent-child pair, so $R_i$ is invariant to which one is used.

Indirect attributions through dependency paths of length two (referred to as second-order) can be incorporated by extending the formula with additional terms:
\begin{equation} \label{eq:attributions_fully_connected_dag_two_hop}
\begin{aligned}
    R_i &= \int_{0}^{1} \frac{\partial F}{\partial \gamma_i} \frac{\partial \gamma_i}{\partial \alpha} \, d\alpha + \sum_{j > i} \int_{0}^{1} \frac{\partial F}{\partial \gamma_j} \frac{\partial \gamma_j}{\partial \gamma_i} \frac{\partial \gamma_i}{\partial \alpha} \, d\alpha + \sum_{\substack{j > i \\ i < k < j}} \int_{0}^{1} \frac{\partial F}{\partial \gamma_j} \frac{\partial \gamma_j}{\partial \gamma_k} \frac{\partial \gamma_k}{\partial \gamma_i} \frac{d\gamma_i}{d \alpha} \, d\alpha.
\end{aligned}
\end{equation}
The last (two-hop) term uses the full derivative $\frac{d\gamma_i}{d\alpha}$ rather than the local $\frac{\partial\gamma_i}{\partial\alpha}$ used in the direct and one-hop terms: it is the deepest dependency explicitly tracked, so it must fold in any effect of $x_i$'s own ancestors that is not otherwise given its own term, exactly as the one-hop formula's indirect term already does for its (only) tracked layer. Paths of greater length can be handled analogously, tracking one additional layer with local partials and moving the full-derivative fold to the new deepest layer; our current implementation stops at two hops, which sufficed in all our experiments. This construction keeps $\sum_i R_i = F(\boldsymbol{x})-F(b)$ exact at any truncation depth $H$ (the number of explicitly tracked hops): the untracked remainder beyond the deepest tracked layer is always folded into that layer's full derivative rather than discarded.

\paragraph{The general $H$-hop formula.} The one-hop (Equation~\ref{eq:attributions_fully_connected_dag}) and two-hop (Equation~\ref{eq:attributions_fully_connected_dag_two_hop}) formulas above are the $H=1$ and $H=2$ instances of one pattern. Writing $(J^m)_{ji}$ for the $(j,i)$ entry of the $m$-th power of the local-Jacobian matrix $J_{ji}=\partial\gamma_j/\partial\gamma_i$ — the sum, over every directed path of length exactly $m$ from $i$ to $j$, of the product of $J$ along it, with $(J^0)_{ji}=\delta_{ji}$ — this pattern is, for general $H\ge1$,
\begin{equation}
\begin{aligned}
R_i^{(H)} &= \int_0^1\frac{\partial F}{\partial\gamma_i}\frac{\partial\gamma_i}{\partial\alpha}\,d\alpha
+ \sum_{m=1}^{H-1}\sum_{j:\,(J^m)_{ji}\neq0}\int_0^1\frac{\partial F}{\partial\gamma_j}(J^m)_{ji}\frac{\partial\gamma_i}{\partial\alpha}\,d\alpha \\
&\quad+ \sum_{j:\,(J^H)_{ji}\neq0}\int_0^1\frac{\partial F}{\partial\gamma_j}(J^H)_{ji}\frac{d\gamma_i}{d\alpha}\,d\alpha.
\end{aligned}
\label{eq:h-hop-general}
\end{equation}
Appendix~\ref{appendix:axiomatic_properties} proves this satisfies the same axioms as the one-hop and two-hop instances above, at every $H$.

\section{Details of the basis expansion} \label{appendix:basis_expansion}
To increase the expressive capacity of the surrogate while preserving interpretability and edge-wise decomposition, each parent feature is transformed through a structured basis expansion applied independently per feature. Let $x_i$ denote the value of feature $i$. Inputs are first standardized:
\begin{equation}
z_i = \frac{x_i - \mu_i}{\sigma_i},
\end{equation}
where $\mu_i$ and $\sigma_i$ are the feature-wise mean and standard deviation estimated from the training data. Standardization ensures that the basis functions operate on comparable scales across features and improves numerical stability during optimization.

The resulting basis representation $\phi(x_i)$ combines three complementary function families:
\begin{equation}
\phi(x_i) =
\left[
\underbrace{z_i, z_i^2, \dots, z_i^D}_{\text{Polynomial basis}},
\underbrace{\left( \max(0, z_i - c_k) \right)^p,\ k=1,\dots,K_t}_{\text{Truncated power basis}},
\underbrace{\sin(\omega_f z_i), \cos(\omega_f z_i),\ f=1,\dots,K_f}_{\text{Fourier basis}}
\right].
\end{equation}

The polynomial component models smooth low-order nonlinearities. For a polynomial degree $D$, the expansion contributes:
\begin{equation}
\phi_{\text{poly}}(x_i)
=
[z_i, z_i^2, \dots, z_i^D],
\end{equation}
yielding $D$ basis functions per feature.

The truncated power basis introduces localized nonlinear behavior through learnable knot locations $\{c_k\}_{k=1}^{K_t}$. Each basis element is defined as:
\begin{equation}
\phi_{\text{tp},k}(x_i)
=
\left( \max(0, z_i - c_k) \right)^p,
\end{equation}
where $p$ is the knot power controlling smoothness and curvature. These functions behave similarly to spline basis functions and allow the surrogate to capture sharp transitions or regime-specific effects. This component contributes $K_t$ basis functions per feature.

The Fourier component captures periodic structure:
\begin{equation}
\phi_{\text{fourier}}(x_i)
=
[\sin(\omega_1 z_i), \cos(\omega_1 z_i), \dots,
\sin(\omega_{K_f} z_i), \cos(\omega_{K_f} z_i)],
\end{equation}
where $\omega_f$ denotes the frequency associated with Fourier feature $f$; these frequencies are set as hyperparameters. Since both sine and cosine terms are included for each frequency, this component contributes $2K_f$ basis functions per feature.

The total number of basis functions per feature is:
\begin{equation}
Q = D + K_t + 2K_f.
\end{equation}
This construction provides a flexible yet structured functional approximation that can model smooth, localized, and periodic dependencies while preserving edge-level interpretability within the directed graph.

\section{Axiomatic properties} \label{appendix:axiomatic_properties} \label{appendix:sensitivity_a_proof}

Since our method belongs to the family of path methods \citep{sundararajan2017axiomatic}, it inherits many of the axiomatic properties of integrated gradients. We discuss each property in turn: a statement, a proof (or, for the one axiom we do not satisfy, an explanation of why), and, where our method requires more than the standard axiom does, the additional preconditions this needs. We would expect other dependency-aware attribution methods (e.g.\ causal Shapley values, \citealp{heskes2020causal}; conditional SHAP, \citealp{aas2021explaining}) to require analogous preconditions on the supplied dependency structure, though to our knowledge this has not previously been examined for those methods. Proofs are given for the one-hop formula (Equation~\ref{eq:attributions_fully_connected_dag}); implementation invariance, linearity, and sensitivity(a) carry over unchanged to the general $H$-hop formula (Equation~\ref{eq:h-hop-general}), since its extra factors depend only on the graph and mechanisms, never on $F$, and its direct term is the same for every $H$. Completeness and symmetry need a short additional argument, given at the end of their proofs.

\subsection*{Completeness}
\textbf{Statement.} The sum of attributions equals the difference in model output between the input and the baseline: $\sum_i R_i = F(\boldsymbol{x}) - F(b)$.

\textit{Proof.} By the gradient theorem (fundamental theorem of calculus along the path $\gamma$),
\begin{equation}
\int_0^1 \nabla F(\gamma(\alpha)) \cdot \frac{d\gamma(\alpha)}{d\alpha}\,d\alpha = F(\gamma(1)) - F(\gamma(0)) = F(\boldsymbol{x}) - F(b).
\end{equation}
Expanding the dot product into per-feature terms as in Equation~\ref{eq:attributions}, and grouping each feature $i$'s direct and indirect contributions as in Equation~\ref{eq:attributions_fully_connected_dag}, gives $\sum_i R_i = F(\boldsymbol{x}) - F(b)$. The fundamental theorem of calculus makes no reference to how $\gamma$ is parametrized internally, so this holds exactly, regardless of the causal graph or which differentiable function is used for the mechanism $c_i$ — whether the true structural mechanism or a learned surrogate $s_i$ approximating it. $\square$

\paragraph{Completeness at any truncation depth.} The one-hop (Equation~\ref{eq:attributions_fully_connected_dag}) and two-hop (Equation~\ref{eq:attributions_fully_connected_dag_two_hop}) formulas are the $H=1$ and $H=2$ instances of the general $H$-hop formula, Equation~\ref{eq:h-hop-general} (Appendix~\ref{appendix:multiple_hop_implementation}), which tracks dependencies explicitly up to $H$ hops with the local partial $\partial\gamma_i/\partial\alpha$ and folds every hop beyond $H$ into that boundary layer's full derivative $d\gamma_i/d\alpha$. We show $\sum_i R_i^{(H)}=F(\boldsymbol{x})-F(b)$ for every $H\ge1$. Writing $\tilde J := (I-J)^{-1} = \sum_{m\ge0}J^m$ (a finite sum, by nilpotency of $J$), $\tilde J = \sum_{m=0}^{H-1}J^m + J^H\tilde J$ for every $H$ (both sides equal $\sum_{m\ge0}J^m$). Multiplying by $\partial\gamma/\partial\alpha$ and using $d\gamma/d\alpha = \tilde J\,\partial\gamma/\partial\alpha$ gives, component-wise, $\frac{d\gamma_j}{d\alpha} = \sum_{m=0}^{H-1}\sum_i(J^m)_{ji}\frac{\partial\gamma_i}{\partial\alpha} + \sum_i(J^H)_{ji}\frac{d\gamma_i}{d\alpha}$ exactly, for every $H$. Substituting this into the completeness identity above and regrouping by attribution index $i$ recovers exactly $\sum_i R_i^{(H)}$, so $\sum_i R_i^{(H)}=F(\boldsymbol{x})-F(b)$ holds at any truncation depth $H$ — provided the deepest tracked layer uses the full derivative, as in Equations~\ref{eq:attributions_fully_connected_dag} and \ref{eq:attributions_fully_connected_dag_two_hop}.

\paragraph{Practical numerical error.} The proof above is exact in the continuous limit for any differentiable choice of mechanism $c_i$, since the path itself does not depend on the accuracy of that choice. Two sources of numerical error remain when evaluating it in practice: (i) Riemann-sum discretization over $K$ steps, identical in kind to vanilla integrated gradients, and (ii) the backward finite difference used for $\partial\gamma_i/\partial\alpha$ (Equation~\ref{eq:finite_difference}), which vanilla IG never needs since its path derivative is closed-form — genuinely new to our method, but vanishing as $\Delta\alpha \to 0$. Implementation invariance, symmetry, and linearity (below) have no such gap, since they hold exactly for any $K$. None of our experiments apply post-hoc normalization; all reported attributions are raw.

\begin{figure}[h]
    \centering
    \includegraphics[width=0.6\textwidth]{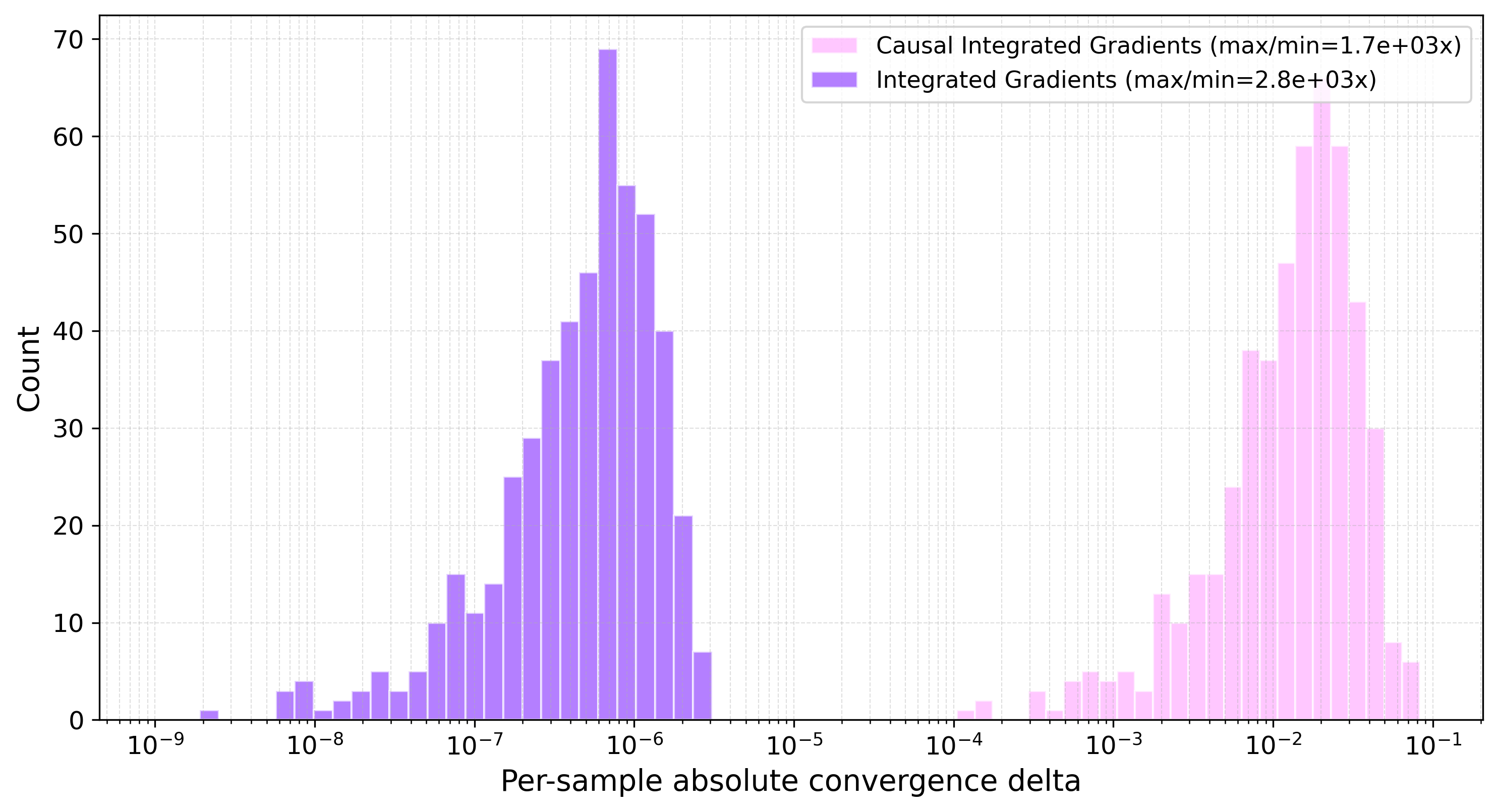}
    \caption{Per-sample absolute completeness delta $|\delta^{(n)}|$ at 774 features ($n=500$, pooled across 5 seeds). Median: CIG $1.48\times10^{-2}$, IG $5.74\times10^{-7}$. Proportional spread (max/min): CIG $1.71\times10^3\times$, IG $2.77\times10^3\times$ — the same order of magnitude.}
    \label{fig:per_sample_convergence_delta}
\end{figure}

\begin{figure}
    \centering
    \includegraphics[width=\textwidth]{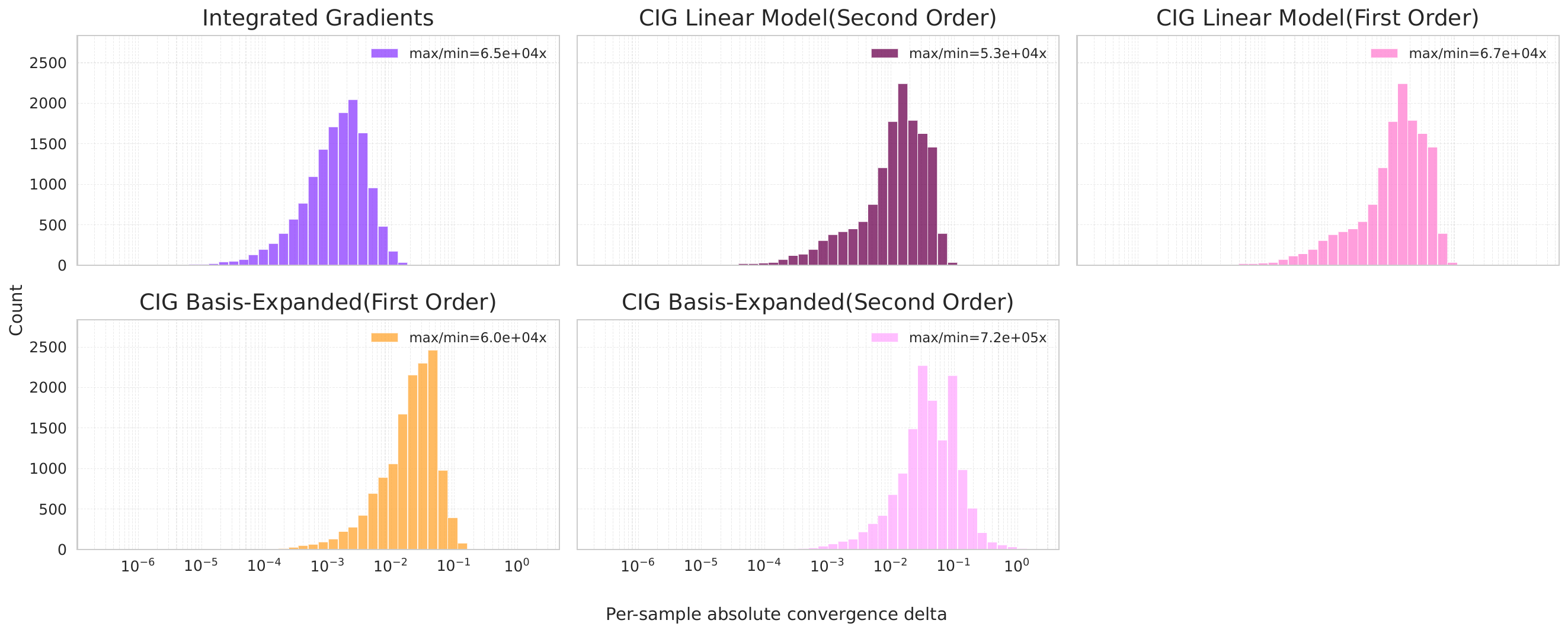}
    \caption{Per-sample absolute completeness delta $|\delta^{(n)}|$ for the Dyngen dataset ($n=14{,}076$, pooled across 5 seeds).}
    \label{fig:completeness_delta_histograms_dyngen}
\end{figure}

\begin{figure}[h]
    \centering
    \includegraphics[width=0.6\textwidth]{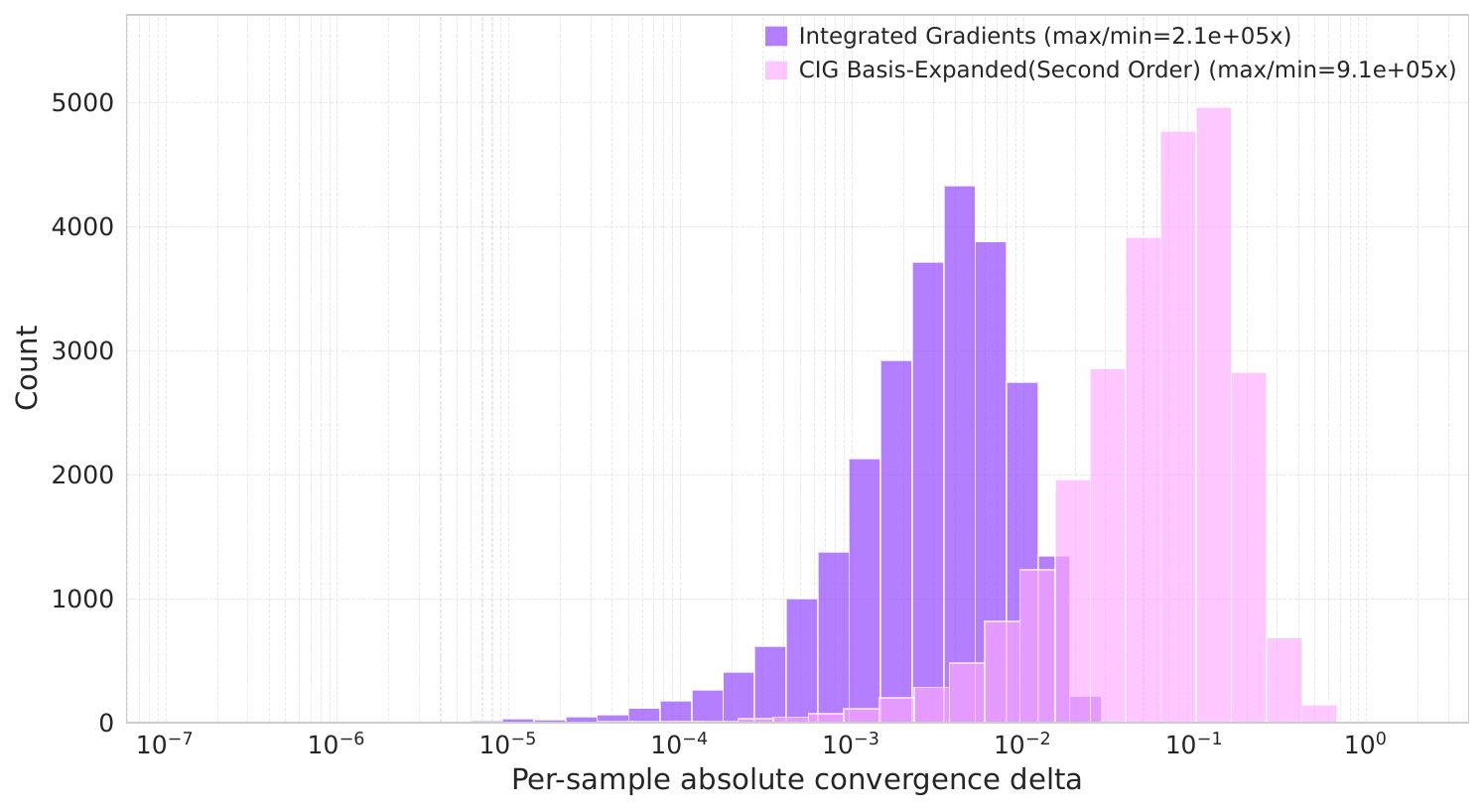}
    \caption{Per-sample absolute completeness delta $|\delta^{(n)}|$ for the IFN-$\beta$ dataset (6,941 features; $n=25{,}450$, pooled across 5 seeds). Median: CIG $6.75\times10^{-2}$, IG $3.34\times10^{-3}$. Proportional spread (max/min): CIG $9.05\times10^{5}\times$, IG $2.11\times10^{5}\times$ — the same order of magnitude.}
    \label{fig:per_sample_convergence_delta_kang}
\end{figure}

We quantify this residual with the per-sample completeness delta $\delta^{(n)} = \sum_i R_i^{(n)} - (F(x^{(n)})-F(b^{(n)}))$. On the synthetic benchmark ($K{=}100$), we report it relative to the output difference (Table~\ref{tab:completeness_delta}), computed batch-wise as a weighted absolute percentage error (WAPE), $\sum_n |\delta^{(n)}| \big/ \sum_n |F(x^{(n)})-F(b^{(n)})|$ over evaluation samples $n$ within each seed, then averaged (mean $\pm$ std) across seeds. We use WAPE rather than the per-sample mean absolute percentage error (MAPE), since the latter is unstable whenever a single sample's $F(x^{(n)})-F(b^{(n)})$ is near zero; WAPE is the standard fix for this in forecasting \citep{kolassa2007advantages}. To also show the absolute magnitude of the residual, Figure~\ref{fig:per_sample_convergence_delta} gives the distribution of $|\delta^{(n)}|$ at one representative setting (774 synthetic features). For the two biological datasets, where near-zero output differences are more common, we report only this absolute delta (Figures~\ref{fig:completeness_delta_histograms_dyngen} and \ref{fig:per_sample_convergence_delta_kang}), which sidesteps the ratio entirely.

\begin{table}[H]
\centering
\caption{Relative completeness delta (WAPE), mean $\pm$ std over seeds.}
\label{tab:completeness_delta}
\begin{tabular}{lcc}
\toprule
Setting & CIG & IG \\
\midrule
Synthetic, 100 features & $1.90\times10^{-3} \pm 1.16\times10^{-3}$ & $5.77\times10^{-8} \pm 6.45\times10^{-9}$ \\
Synthetic, 129 features & $1.40\times10^{-3} \pm 3.78\times10^{-4}$ & $5.53\times10^{-8} \pm 2.34\times10^{-9}$ \\
Synthetic, 166 features & $1.38\times10^{-3} \pm 2.99\times10^{-4}$ & $5.96\times10^{-8} \pm 4.62\times10^{-9}$ \\
Synthetic, 215 features & $1.60\times10^{-3} \pm 4.11\times10^{-4}$ & $6.56\times10^{-8} \pm 4.21\times10^{-9}$ \\
Synthetic, 278 features & $1.75\times10^{-3} \pm 5.08\times10^{-4}$ & $6.59\times10^{-8} \pm 3.51\times10^{-9}$ \\
Synthetic, 359 features & $1.96\times10^{-3} \pm 3.64\times10^{-4}$ & $7.09\times10^{-8} \pm 5.30\times10^{-9}$ \\
Synthetic, 464 features & $1.91\times10^{-3} \pm 3.30\times10^{-4}$ & $7.49\times10^{-8} \pm 1.15\times10^{-8}$ \\
Synthetic, 599 features & $1.82\times10^{-3} \pm 2.17\times10^{-4}$ & $8.22\times10^{-8} \pm 3.17\times10^{-9}$ \\
Synthetic, 774 features & $1.99\times10^{-3} \pm 3.83\times10^{-4}$ & $7.59\times10^{-8} \pm 5.66\times10^{-9}$ \\
Synthetic, 1000 features & $1.95\times10^{-3} \pm 2.64\times10^{-4}$ & $8.01\times10^{-8} \pm 5.38\times10^{-9}$ \\
\bottomrule
\end{tabular}
\end{table}

CIG's delta is consistently one or more orders of magnitude larger than IG's, as expected rather than indicative of worse attribution quality: IG's path derivative is closed-form, while CIG additionally estimates the surrogate-residual derivative via finite differences, an extra discretization-error layer IG lacks. Real-data residuals exceed the synthetic ones: on synthetic data the mechanism is the exact generating function, so $h_i(\alpha)$ reduces to pure exogenous noise, whereas on real data $h_i(\alpha)$ additionally conflates the surrogate's fitting error, which need not vary smoothly with $\alpha$ along the path. Since all our evaluation metrics depend only on how attributions rank features, a residual of this size is unlikely to affect them.

\begin{figure}[h]
    \centering
    \includegraphics[width=0.55\textwidth]{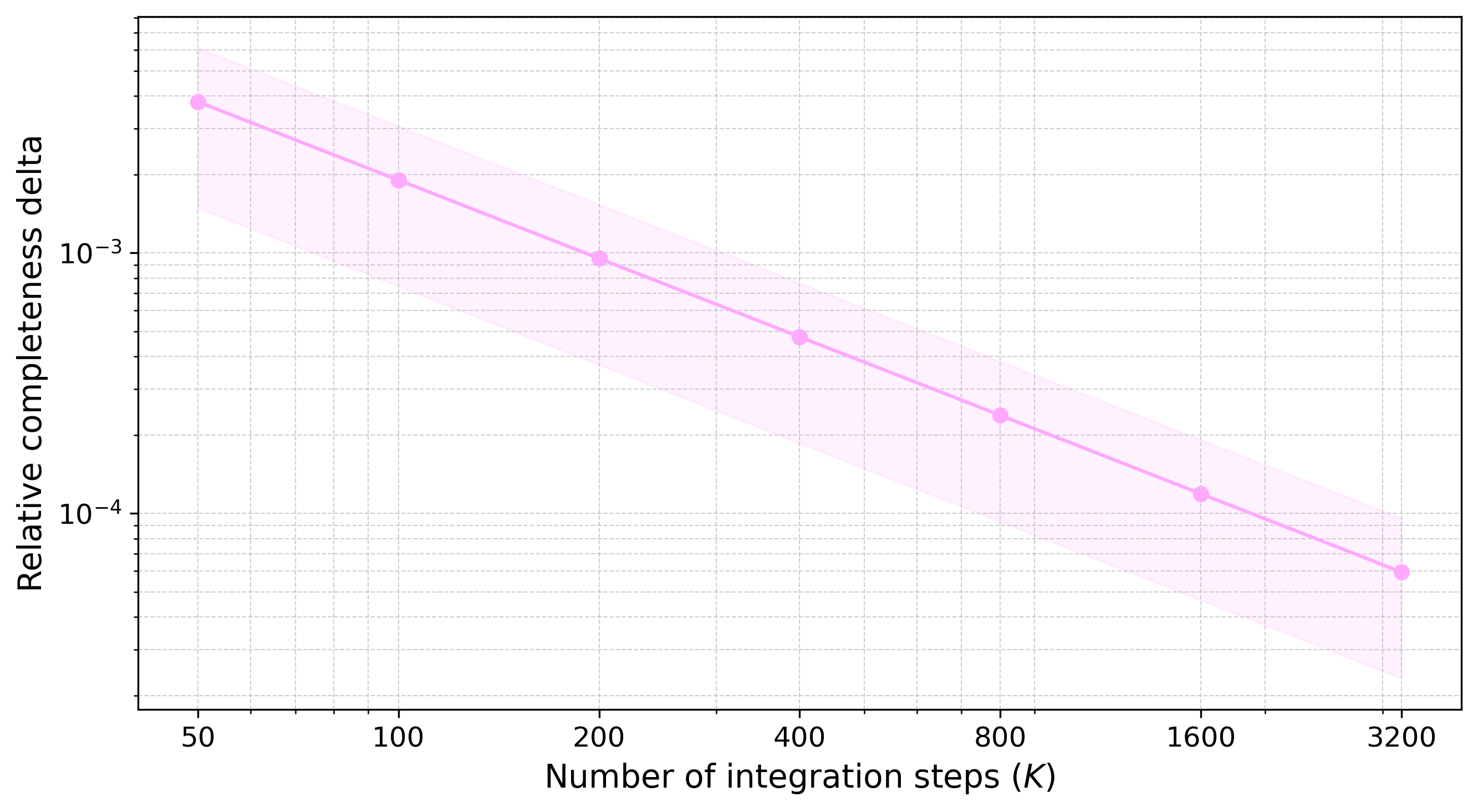}
    \caption{Relative completeness delta of CIG as a function of the number of integration steps $K$, at 100 synthetic features (log-log axes). The shaded band shows the standard deviation over seeds. The linear trend on log-log axes is the $O(1/K)$ signature of first-order discretization error.}
    \label{fig:completeness_delta_vs_nsteps}
\end{figure}

This residual converges as expected for a first-order discretization scheme: Figure~\ref{fig:completeness_delta_vs_nsteps} sweeps $K$ at 100 synthetic features, showing the relative completeness delta falling off linearly on a log-log plot, roughly halving with every doubling of $K$ — the textbook $O(1/K)$ signature shared by both the Riemann sum and the finite-difference derivative. This sweep does not isolate which of the two dominates, only that the combined numerical error vanishes at the expected rate.

\subsection*{Implementation invariance}
\textbf{Statement.} Attributions are invariant to the implementation of the model, depending only on the input-output behavior of $F$.

\textbf{Precondition.} The causal graph and the mechanism's input-output relation ($c_i$ as a function, or $s_i$ if using surrogates, not its specific implementation) are held fixed; only the model $F$ varies.

\textit{Proof.} Let $F, F'$ be two models with identical input-output behavior, i.e.\ $F(z) = F'(z)$ for all $z$, evaluated under the same causal graph and the same mechanisms $c_i$ (or surrogates $s_i$). The path points $\gamma^{(p)}$ visited by the Riemann sum, and the gradients $\partial\gamma_i/\partial\alpha$, $\partial\gamma_j/\partial\gamma_i$ evaluated at those points (Equation~\ref{eq:finite_difference}), depend only on the graph, $\alpha^{(p)}$, and $c_i$ (or $s_i$) — never on $F$. Hence at every path point $p$,
\begin{equation}
\frac{\partial F(\gamma^{(p)})}{\partial \gamma_i} = \frac{\partial F'(\gamma^{(p)})}{\partial \gamma_i}
\end{equation}
follows directly from $F \equiv F'$. Since every term of Equation~\ref{eq:attributions_fully_connected_dag} involves $F$ only through such partial derivatives evaluated at these shared path points, $R_i^F = R_i^{F'}$ for every feature $i$ and every $K$. $\square$

This precondition is not an additional restriction beyond the standard axiom — it is what implementation invariance means: comparing two models under one shared explanation setup, not two different setups. Changing the graph or the mechanisms is meant to change attributions.

\subsection*{Sensitivity(a)}
\textbf{Statement.} If the input and baseline differ in exactly one feature $i$ and yield different predictions, then feature $i$ should receive a non-zero attribution. We prove this for the direct attribution component.

\textit{Proof.}
Since the line integral uses a straight-line path between baseline and input, all features $j \neq i$ remain constant along the path, so their total derivatives vanish:
\begin{equation}
\frac{d\gamma_j}{d\alpha} = 0, \quad j \neq i.
\end{equation}

By the chain rule, the total derivative of the varying feature $\gamma_i$ along the path is:
\begin{equation}
\frac{d\gamma_i}{d\alpha} = \frac{\partial \gamma_i}{\partial \alpha} + \sum_{j \neq i} \frac{\partial \gamma_i}{\partial \gamma_j} \frac{d\gamma_j}{d\alpha} = \frac{\partial \gamma_i}{\partial \alpha} + \sum_{j \neq i} \frac{\partial \gamma_i}{\partial \gamma_j} \cdot 0 = \frac{\partial \gamma_i}{\partial \alpha}.
\end{equation}

Consequently, the direct attribution of feature $i$ reduces to the standard integrated gradients attribution:
\begin{equation}
R_{i, \mathrm{direct}}^{\mathrm{CIG}} = \int_{0}^{1} \frac{\partial F}{\partial \gamma_i} \frac{\partial \gamma_i}{\partial \alpha} \, d\alpha = \int_{0}^{1} \frac{\partial F}{\partial \gamma_i} \frac{d\gamma_i}{d\alpha} \, d\alpha = R_i^{\mathrm{IG}}.
\end{equation}

Since integrated gradients satisfies sensitivity(a), it follows that the direct attribution component of causal integrated gradients does as well. $\square$

\subsection*{Symmetry}
\textbf{Statement.} If two features have the same value and baseline and are symmetric with respect to $F$, they receive equal attributions.

\textbf{Precondition.} Beyond the standard requirement (that $F$ be symmetric in features $i,k$ with $x_i=x_k$, $b_i=b_k$), our method additionally requires: (S1) $i$ and $k$ have identical parents and identical children in the causal graph — which also implies they cannot be linked by an edge, since $i\to k$ would place $i$ in $k$'s parent set but not its own; and (S2) the mechanism is exchangeable ($c_i\equiv c_k$ as functions, or $s_i\equiv s_k$ if using surrogates, not necessarily identical implementation), with every shared child's mechanism symmetric in $i,k$.

\textit{Proof.} By (S1), $i$ and $k$ share the same parent set. Together with $x_i=x_k$, $b_i=b_k$, and $c_i\equiv c_k$ (S2) applied to that shared parent set, an induction over the topological order gives $\gamma_i^{(p)} = \gamma_k^{(p)}$ at every path point $p$: the base case holds for root features by $x_i=x_k$, $b_i=b_k$ directly, and the inductive step holds because $i$ and $k$ receive identical inputs from their (identical, by (S1)) parents and apply the same mechanism (S2). Consequently $\partial\gamma_i/\partial\alpha = \partial\gamma_k/\partial\alpha$ at every path point, so the direct terms of $R_i$ and $R_k$ agree, since $F$ is symmetric in $i,k$ by hypothesis. For the indirect terms, (S1) gives $i,k$ identical children, and (S2) makes every shared child's mechanism symmetric in $i,k$, so $\partial\gamma_j/\partial\gamma_i = \partial\gamma_j/\partial\gamma_k$ for every shared child $j$; combined with $F$'s symmetry in $i,k$, the indirect sums in Equation~\ref{eq:attributions_fully_connected_dag} agree term-by-term. Hence $R_i = R_k$. $\square$

If $i,k$ occupy different positions in the graph, our method attributes them asymmetrically by design, since it is meant to reflect their differing causal roles.

\paragraph{Symmetry at any truncation depth.} The same preconditions (S1)+(S2) suffice for $R_i^{(H)}=R_k^{(H)}$ (Equation~\ref{eq:h-hop-general}) at every $H$, not only $H=1$. Two extensions of the argument above are needed. First, since $\gamma_i^{(p)}\equiv\gamma_k^{(p)}$ holds identically as a function along the path, not merely pointwise, their derivatives agree at every order too, so $\frac{d\gamma_i}{d\alpha}=\frac{d\gamma_k}{d\alpha}$ in addition to $\frac{\partial\gamma_i}{\partial\alpha}=\frac{\partial\gamma_k}{\partial\alpha}$. Second, $(J^m)_{ji}=(J^m)_{jk}$ for every $m\ge1$ and every $j$: decomposing a length-$m$ path from $i$ by its first step, $(J^m)_{ji}=\sum_{l\,\in\,\mathrm{children}(i)}(J^{m-1})_{jl}J_{li}$, and since $i,k$ have identical children (S1) and $J_{li}=J_{lk}$ for every shared child $l$ (established above), this equals $\sum_{l\,\in\,\mathrm{children}(k)}(J^{m-1})_{jl}J_{lk}=(J^m)_{jk}$ — one identity valid for every $m$ simultaneously, not an induction on $m$. Combining both facts, every term of $R_i^{(H)}$ — direct, every middle layer, and the boundary layer — matches the corresponding term of $R_k^{(H)}$, so $R_i^{(H)}=R_k^{(H)}$ for every $H$.

\subsection*{Linearity}
\textbf{Statement.} If $F = aF_1 + bF_2$, the attributions for $F$ equal the same linear combination of the attributions for $F_1$ and $F_2$: $R_i^F = a R_i^{F_1} + b R_i^{F_2}$.

\textbf{Precondition.} $F_1$ and $F_2$ share the same causal graph and the same mechanism's input-output relation ($c_i$ as a function, or $s_i$ if using surrogates, not its specific implementation), differing only in how they are combined into $F$.

\textit{Proof.} Under this precondition, the path points $\gamma^{(p)}$ and path derivatives $\partial\gamma_i/\partial\alpha$, $\partial\gamma_j/\partial\gamma_i$ are identical for $F_1$ and $F_2$ (they depend only on the shared graph and mechanisms). By linearity of the partial derivative,
\begin{equation}
\frac{\partial F}{\partial \gamma_i} = a\,\frac{\partial F_1}{\partial \gamma_i} + b\,\frac{\partial F_2}{\partial \gamma_i}
\end{equation}
at every path point. Substituting into Equation~\ref{eq:attributions_fully_connected_dag} — both the direct term and every indirect term, which use only these shared path derivatives — gives $R_i^F = a R_i^{F_1} + b R_i^{F_2}$ for every feature $i$. $\square$

This axiom is mainly of practical interest for ensembles: since the graph and mechanisms are properties of the data rather than of any individual ensemble member, an ensemble $F=aF_1+bF_2$ trained on one dataset satisfies the precondition automatically, so linearity holds without needing to verify anything further. It would only fail if $F_1,F_2$ were deliberately given different graphs or different mechanisms — a scenario with little practical relevance.

\subsection*{Dummy}
\textbf{Statement.} A feature that does not appear in $F$ should receive zero attribution.

Our method does not satisfy this axiom. A feature absent from $F$ but causally upstream of a feature that does appear can receive a non-zero indirect attribution: in the two-feature example in the right panel of Figure~\ref{fig:chain_rule_diagrams}, if $F_1$ depends only on $x_2$ and $x_2$ depends on $x_1$, then $x_1$ receives the non-zero indirect term of Equation~\ref{eq:attributions_f1} even though $F_1$ never reads $x_1$ directly. This is a deliberate design choice — capturing such upstream influence is precisely the goal of incorporating causal dependencies — and the same behavior is exhibited by other dependency-aware methods such as conditional SHAP \citep{aas2021explaining} and causal Shapley values \citep{heskes2020causal}.

\section{Computational complexity analysis} \label{appendix:complexity}

We give a theoretical, not just empirical, complexity comparison against causal Shapley values \citep{heskes2020causal} and conditional SHAP \citep{aas2021explaining}. Let $d$ be the number of input features, $C$ the number of output classes attributed (typically 1), $K$ the number of Riemann integration steps, $M$ the number of SHAP permutations, and $E$ the number of edges in the causal graph. Throughout, we assume a fixed-width, fixed-depth network, so a single forward(+backward) pass costs $O(d)$, dominated by the input/output layers rather than by hidden width; every entry below inherits this. Table~\ref{tab:complexity} gives the cost per explained instance.

\begin{table}[H]
\centering
\caption{Complexity per explained instance.}
\label{tab:complexity}
\begin{tabularx}{\linewidth}{l X}
\toprule
Method & Cost \\
\midrule
IG (vanilla) & $O(K \cdot d \cdot C)$ \\
CIG (ours) & $O(K \cdot d \cdot C)$ for a sparse causal graph ($E{=}O(d)$); $O(K \cdot d^2)$ for a dense graph (worst case) \\
Causal Shapley values (sampled, as implemented) & $O(M \cdot d^2)$ \\
Conditional SHAP (sampled, as implemented) & $O(M \cdot d^3)$ (one-time $O(d^3)$ Cholesky) \\
Exact Shapley value (causal or conditional) & $O(2^d)$ \\
\bottomrule
\end{tabularx}
\end{table}

CIG's cost beyond a single $O(d)$ forward/backward pass comes from the surrogate Jacobian needed for the indirect term. With a dense causal graph, computing this Jacobian via autograd needs one backward pass per output feature — $d$ passes total, each itself $O(d)$ — giving $O(d^2)$: this is the generic cost of any dense $d\to d$ Jacobian, not specific to our method. With a sparse causal graph (low mean degree), our sparse masked surrogate's Jacobian is computed analytically via the chain rule, evaluating derivatives only for active edges — touching only the $O(d)$ entries that actually exist, and skipping this $d$-fold blow-up entirely. Our synthetic scaling experiments (right panel of Figure~\ref{fig:causal_scaling}) in fact compute the full (dense) Jacobian rather than the closed-form sparse one, which is why the measured runtimes there follow the $O(d^2)$ line above rather than the $O(d)$ sparse case. Our real graphs support the sparse regime: the OmniPath-derived network (6,941 genes) has mean degree 3.38, and the Dyngen GRN (2,300--2,500 genes) has mean degree 3.07, both low and, notably, not growing with graph size, though both also contain a small number of high-degree hub genes (up to several hundred children each), consistent with the heavy-tailed degree distributions typical of real regulatory networks; what the $O(d)$ argument needs is the low mean, not a uniform cap on every node.

\paragraph{Second-order indirect effects.} The optional second-order extension (Appendix~\ref{appendix:multiple_hop_implementation}), which attributes effects mediated through two successive surrogate applications rather than one, applies the same Jacobian-and-contraction step a second time and so does not change CIG's complexity class in Table~\ref{tab:complexity} — only its constant, roughly $2\times$ the first-order-only cost.

Causal Shapley values and conditional SHAP, as commonly implemented via permutation sampling, are quadratic and cubic in $d$, respectively (conditional SHAP pays an extra factor of $d$ for Gaussian conditioning per sample). The exact Shapley value is exponential, $O(2^d)$, regardless of sparsity. So CIG's polynomial cost is exact (up to ordinary path discretization) and improves with sparsity, while the Shapley baselines' polynomial cost is already a sampling approximation of an otherwise exponential quantity. This is not only a theoretical gap: at the scale of our real Dyngen and IFN-$\beta$ experiments, causal Shapley values and conditional SHAP are excluded outright as computationally infeasible, and on our larger synthetic benchmarks conditional SHAP is excluded and causal Shapley values must run on a reduced sampling budget, so the polynomial/exponential gap above corresponds to the difference between applications that are and are not reachable at all in practice.

One more cost worth disclosing: surrogate training. With $n$ the training-set size and $T$ epochs, our closed-form sparse surrogate trains in $O(T\cdot n\cdot d)$ (sparse graph, $E{=}O(d)$) or $O(T\cdot n\cdot d^2)$ (dense); a generic dense-MLP surrogate is $O(T\cdot n\cdot d)$ regardless of sparsity. This is a one-time, offline cost amortized over all explanations, not a per-instance one: over $N_{\text{explain}}$ instances it adds (training cost)$/N_{\text{explain}}$ to each, vanishing as $N_{\text{explain}}$ grows.

\section{Experimental setup and end-to-end cost accounting} \label{appendix:experimental_setup}
All experiments were run on a server equipped with two Intel Xeon Gold 6438Y+ CPUs (32 cores each), two NVIDIA A40 GPUs (46 GB VRAM each), and approximately 1 TB of system RAM. We report here a full end-to-end cost breakdown (surrogate training vs.\ CIG's own attribution computation) for both biological experiments at $K{=}100$ integration steps in Table~\ref{tab:profiling}.

\begin{table}[H]
\centering
\caption{End-to-end pipeline profiling: wall-clock time and peak CUDA memory per stage, for the full test set at $K{=}100$ integration steps. Dyngen benchmark ($2{,}514$--$3{,}000$ test cells per seed); IFN-$\beta$ ($5{,}090$ cells $\times$ $6{,}941$ genes). Values are mean $\pm$ std over 5 seeds. CEG averages over $100$ random baselines; all CIG rows use a zero baseline.}
\label{tab:profiling}
\resizebox{\textwidth}{!}{%
\begin{tabular}{lllrrr}
\toprule
Dataset & Method & Stage & Time (s) & Peak CUDA alloc.\ (MB) & Peak CUDA reserved (MB) \\
\midrule
Dyngen & CIG (linear basis, 1st order) & Surrogate training & 129.7 $\pm$ 4.2 ($\sim$130\,s) & 224 $\pm$ 7 & 980 $\pm$ 958 \\
 &  & Attribution computation & 175.4 $\pm$ 19.9 ($\sim$175\,s) & 17818 $\pm$ 927 & 18680 $\pm$ 1333 \\
 &  & \textbf{Total} & \textbf{305.1 $\pm$ 23.6 ($\sim$305\,s)} & \textbf{17818 $\pm$ 927} & \textbf{18680 $\pm$ 1333} \\
\midrule
Dyngen & CIG (linear basis, 2nd order) & Surrogate training & 130.4 $\pm$ 5.0 ($\sim$130\,s) & 221 $\pm$ 12 & 966 $\pm$ 970 \\
 &  & Attribution computation & 349.6 $\pm$ 39.6 ($\sim$350\,s) & 26653 $\pm$ 1416 & 27500 $\pm$ 2004 \\
 &  & \textbf{Total} & \textbf{480.0 $\pm$ 44.6 ($\sim$480\,s)} & \textbf{26653 $\pm$ 1416} & \textbf{27500 $\pm$ 2004} \\
\midrule
Dyngen & CIG (expanded basis, 1st order) & Surrogate training & 1246.4 $\pm$ 35.4 ($\sim$21\,min) & 8966 $\pm$ 305 & 12818 $\pm$ 374 \\
 &  & Attribution computation & 248.1 $\pm$ 25.3 ($\sim$248\,s) & 19393 $\pm$ 1020 & 30485 $\pm$ 1285 \\
 &  & \textbf{Total} & \textbf{1494.5 $\pm$ 59.1 ($\sim$25\,min)} & \textbf{19393 $\pm$ 1020} & \textbf{30485 $\pm$ 1285} \\
\midrule
Dyngen & CIG (expanded basis, 2nd order) & Surrogate training & 1246.6 $\pm$ 35.3 ($\sim$21\,min) & 8966 $\pm$ 305 & 12815 $\pm$ 376 \\
 &  & Attribution computation & 494.3 $\pm$ 50.1 ($\sim$494\,s) & 28228 $\pm$ 1511 & 39314 $\pm$ 1774 \\
 &  & \textbf{Total} & \textbf{1740.9 $\pm$ 83.3 ($\sim$29\,min)} & \textbf{28228 $\pm$ 1511} & \textbf{39314 $\pm$ 1774} \\
\midrule
Dyngen & CEG (expanded basis, 2nd order) & Surrogate training & 1417.5 $\pm$ 251.9 ($\sim$24\,min) & 8966 $\pm$ 305 & 13550 $\pm$ 1572 \\
 &  & Attribution computation & 66089.2 $\pm$ 2021.1 ($\sim$18.4\,hr) & 29174 $\pm$ 1532 & 35869 $\pm$ 2056 \\
 &  & \textbf{Total} & \textbf{67506.7 $\pm$ 1971.1 ($\sim$18.8\,hr)} & \textbf{29174 $\pm$ 1532} & \textbf{35869 $\pm$ 2056} \\
\midrule
IFN-$\beta$ & CIG (expanded basis, 2nd order) & Surrogate training & 6465.1 $\pm$ 233.3 ($\sim$1.8\,hr) & 10627 $\pm$ 7 & 14255 $\pm$ 12 \\
 &  & Attribution computation & 7152.1 $\pm$ 14.0 ($\sim$2.0\,hr) & 19112 $\pm$ 1 & 30143 $\pm$ 12 \\
 &  & \textbf{Total} & \textbf{13617.1 $\pm$ 242.2 ($\sim$3.8\,hr)} & \textbf{19112 $\pm$ 1} & \textbf{30143 $\pm$ 12} \\
\bottomrule
\end{tabular}%
}
\end{table}

The pipeline is split into surrogate fitting and attribution, and the stage that dominates depends on the surrogate. For CIG both are modest, and the cost of moving from first- to second-order indirect effects is roughly a doubling of attribution time with a smaller increase in memory. CEG reuses the same surrogate and the same integration algorithm but repeats the integration for each reference baseline, so its attribution stage is two orders of magnitude slower. However, this is purely a time cost and its peak memory is essentially that of CIG. Peak memory is set by the attribution stage rather than by surrogate training, and grows mainly with the order of indirect effects. The basis expansion shifts the burden entirely onto training, raising both fitting time and training memory by an order of magnitude while increasing attribution time only moderately. It is a one-off amortizable surcharge for a non-linear conditional model and not a recurring per-explanation cost.

\section{Details of the synthetic causal attribution benchmark} \label{appendix:causal_attributions_extra} \label{appendix:root_node_intuition}

\begin{figure}[h]
    \centering
    \includegraphics[width=0.55\textwidth]{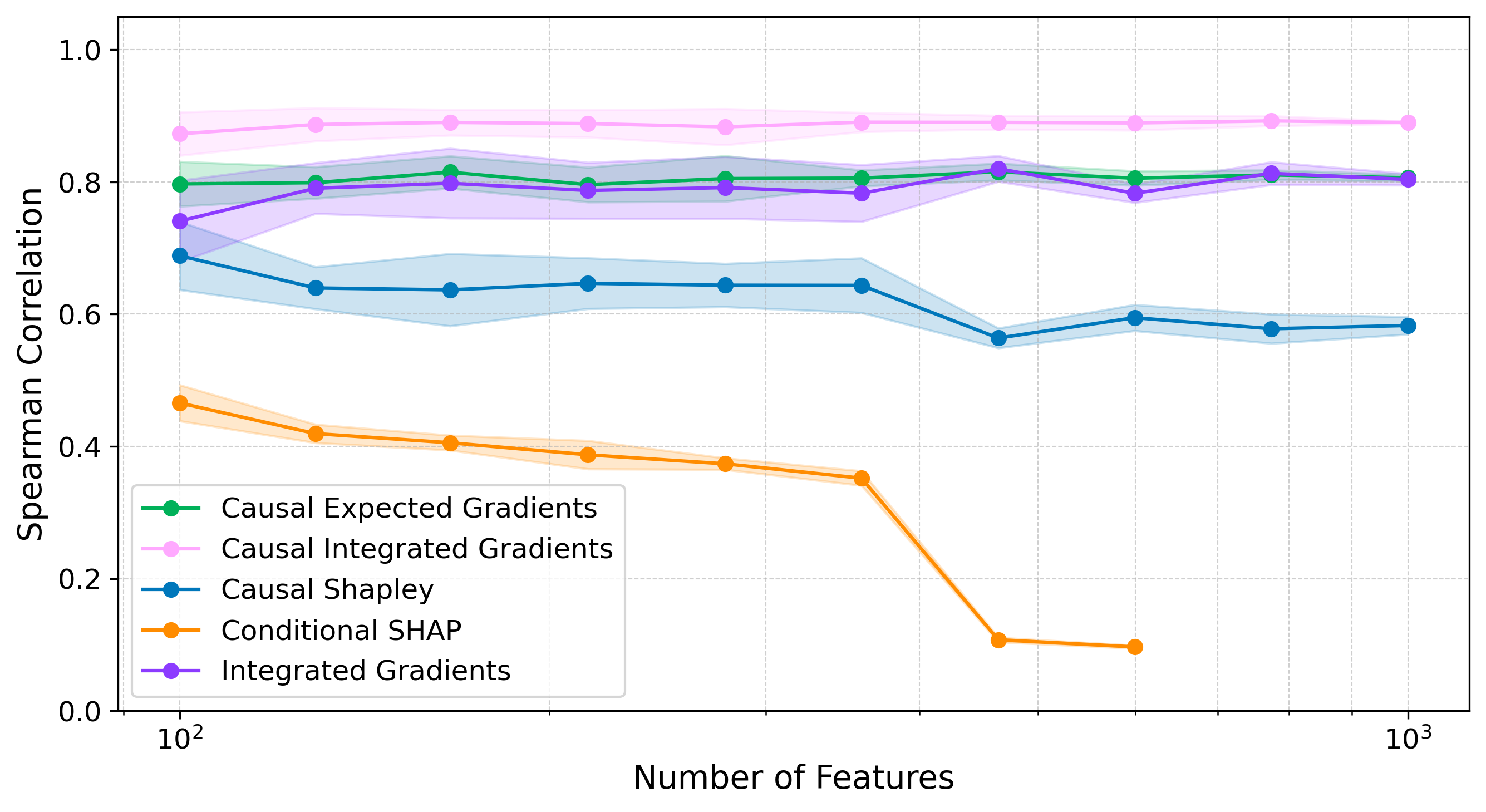}
    \caption{Spearman correlation with the CATE averaged over all nodes, as a function of the number of features. Shaded areas denote the standard deviation over 5 runs.}
    \label{fig:causal_scaling_spearman_all_nodes}
\end{figure}

\paragraph{Experimental setup.}
For each feature dimensionality and each of the 5 independent runs, a random causal graph of maximum depth 2 is generated, with expected node degree $k=5$. Root nodes (those without parents) receive no structural assignment function; their value is pure exogenous noise. For every non-root node $j$, a mechanism type is drawn independently and uniformly at random from three options — linear, tanh, and interaction, each with probability $1/3$:
\begin{itemize}
    \item \textbf{Linear:} $c_j(x_{\mathrm{pa}(j)}) = \sum_{i\in\mathrm{pa}(j)} w_i x_i$, with weights $w_i \sim \mathrm{Uniform}(-1,1)$ drawn i.i.d.\ per parent.
    \item \textbf{Tanh:} $c_j(x_{\mathrm{pa}(j)}) = \tanh\!\left(\sum_{i\in\mathrm{pa}(j)} w_i x_i\right)$, with the same weight distribution as the linear case.
    \item \textbf{Interaction:} $c_j(x_1,x_2) = w_1 x_1 + w_2 x_2 + w_1 w_2 x_1 x_2$, with $w_1,w_2\sim\mathrm{Uniform}(-1,1)$. Only the first two parents are used regardless of how many parents the node actually has in the DAG; any additional parents receive a true structural weight of exactly zero — this only matters for nodes with more than two parents. If a node has fewer than two parents, this case degenerates to the linear assignment above.
\end{itemize}
Input features are sampled according to these assignments with additive exogenous Gaussian noise. Targets are generated as a random linear combination of a randomly chosen 20\% subset of the input features, with added Gaussian noise. The reported metric is the sample-wise Spearman correlation (averaged over 100 samples) between the local attributions and the individual conditional average treatment effects (CATEs), computed either over all input features (Figure~\ref{fig:causal_scaling_spearman_all_nodes}) or over root nodes only (left panel of Figure~\ref{fig:causal_scaling}).

Conditional SHAP is implemented assuming a multivariate Gaussian distribution over the inputs, which allows the conditional distribution of any feature subset to be computed in closed form \citep{aas2021explaining}. Background samples are drawn from these conditional distributions. Causal Shapley values directly use the available structural causal model to sample from the interventional distribution implied by the do-operator, avoiding the need for distributional approximations. Specifically, the expectation
\begin{equation}
E[F \mid do(X_S = x_S)],
\end{equation}
which appears in the computation of causal Shapley values, is evaluated by running the structural causal model (including exogenous noise) with features $S$ fixed to $x_S$ and all remaining features drawn from the background dataset. The model output is averaged over these background samples.

Both methods are implemented using the \texttt{PermutationExplainer} from the \texttt{shap} package \citep{lundberg2017unified} with 10 permutations and 50 background samples. For experiments with 464 or more features, where runtime became prohibitive, these were reduced to 4 permutations and 10 background samples. As a result, the exponential scaling of exact Shapley-value computation is not fully visible in the runtime plot (right panel of Figure~\ref{fig:causal_scaling}): maintaining comparable approximation quality at higher feature counts would require \emph{increasing}, not decreasing, these parameters, so the measured curve underestimates the true asymptotic gap.

Both causal integrated gradients and integrated gradients use 100 steps for the Riemann sum and a zero baseline. Causal expected gradients uses 10 random baselines and 10 steps for each baseline.

\paragraph{Root-node intuition.}
The intuition behind the root-node results is best illustrated with a simple example. Consider a model $F$ with two equally contributing features $x_1$ and $x_2$, where $x_1$ fully causes $x_2$ (i.e.\ $x_1 \rightarrow x_2 \rightarrow F$, $x_1 \rightarrow F$):
\begin{itemize}
    \item \textbf{Conditional SHAP} gives $R_1 = R_2$, since it does not account for the direction of the dependency and therefore does not preferentially attribute importance to the root node $x_1$.
    \item \textbf{Causal Shapley values} give $R_1 > R_2$, since they take into account the direction of the dependency but still attribute some importance to $x_2$, because they consider a situation where $x_1$ is not set to a value, in which case $x_2$ does have an effect on $F$.
    \item \textbf{Causal integrated gradients} gives $R_2 = 0$, correctly identifying that $x_2$ contributes no independent effect beyond what is already captured by $x_1$.
\end{itemize}

\paragraph{All-node results.}
Figure~\ref{fig:causal_scaling_spearman_all_nodes} shows attribution quality when evaluated over all nodes.

\begin{figure}[H]
    \centering
    \begin{subfigure}[c]{0.48\textwidth}
        \centering
        \includegraphics[width=\textwidth]{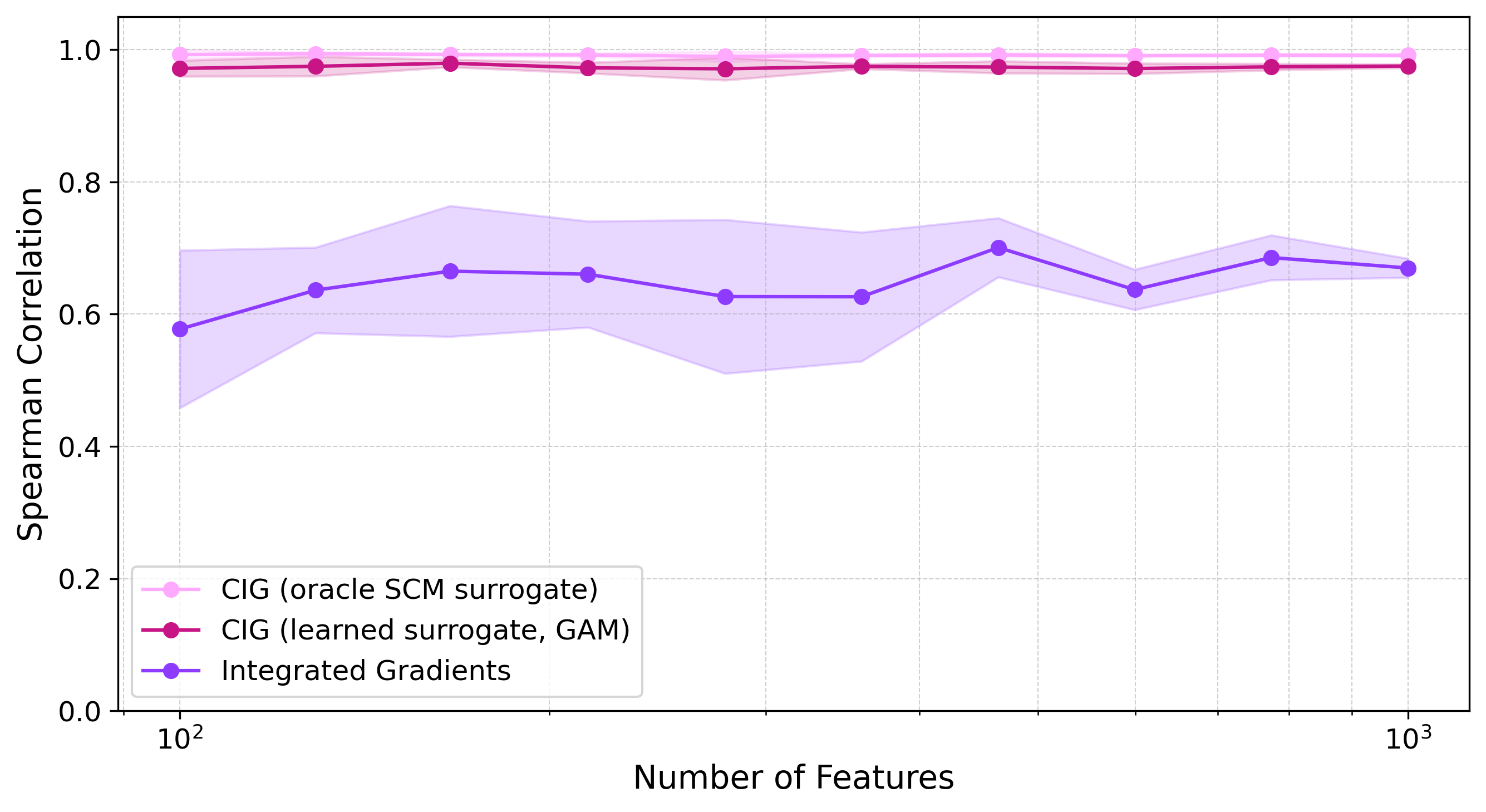}
        \caption{Root-cause (eldest-ancestor) features.}
        \label{fig:surrogate_comparison_eldest_ancestors}
    \end{subfigure}
    \hfill
    \begin{subfigure}[c]{0.48\textwidth}
        \centering
        \includegraphics[width=\textwidth]{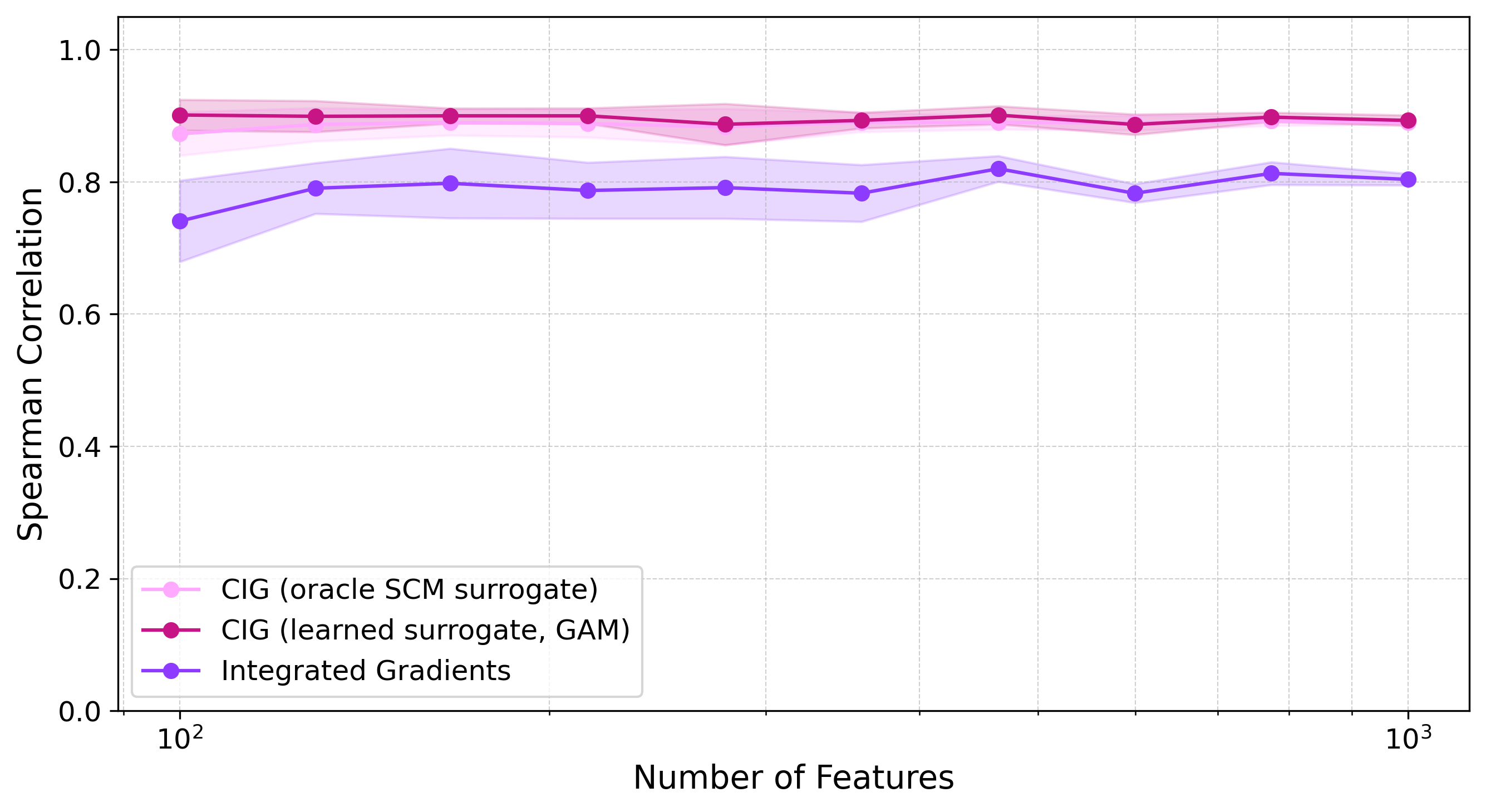}
        \caption{All nodes.}
        \label{fig:surrogate_comparison_all_nodes}
    \end{subfigure}
    \caption{Spearman correlation with ground-truth CATE as a function of the number of features: CIG using the SCM's true mechanisms directly (oracle SCM surrogate) vs.\ CIG using our learned, expanded-basis surrogate (labeled ``GAM'' in the legend) vs.\ graph-agnostic integrated gradients. Shaded areas denote the standard deviation over 5 seeds.}
    \label{fig:surrogate_comparison}
\end{figure}

\paragraph{Learned surrogate vs.\ true mechanisms.}
The main benchmark uses the SCM's true structural mechanisms directly when computing CIG's attributions, isolating attribution quality from surrogate error. To check the resulting gap between this idealized setting and practical use, we rerun the benchmark using our learned surrogate — the expanded-basis variant of the sparse masked surrogate from Section~\ref{sec:surrogate} — in place of the true mechanisms, given only the causal graph's topology as prior knowledge; mechanisms themselves are learned from data, exactly as in our biological experiments. Figure~\ref{fig:surrogate_comparison} reports Spearman correlation with ground-truth CATE across the full 100--1000 feature sweep, for both root-cause (eldest-ancestor) and all-node features, alongside graph-agnostic integrated gradients for reference.

The learned surrogate tracks the true-mechanism numbers closely across the entire sweep, and both remain far above integrated gradients, which ignores the graph entirely. The gap motivating this comparison is therefore negligible in practice and does not materially change attribution values or rankings relative to the idealized true-mechanism setting.

\section{Details of the Dyngen experiment} \label{appendix:gene_recovery_extra_dyngen}

\begin{figure}[h]
\centering
\begin{minipage}[b]{0.40\linewidth}
\centering
\captionof{table}{Summary statistics of the generated Dyngen linear dataset.}
\label{tab:dyngen_stats}
\begin{tabular}{lr}
\toprule
\textbf{Statistic} & \textbf{Value} \\
\midrule
Trajectory topology & Linear \\
Cells & 10{,}000 \\
Transcription factors & 48 \\
Target genes & 2{,}000 \\
Housekeeping genes & 500 \\
Total genes & 2{,}548 \\
Total simulation time & 1{,}000 \\
Simulations & 100 \\
\bottomrule
\end{tabular}
\end{minipage}\hfill
\begin{minipage}[b]{0.56\linewidth}
\centering
\includegraphics[width=\linewidth,trim={28 105 0 100},clip]{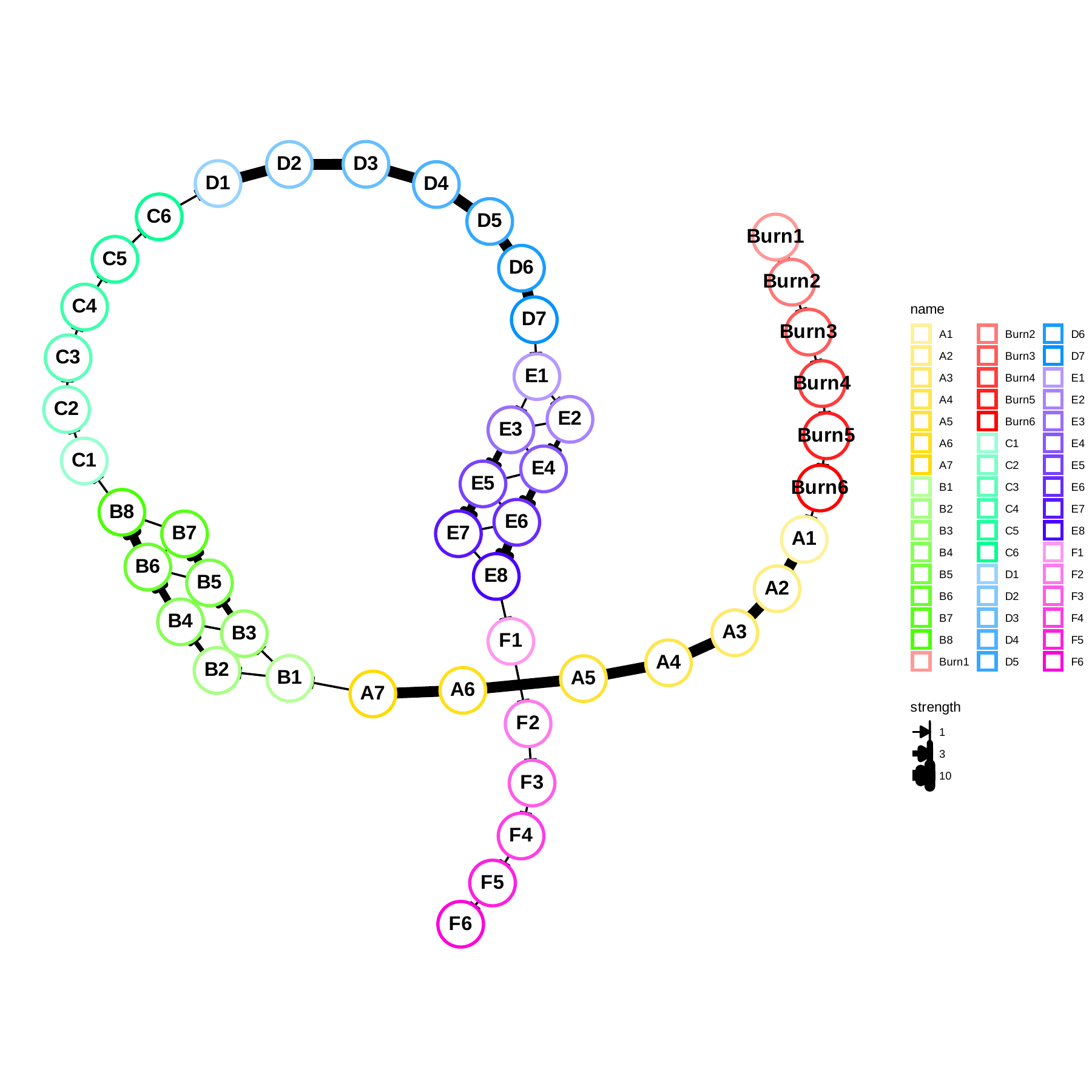}
\captionof{figure}{Backbone module network for the generated Dyngen linear dataset.}
\label{fig:backbone_modulenet}
\end{minipage}
\end{figure}

\begin{table}[H]
\centering
\caption{Dyngen regression datasets across the five simulation
seeds. Cells and genes are reported before and after quality-control
filtering; the train/validation/test split is a nested 70/30 split of the
filtered cells (49\%/21\%/30\%).}
\label{tab:dyngen-regression-datasets}
\begin{tabular}{lrrrrrrr}
\toprule
 & \multicolumn{2}{c}{Cells} & \multicolumn{2}{c}{Genes} & \multicolumn{3}{c}{Filtered cells per split} \\
\cmidrule(lr){2-3}\cmidrule(lr){4-5}\cmidrule(lr){6-8}
Seed & Raw & Filtered & Raw & Filtered & Train & Val. & Test \\
\midrule
42   & 10{,}000 & 9{,}790  & 2{,}548 & 2{,}378 & 4{,}797 & 2{,}056 & 2{,}937 \\
420  & 10{,}000 & 8{,}380  & 2{,}548 & 2{,}445 & 4{,}106 & 1{,}760 & 2{,}514 \\
100  & 10{,}000 & 8{,}871  & 2{,}548 & 2{,}320 & 4{,}346 & 1{,}863 & 2{,}662 \\
1000 & 10{,}000 & 10{,}000 & 2{,}548 & 2{,}388 & 4{,}900 & 2{,}100 & 3{,}000 \\
2000 & 10{,}000 & 9{,}876  & 2{,}548 & 2{,}494 & 4{,}839 & 2{,}074 & 2{,}963 \\
\bottomrule
\end{tabular}
\end{table}
\begin{table}[H]
\centering
\begin{minipage}[t]{0.42\textwidth}
\centering
\captionof{table}{Hyperparameters of the surrogate model used for the Dyngen dataset.}
\label{tab:surrogate_training_details_dyngen}
\begin{tabular}{lr}
\toprule
\textbf{Hyperparameter} & \textbf{Value} \\
\midrule
Number of knots & 75 \\
Knot power & 3 \\
Polynomial degree & 3 \\
Fourier features & 10 \\
Training epochs & 2000 \\
Weight decay & $1 \times 10^{-5}$ \\
\bottomrule
\end{tabular}
\end{minipage}\hfill
\begin{minipage}[t]{0.55\textwidth}
\centering
\captionof{table}{Hyperparameters for data preprocessing and for the neural network model used to predict simulation time.}
\label{tab:mlp_hyperparameters}
\begin{tabular}{lr}
\toprule
\textbf{Hyperparameter} & \textbf{Value} \\
\midrule
Min genes expressed per cell & 200 \\
Min cells expressing a gene & 100 \\
Activation function & ReLU \\
Hidden layer sizes & (50, 25, 10) \\
Maximum iterations & 500 \\
Batch size & 512 \\
Solver & Adam \\
Alpha (L2 penalty) & $1 \times 10^{-4}$ \\
Learning rate & $1 \times 10^{-3}$ \\
\bottomrule
\end{tabular}
\end{minipage}
\end{table}

\begin{table}[H]
\centering
\caption{Feature attribution methods evaluated in the simulated single-cell experiment.}
\label{tab:attribution_methods_summary}
\small
\setlength{\tabcolsep}{4pt}
\begin{tabularx}{\linewidth}{lll>{\raggedright\arraybackslash}X}
\toprule
\textbf{Method} & \textbf{Used} & \textbf{Baseline} & \textbf{Notes} \\
\midrule
DeepSHAP \citep{lundberg2017unified} & $\checkmark$ & Random set & \texttt{captum} implementation \\
GradientSHAP \citep{lundberg2017unified} & $\checkmark$ & Random set & \texttt{captum} implementation \\
Integrated gradients \citep{sundararajan2017axiomatic} & $\checkmark$ & Zero & \texttt{captum} implementation \\
DeepLIFT \citep{shrikumar_learning_2019} & $\checkmark$ & Zero & \texttt{captum} implementation \\
Saliency \citep{simonyan_deep_2014} & $\checkmark$ & $\times$ & \texttt{captum} implementation \\
Guided backpropagation \citep{springenberg_striving_2015} & $\checkmark$ & $\times$ & \texttt{captum} implementation \\
Partition explainer \citep{lundberg2017unified} & $\checkmark$ & Correlation & \texttt{shap} implementation \\
CIG (first-order) & $\checkmark$ & Zero & Linear/expanded basis (Table~\ref{tab:surrogate_training_details_dyngen}) \\
CIG (second-order) & $\checkmark$ & Zero & Linear/expanded basis (Table~\ref{tab:surrogate_training_details_dyngen}) \\
CEG extension (second-order) & $\checkmark$ & Random set & Expanded basis (Table~\ref{tab:surrogate_training_details_dyngen}) \\
Conditional SHAP \citep{aas2021explaining} & $\times$ & -- & Computationally infeasible \\
Causal Shapley values \citep{heskes2020causal} & $\times$ & -- & Computationally infeasible \\
\bottomrule
\end{tabularx}
\end{table}
\begin{table}[H]
\centering
\caption{Gene relevance scoring scheme based on the simulated gene regulatory network (GRN). The DCG at cutoff $K$ is $\mathrm{DCG}@K = \sum_{i=1}^{K} \mathrm{rel}_i / \log_2(i+1)$, where $\mathrm{rel}_i$ is the score of the gene at rank $i$.}
\label{tab:gene_relevance_scores}
\begin{tabular}{p{4.8cm}cp{6.6cm}}
\toprule
\textbf{Gene Category} & \textbf{Score} & \textbf{Description} \\
\midrule
Backbone transcription factors (TFs) & +1.00 & TFs defining main differentiation trajectory in GRN \\
Direct targets of backbone TFs & +0.25 & Genes regulated by one backbone TF \\
Joint targets with opposing interaction signs & +0.25 & Genes jointly regulated by two TFs with opposite regulatory signs \\
Burn-in phase TFs & -0.25 & TFs active only during burn-in phase \\
Targets of burn-in TFs & -0.25 & Genes regulated by burn-in phase TFs \\
Housekeeping genes & -0.25 & Non-informative constitutively expressed genes \\
Other genes & 0.00 & Genes not belonging to the categories above \\
\bottomrule
\end{tabular}
\end{table}

\section{Details of the \texorpdfstring{IFN-$\beta$}{IFN-beta} experiment} \label{appendix:gene_recovery_extra_ifnb}
\begin{table}[H]
\centering
\begin{minipage}[t]{0.42\textwidth}
\centering
\captionof{table}{Hyperparameters of the surrogate model used for the IFN-$\beta$ dataset.}
\label{tab:surrogate_training_details_ifnb}
\begin{tabular}{lr}
\toprule
\textbf{Hyperparameter} & \textbf{Value} \\
\midrule
Number of knots & 50 \\
Knot power & 3 \\
Polynomial degree & 5 \\
Fourier features & 5 \\
Training epochs & 1000 \\
Weight decay & $1 \times 10^{-5}$ \\
\bottomrule
\end{tabular}
\end{minipage}\hfill
\begin{minipage}[t]{0.55\textwidth}
\centering
\captionof{table}{Hyperparameters for data preprocessing and for the neural network model used to predict IFN-stimulated versus control cells.}
\label{tab:perturbation_experiment_details}
\begin{tabular}{lr}
\toprule
\textbf{Hyperparameter} & \textbf{Value} \\
\midrule
Min genes expressed per cell & 200 \\
Min counts per gene & 250 \\
Genes before filtering & 15{,}706 \\
Cells before filtering & 24{,}673 \\
Genes after filtering & 6{,}941 \\
Cells after filtering & 24{,}562 \\
Input dimension & 6{,}941 \\
Activation function & ReLU \\
Hidden layer sizes & (25, ) \\
Maximum iterations & 10 \\
Batch size & 512 \\
Solver & Adam \\
Alpha (L2 penalty) & $1 \times 10^{-4}$ \\
Learning rate & $1 \times 10^{-3}$ \\
Train/validation/test sizes & 17{,}448/2{,}024/5{,}090 \\
\bottomrule
\end{tabular}
\end{minipage}
\end{table}

\paragraph{OmniPath prior graph construction.}
Since no causal graph is available for this dataset, we leverage prior knowledge from OmniPath \citep{turei_integrated_2021}. After restricting the graph to genes present in the dataset and to the interactions annotated as transcriptional, the resulting network contains 6,941 genes and 22,887 edges. Each edge was then assigned a confidence score as a weighted mean of four evidence channels: curation effort, number of references, number of primary sources, and consensus direction. Duplicate source–target records were collapsed into a single edge by taking the maximum of each evidence channel, reducing the edges to 22,624 aggregated interactions. Edges without supporting evidence in any channel were removed, and the edges with the top 75\% of weights were kept, leaving 12,136 edges. Because a topological ordering of the features is required, we converted the cyclic graph to a DAG by approximate minimum feedback arc set removal. Specifically, we first computed a linear vertex ordering using the weighted greedy heuristic of \citet{EADES1993319}, removed arcs inconsistent with this ordering, and then reconsidered removed arcs in descending confidence order, reinserting an arc whenever doing so preserves acyclicity. This removed 399 of 12,136 edges (3.3\% of edges and 3.6\% of total confidence mass), yielding an acyclic prior of 11,737 edges over 3,792 connected genes.
\paragraph{SCENIC evaluation graph construction.}

\textbf{Data.} We started from the PBMC scRNA-seq dataset of \citet{kang_multiplexed_2018}, comprising 8 lupus patients, with each patient's cells split into IFN-$\beta$-stimulated and control samples. We used the raw UMI counts throughout the SCENIC steps.

\textbf{Quality control.} We kept cells with more than 200 detected genes, fewer than 4,000 detected genes, and less than 15\% mitochondrial reads. We kept genes with more than 250 total counts across all cells, which matches the SCENIC recommendation of one count in 1\% of cells. The filtered matrix contained 24,562 cells $\times$ 6,941 genes. We restricted the data to six training patients (1244, 1016, 1039, 1256, 1488, and 107), comprising 17,448 cells, leaving patients 1015 and 101 out.

\textbf{Multi-seed SCENIC.} GRNBoost2 is stochastic, so we ran pySCENIC ten times with seeds \{0, 1, 3, 4, 5, 20, 42, 420, 1000, 2000\}. Each run had three steps:
\begin{enumerate}
    \item \textit{GRN inference}: GRNBoost2 on the raw count matrix, with the hg38 list of human TFs as candidate regulators. Output: TF--target adjacencies weighted by importance.
    \item \textit{Module creation and cisTarget pruning}: co-expression modules were built with the default settings, so TF--target correlations use only cells where both genes are detected. Modules were then pruned by motif enrichment. We used two hg38 RefSeq-r80 mc9nr gene-based ranking databases (10 kb up/downstream of the TSS, and 500 bp up/100 bp down) and the motif-to-TF annotation motifs-v9-nr.hgnc-m0.001-o0.0.
    \item \textit{Regulon filtering}: we kept enriched motifs with NES greater than 3.0 and motif-similarity $q$-value less than 0.001. Surviving motifs were merged into one regulon per TF and context, which takes the union of their targets.
\end{enumerate}

\textbf{Consensus network across seeds.} Each seed's regulons were flattened into directed TF-to-target edges. When a TF had several contexts, an edge kept its largest GRNBoost2 weight. Seeds agreed moderately: pairwise Jaccard similarity of the edge sets was 0.58 to 0.60. We chose the recurrence threshold from the edge-recurrence curve (Table~\ref{tab:edges_recurrence}). An edge entered the consensus if it appeared in at least 8 of 10 seeds, and its weight is the median over the seeds that contained it. The training-patient data give 23,451 edges from 155 TFs to 5,442 targets.

\begin{table}[ht]
\centering
\caption{Edges kept at each recurrence threshold (full data).}
\label{tab:edges_recurrence}
\begin{tabular}{lrrrrrr}
\toprule
\textbf{Min.\ seeds} & \textbf{1} & \textbf{2} & \textbf{4} & \textbf{6} & \textbf{8} & \textbf{10} \\
\midrule
\textbf{Edges} & 74{,}426 & 58{,}082 & 42{,}448 & 33{,}172 & 26{,}379 & 19{,}055 \\
\textbf{TFs}   & 252    & 229    & 193    & 174    & 156    & 131    \\
\bottomrule
\end{tabular}
\end{table}

\textbf{Consensus regulon activity (AUCell).} For each seed we scored regulon activity per cell with pyscenic aucell, using an AUC threshold of 0.10. That means the top 10\% of each cell's ranked genes; the median cell detects about 500 genes. A regulon was kept if (i) its TF is in the consensus network and (ii) AUCell scored it in at least 8 of 10 seeds. Its activity is the mean AUC over the seeds that scored it, and a seed that did not score it counts as missing rather than zero. The result is 155 regulons.

\textbf{Extraction of stimulation-responsive SCENIC regulons.} For each cell type, SCENIC regulon activities, quantified using AUCell on the training replicates, were averaged within each replicate and contrasted between the stimulated and control groups. Regulons significant according to a one-sample $t$-test across replicates (BH-FDR $< 0.05$) were retained, and the 50 regulons with the largest $|\Delta\mathrm{AUC}|$ were selected.

\section{Additional \texorpdfstring{IFN-$\beta$}{IFN-beta} results}
\subsection{Attribution rankings by hop distance from the roots} \label{appendix:hop_distance}
 The rankings are reproducible across training seeds: the per-gene spread of the percentile rank across seeds stays well below what independent seed-to-seed rankings would produce (Tables~\ref{tab:sd_median_omnipath} and~\ref{tab:sd_median_scenic}).
\begin{figure}[H]
    \centering
    \includegraphics[width=\linewidth]{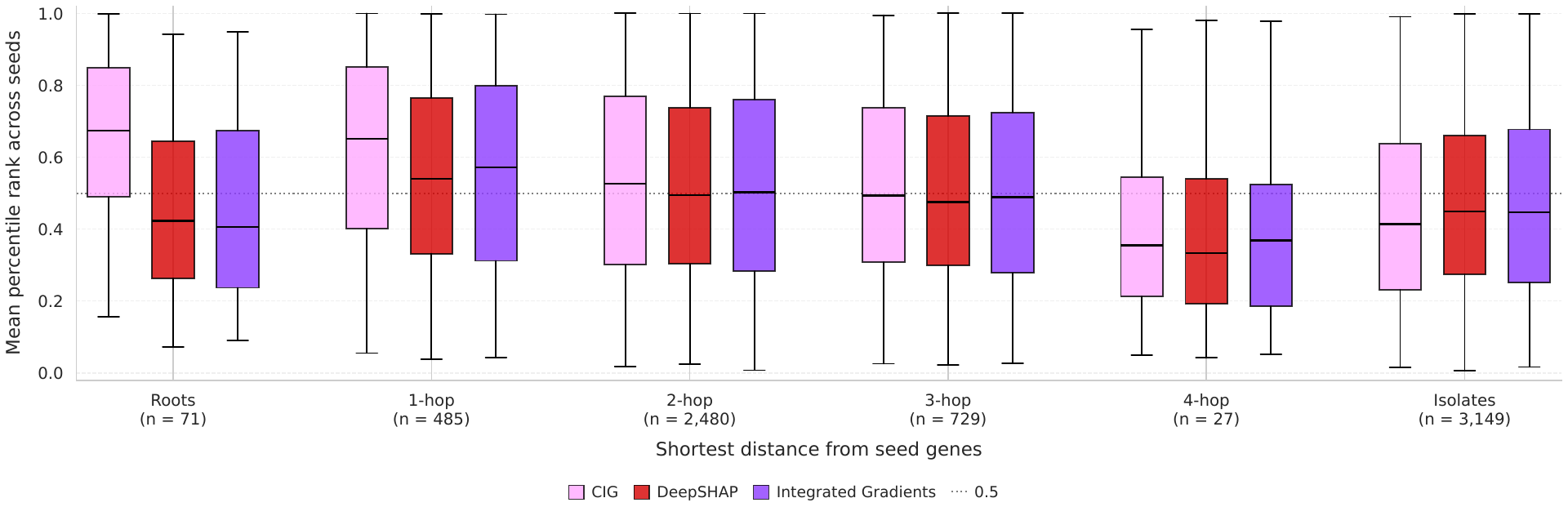}
    \caption{Distribution of attribution ranks for genes grouped by their hop distance from the roots of the OmniPath prior graph. Each boxplot represents the distribution of ranks for genes within a specific hop-distance group, with the median rank indicated by a horizontal line within each box.}
    \label{fig:ifnb_hop_distance_omnipath}
\end{figure}

\begin{table}[H]
\centering
\caption{Median standard deviations of percentile rank across seeds (OmniPath); chance $=0.278$, perfectly stable $=0.0$.}
\label{tab:sd_median_omnipath}
\begin{tabular}{lrrrrrr}
\toprule
 & Roots & 1-hop & 2-hop & 3-hop & 4-hop & Isolates \\
\midrule
CIG & 0.1220 & 0.1085 & 0.1126 & 0.1197 & 0.1445 & 0.0959 \\
DeepSHAP & 0.1312 & 0.1194 & 0.1292 & 0.1335 & 0.1598 & 0.1428 \\
Integrated gradients & 0.0980 & 0.0891 & 0.0956 & 0.0987 & 0.1065 & 0.1017 \\
\bottomrule
\end{tabular}
\end{table}

\begin{table}[H]
\centering
\caption{Median standard deviations of percentile rank across seeds (SCENIC); chance $=0.278$, perfectly stable $=0.0$.}
\label{tab:sd_median_scenic}
\begin{tabular}{lrrrrrrr}
\toprule
 & Roots & 1-hop & 2-hop & 3-hop & 4-hop & 5-hop & Isolates \\
\midrule
CIG & 0.1057 & 0.1026 & 0.1007 & 0.1041 & 0.1145 & 0.0872 & 0.1095 \\
DeepSHAP & 0.1940 & 0.1326 & 0.1242 & 0.1295 & 0.1502 & 0.1323 & 0.1535 \\
Integrated gradients & 0.1309 & 0.0974 & 0.0929 & 0.0969 & 0.1085 & 0.0535 & 0.1026 \\
\bottomrule
\end{tabular}
\end{table}

\subsection{\texorpdfstring{Top-$k$}{Top-k} enrichment of stimulation-responsive SCENIC regulons} \label{appendix:topk_enrichment}
\begin{figure}[H]
    \centering
    \includegraphics[width=\linewidth]{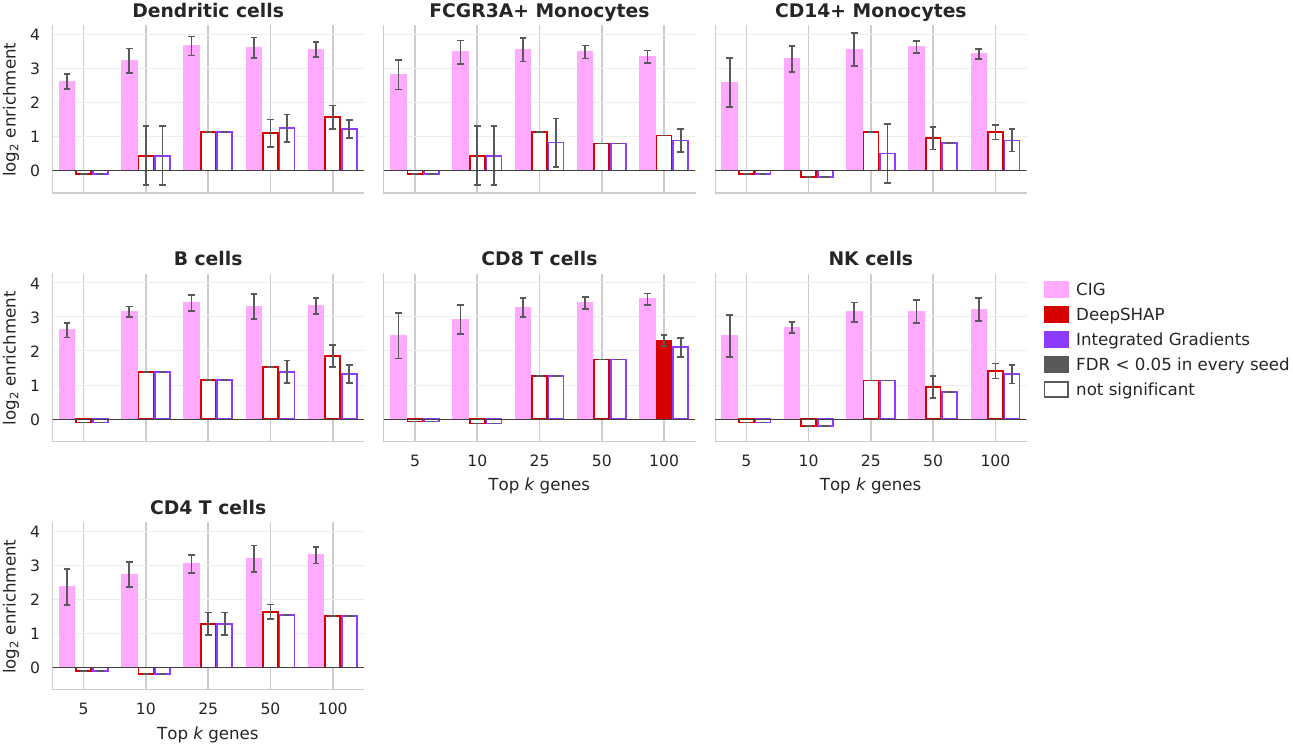}
    \caption{Top-$k$ enrichment of stimulation-responsive SCENIC regulons. The $z$-score of the hypergeometric test statistic is shown for each method, cell type, and top-$k$ cutoff, averaged across five random seeds. Higher values indicate that responsive TFs are over-represented among the top-$k$ genes. Error bars denote the standard deviation across seeds. $P$-values were BH-corrected across cell types and $k$. Only significant enrichment bars are shaded.
}
    \label{fig:top_tf_enrichment}
\end{figure}

\subsection{Recovery of pseudobulk DEGs} \label{appendix:bulk_deg_recovery}
As an orthogonal readout, we asked how well each attribution ranking recovers the per-cell-type pseudobulk DEGs. Within each cell type, genes were ranked by their \emph{absolute attribution difference} $\delta_g = \lvert \bar{R}_g^{\,\text{stim}} - \bar{R}_g^{\,\text{ctrl}} \rvert$, where $\bar{R}_g^{\,\text{stim}}$ and $\bar{R}_g^{\,\text{ctrl}}$ denote the attribution of gene $g$ averaged over the stimulated and over the control cells of that cell type. Then, for each cell type, the rankings were scored against that cell type's DESeq2 \citep{deseq2} DEG set using precision, recall, F1, normalized recovery AUC, and normalized discounted cumulative gain (NDCG; the DCG divided by that of the ideal ranking) graded by the absolute Wald statistic, at several fixed cutoffs and at a cutoff equal to the number of DEGs. The normalized recovery AUC is the area under the curve of recovered DEGs against rank depth, rescaled so that a random ranking scores $0$ and the optimal ranking scores $1$. At $k = n_{\text{DEGs}}$, precision, recall, and F1 coincide by construction, so only F1 is reported at that cutoff. We also measured the Spearman correlation between $\delta_g$ and the absolute DESeq2 Wald statistic over the entire gene panel. Every quantity was computed independently for all five seeds and is reported as mean $\pm$ s.d. over them. Megakaryocytes contributed no significant DEGs and are therefore absent from the recovery metrics.

The whole-panel correlation is stable seed-to-seed. No method tracks the DE statistic closely (Table~\ref{tab:degs_spearman}), so none of these methods can be considered equivalent to a differential expression test. CIG is the least DE-correlated of the three in all seven cell types. Over the full DEG set, scored at a cutoff equal to the number of DEGs (Table~\ref{tab:degs_scores}), CIG has the lowest F1 in six of the seven cell types and the lowest recovery AUC in six of the seven; the exceptions are dendritic cells, where CIG and integrated gradients are tied on F1, and CD4 T cells, where CIG's recovery AUC exceeds that of DeepSHAP. CIG's NDCG is below that of both other methods in every cell type. Apart from dendritic cells and the CIG--DeepSHAP recovery-AUC comparison in CD4 T cells, these gaps are larger than the seed-to-seed spread.

The methods separate most sharply at the top of the ranking, and in the same direction in every cell type. Among their first ten genes, DeepSHAP and integrated gradients reach $94$--$100\%$ precision and an NDCG of $64$--$91$, against $34$--$52\%$ and $22$--$32$ for CIG (Table~\ref{tab:degs_cutoff10}). The gap then narrows steadily with depth (Tables~\ref{tab:degs_cutoff25}--\ref{tab:degs_cutoff250}). CIG's precision is also non-monotone in $k$. It is lower at $k = 10$ than at any cutoff up to $k = 100$, peaks at $k = 100$ in every cell type except NK cells (where it peaks at $k = 50$), and only then declines, whereas DeepSHAP and integrated gradients are at or near their maximum at $k = 10$ and decline thereafter. The genes CIG places in its first few dozen positions are therefore less likely to be DEGs than the ones it ranks immediately below them.

Taken together, these results are consistent with the following behavior. DeepSHAP and integrated gradients rank the downstream transcriptional response genes first, which lifts their precision and NDCG at shallow cutoffs and gives them a marginally closer alignment with the DE statistic overall. CIG instead prioritizes the upstream regulators driving that response, genes that are rarely strong DEGs themselves.

\begin{table}[H]
\centering
\caption{Whole-panel agreement between attribution and differential expression. For each cell type and method, the Spearman correlation between the absolute attribution difference $\delta_g$ (stimulated vs.\ control) and the absolute DESeq2 Wald statistic is computed over the full measured gene panel. Values are the mean and standard deviation across five random seeds.}
\label{tab:degs_spearman}
\begin{tabular}{llrr}
\toprule
Cell type & Method & Mean Spearman $\rho$ & S.d. across seeds \\
\midrule
B cells & CIG & 0.3098 & 0.0072 \\
B cells & DeepSHAP & 0.3724 & 0.0015 \\
B cells & Integrated gradients & 0.3465 & 0.0039 \\
CD14$^+$ monocytes & CIG & 0.3530 & 0.0343 \\
CD14$^+$ monocytes & DeepSHAP & 0.3945 & 0.0215 \\
CD14$^+$ monocytes & Integrated gradients & 0.4119 & 0.0395 \\
CD4 T cells & CIG & 0.2678 & 0.0069 \\
CD4 T cells & DeepSHAP & 0.2926 & 0.0076\\
CD4 T cells & Integrated gradients & 0.3273 & 0.0048\\
CD8 T cells & CIG & 0.2253 & 0.0113 \\
CD8 T cells & DeepSHAP & 0.2800 & 0.0079\\
CD8 T cells & Integrated gradients & 0.2429 & 0.0070\\
Dendritic cells & CIG & 0.1524 & 0.0106\\
Dendritic cells & DeepSHAP & 0.1571 & 0.0080\\
Dendritic cells & Integrated gradients & 0.1537 & 0.0041 \\
FCGR3A$^+$ monocytes & CIG & 0.3918 & 0.0275\\
FCGR3A$^+$ monocytes & DeepSHAP & 0.4358 & 0.0167 \\
FCGR3A$^+$ monocytes & Integrated gradients & 0.4378 & 0.0279 \\
NK cells & CIG & 0.1548 & 0.0065 \\
NK cells & DeepSHAP & 0.1566 & 0.0065\\
NK cells & Integrated gradients & 0.1658 & 0.0033 \\
\bottomrule
\end{tabular}
\end{table}

\begin{table}[H]
\centering
\caption{Recovery of pseudobulk DEGs at a cutoff equal to the number of DEGs ($k = n_{\text{DEGs}}$). For each cell type, genes are ranked by the absolute attribution difference $\delta_g$ and scored against that cell type's DESeq2 DEG set. The recovery AUC is normalized so that a random ranking scores $0$ and the optimal ranking scores $1$; NDCG is graded by the absolute Wald statistic. At this cutoff, precision, recall, and F1 coincide by construction, so only F1 is reported. All values are percentages, given as mean $\pm$ s.d. across five random seeds. Megakaryocytes are omitted as they contributed no significant DEGs.}
\label{tab:degs_scores}
\begin{tabular}{lrrr}
\toprule
Cell type & CIG & DeepSHAP & Integrated gradients \\
\midrule
\multicolumn{4}{l}{\textit{F1-score} at $k = n_{\text{DEGs}}$} \\
\midrule
B cells & $47.12 \pm 0.48$ & $57.56 \pm 0.44$ & $54.78 \pm 0.40$ \\
CD14$^+$ monocytes & $43.70 \pm 1.56$ & $46.41 \pm 0.83$ & $46.78 \pm 1.72$ \\
CD4 T cells & $35.56 \pm 1.07$ & $38.10 \pm 1.06$ & $39.39 \pm 1.06$ \\
CD8 T cells & $49.20 \pm 1.90$ & $67.15 \pm 1.55$ & $65.55 \pm 0.80$ \\
Dendritic cells & $35.55 \pm 0.83$ & $36.19 \pm 0.93$ & $35.52 \pm 0.78$ \\
FCGR3A$^+$ monocytes & $48.38 \pm 2.90$ & $55.04 \pm 0.99$ & $53.76 \pm 2.63$ \\
NK cells & $34.12 \pm 1.33$ & $39.88 \pm 0.93$ & $38.70 \pm 1.16$ \\
\midrule
\multicolumn{4}{l}{\textit{Normalized recovery AUC}} \\
\midrule
B cells & $74.45 \pm 1.41$ & $79.57 \pm 1.35$ & $79.78 \pm 0.59$ \\
CD14$^+$ monocytes & $30.18 \pm 2.23$ & $33.19 \pm 1.72$ & $35.05 \pm 2.93$ \\
CD4 T cells & $44.18 \pm 3.56$ & $41.47 \pm 2.68$ & $49.85 \pm 3.12$ \\
CD8 T cells & $89.53 \pm 1.35$ & $93.90 \pm 1.66$ & $93.40 \pm 2.17$ \\
Dendritic cells & $21.55 \pm 1.76$ & $21.79 \pm 1.01$ & $21.58 \pm 0.89$ \\
FCGR3A$^+$ monocytes & $63.78 \pm 2.97$ & $67.52 \pm 1.70$ & $68.07 \pm 3.12$ \\
NK cells & $44.05 \pm 2.49$ & $49.29 \pm 1.93$ & $46.90 \pm 1.61$ \\
\midrule
\multicolumn{4}{l}{\textit{NDCG} at $k = n_{\text{DEGs}}$} \\
\midrule
B cells & $69.64 \pm 2.43$ & $89.46 \pm 0.65$ & $87.22 \pm 0.79$ \\
CD14$^+$ monocytes & $73.01 \pm 2.69$ & $79.38 \pm 1.21$ & $79.43 \pm 2.31$ \\
CD4 T cells & $68.77 \pm 1.59$ & $84.35 \pm 0.67$ & $83.27 \pm 0.47$ \\
CD8 T cells & $59.80 \pm 2.83$ & $88.49 \pm 0.95$ & $87.78 \pm 1.10$ \\
Dendritic cells & $69.02 \pm 0.45$ & $76.70 \pm 0.23$ & $75.74 \pm 0.78$ \\
FCGR3A$^+$ monocytes & $70.86 \pm 3.30$ & $83.89 \pm 1.69$ & $81.87 \pm 4.07$ \\
NK cells & $67.25 \pm 2.23$ & $82.89 \pm 1.08$ & $80.73 \pm 1.44$ \\
\bottomrule
\end{tabular}
\end{table}

\begin{table}[H]
\centering
\caption{Recovery of pseudobulk DEGs per cell type at a top-$k$ cutoff of $k = 10$. Genes are ranked by the absolute attribution difference $\delta_g$ and scored against that cell type's DESeq2 DEG set. NDCG is graded by the absolute Wald statistic. All values are percentages, given as mean $\pm$ s.d. across five random seeds. Megakaryocytes are omitted as they contributed no significant DEGs.}
\label{tab:degs_cutoff10}
\small
\begin{tabular}{lccc}
\toprule
Cell type & CIG & DeepSHAP & Integrated gradients \\
\midrule
\multicolumn{4}{l}{\textit{Precision}} \\
\midrule
B cells & $52.00 \pm 13.04$ & $100.00 \pm 0.00$ & $100.00 \pm 0.00$ \\
CD14$^+$ monocytes & $48.00 \pm 8.37$ & $100.00 \pm 0.00$ & $100.00 \pm 0.00$ \\
CD4 T cells & $34.00 \pm 13.42$ & $96.00 \pm 5.48$ & $96.00 \pm 5.48$ \\
CD8 T cells & $40.00 \pm 12.25$ & $100.00 \pm 0.00$ & $100.00 \pm 0.00$ \\
Dendritic cells & $48.00 \pm 13.04$ & $100.00 \pm 0.00$ & $96.00 \pm 5.48$ \\
FCGR3A$^+$ monocytes & $44.00 \pm 16.73$ & $96.00 \pm 5.48$ & $94.00 \pm 8.94$ \\
NK cells & $42.00 \pm 13.04$ & $100.00 \pm 0.00$ & $100.00 \pm 0.00$ \\
\midrule
\multicolumn{4}{l}{\textit{Recall}} \\
\midrule
B cells & $1.09 \pm 0.27$ & $2.11 \pm 0.00$ & $2.11 \pm 0.00$ \\
CD14$^+$ monocytes & $0.24 \pm 0.04$ & $0.51 \pm 0.00$ & $0.51 \pm 0.00$ \\
CD4 T cells & $0.47 \pm 0.19$ & $1.32 \pm 0.08$ & $1.32 \pm 0.08$ \\
CD8 T cells & $2.92 \pm 0.89$ & $7.30 \pm 0.00$ & $7.30 \pm 0.00$ \\
Dendritic cells & $0.30 \pm 0.08$ & $0.63 \pm 0.00$ & $0.61 \pm 0.03$ \\
FCGR3A$^+$ monocytes & $0.54 \pm 0.21$ & $1.18 \pm 0.07$ & $1.16 \pm 0.11$ \\
NK cells & $0.85 \pm 0.26$ & $2.03 \pm 0.00$ & $2.03 \pm 0.00$ \\
\midrule
\multicolumn{4}{l}{\textit{F1-score}} \\
\midrule
B cells & $2.14 \pm 0.54$ & $4.12 \pm 0.00$ & $4.12 \pm 0.00$ \\
CD14$^+$ monocytes & $0.48 \pm 0.08$ & $1.01 \pm 0.00$ & $1.01 \pm 0.00$ \\
CD4 T cells & $0.93 \pm 0.37$ & $2.61 \pm 0.15$ & $2.61 \pm 0.15$ \\
CD8 T cells & $5.44 \pm 1.67$ & $13.61 \pm 0.00$ & $13.61 \pm 0.00$ \\
Dendritic cells & $0.60 \pm 0.16$ & $1.25 \pm 0.00$ & $1.20 \pm 0.07$ \\
FCGR3A$^+$ monocytes & $1.07 \pm 0.41$ & $2.34 \pm 0.13$ & $2.29 \pm 0.22$ \\
NK cells & $1.67 \pm 0.52$ & $3.98 \pm 0.00$ & $3.98 \pm 0.00$ \\
\midrule
\multicolumn{4}{l}{\textit{NDCG}} \\
\midrule
B cells & $22.47 \pm 6.81$ & $80.36 \pm 1.79$ & $78.90 \pm 2.41$ \\
CD14$^+$ monocytes & $26.74 \pm 7.25$ & $65.73 \pm 1.52$ & $64.32 \pm 2.23$ \\
CD4 T cells & $31.53 \pm 8.07$ & $90.60 \pm 1.90$ & $90.09 \pm 1.79$ \\
CD8 T cells & $21.78 \pm 5.40$ & $85.18 \pm 2.20$ & $86.21 \pm 3.02$ \\
Dendritic cells & $23.25 \pm 1.82$ & $89.19 \pm 3.35$ & $86.88 \pm 4.71$ \\
FCGR3A$^+$ monocytes & $22.01 \pm 4.67$ & $80.28 \pm 3.64$ & $75.57 \pm 10.85$ \\
NK cells & $23.80 \pm 7.67$ & $82.10 \pm 3.39$ & $79.07 \pm 5.60$ \\
\bottomrule
\end{tabular}
\end{table}

\begin{table}[H]
\centering
\caption{Recovery of pseudobulk DEGs per cell type at a top-$k$ cutoff of $k = 25$. Metrics and conventions are as in Table~\ref{tab:degs_cutoff10}.}
\label{tab:degs_cutoff25}
\small
\begin{tabular}{lccc}
\toprule
Cell type & CIG & DeepSHAP & Integrated gradients \\
\midrule
\multicolumn{4}{l}{\textit{Precision}} \\
\midrule
B cells & $56.80 \pm 12.46$ & $96.00 \pm 0.00$ & $96.00 \pm 0.00$ \\
CD14$^+$ monocytes & $60.80 \pm 11.45$ & $100.00 \pm 0.00$ & $97.60 \pm 3.58$ \\
CD4 T cells & $49.60 \pm 8.29$ & $96.00 \pm 0.00$ & $96.00 \pm 0.00$ \\
CD8 T cells & $48.80 \pm 5.93$ & $97.60 \pm 2.19$ & $94.40 \pm 3.58$ \\
Dendritic cells & $50.40 \pm 8.29$ & $98.40 \pm 2.19$ & $93.60 \pm 6.07$ \\
FCGR3A$^+$ monocytes & $52.00 \pm 9.38$ & $96.80 \pm 1.79$ & $89.60 \pm 10.43$ \\
NK cells & $55.20 \pm 11.10$ & $100.00 \pm 0.00$ & $94.40 \pm 4.56$ \\
\midrule
\multicolumn{4}{l}{\textit{Recall}} \\
\midrule
B cells & $2.99 \pm 0.66$ & $5.05 \pm 0.00$ & $5.05 \pm 0.00$ \\
CD14$^+$ monocytes & $0.77 \pm 0.15$ & $1.27 \pm 0.00$ & $1.24 \pm 0.05$ \\
CD4 T cells & $1.71 \pm 0.29$ & $3.31 \pm 0.00$ & $3.31 \pm 0.00$ \\
CD8 T cells & $8.91 \pm 1.08$ & $17.81 \pm 0.40$ & $17.23 \pm 0.65$ \\
Dendritic cells & $0.79 \pm 0.13$ & $1.55 \pm 0.03$ & $1.48 \pm 0.10$ \\
FCGR3A$^+$ monocytes & $1.60 \pm 0.29$ & $2.98 \pm 0.06$ & $2.76 \pm 0.32$ \\
NK cells & $2.80 \pm 0.56$ & $5.07 \pm 0.00$ & $4.79 \pm 0.23$ \\
\midrule
\multicolumn{4}{l}{\textit{F1-score}} \\
\midrule
B cells & $5.68 \pm 1.25$ & $9.60 \pm 0.00$ & $9.60 \pm 0.00$ \\
CD14$^+$ monocytes & $1.52 \pm 0.29$ & $2.51 \pm 0.00$ & $2.44 \pm 0.09$ \\
CD4 T cells & $3.31 \pm 0.55$ & $6.40 \pm 0.00$ & $6.40 \pm 0.00$ \\
CD8 T cells & $15.06 \pm 1.83$ & $30.12 \pm 0.68$ & $29.14 \pm 1.10$ \\
Dendritic cells & $1.57 \pm 0.26$ & $3.06 \pm 0.07$ & $2.91 \pm 0.19$ \\
FCGR3A$^+$ monocytes & $3.11 \pm 0.56$ & $5.79 \pm 0.11$ & $5.36 \pm 0.62$ \\
NK cells & $5.33 \pm 1.07$ & $9.65 \pm 0.00$ & $9.11 \pm 0.44$ \\
\midrule
\multicolumn{4}{l}{\textit{NDCG}} \\
\midrule
B cells & $36.41 \pm 8.36$ & $84.56 \pm 1.06$ & $84.29 \pm 1.12$ \\
CD14$^+$ monocytes & $37.72 \pm 8.14$ & $72.09 \pm 2.33$ & $68.99 \pm 4.00$ \\
CD4 T cells & $42.29 \pm 7.64$ & $86.70 \pm 0.44$ & $86.42 \pm 1.35$ \\
CD8 T cells & $38.39 \pm 5.40$ & $88.33 \pm 2.70$ & $87.86 \pm 1.61$ \\
Dendritic cells & $29.72 \pm 0.93$ & $88.77 \pm 1.86$ & $86.01 \pm 4.65$ \\
FCGR3A$^+$ monocytes & $33.65 \pm 6.57$ & $81.62 \pm 3.73$ & $75.42 \pm 11.52$ \\
NK cells & $40.16 \pm 7.37$ & $86.75 \pm 1.38$ & $82.96 \pm 2.21$ \\
\bottomrule
\end{tabular}
\end{table}

\begin{table}[H]
\centering
\caption{Recovery of pseudobulk DEGs per cell type at a top-$k$ cutoff of $k = 50$. Metrics and conventions are as in Table~\ref{tab:degs_cutoff10}.}
\label{tab:degs_cutoff50}
\small
\begin{tabular}{lccc}
\toprule
Cell type & CIG & DeepSHAP & Integrated gradients \\
\midrule
\multicolumn{4}{l}{\textit{Precision}} \\
\midrule
B cells & $62.80 \pm 4.82$ & $98.00 \pm 0.00$ & $95.60 \pm 2.19$ \\
CD14$^+$ monocytes & $70.00 \pm 8.60$ & $98.80 \pm 1.10$ & $95.20 \pm 3.35$ \\
CD4 T cells & $59.60 \pm 5.55$ & $96.00 \pm 0.00$ & $96.00 \pm 0.00$ \\
CD8 T cells & $54.40 \pm 9.84$ & $94.80 \pm 1.79$ & $93.20 \pm 2.28$ \\
Dendritic cells & $61.20 \pm 4.60$ & $97.20 \pm 2.28$ & $94.40 \pm 5.37$ \\
FCGR3A$^+$ monocytes & $57.20 \pm 7.01$ & $95.60 \pm 3.58$ & $89.60 \pm 11.70$ \\
NK cells & $57.60 \pm 3.29$ & $97.60 \pm 0.89$ & $92.00 \pm 2.45$ \\
\midrule
\multicolumn{4}{l}{\textit{Recall}} \\
\midrule
B cells & $6.61 \pm 0.51$ & $10.32 \pm 0.00$ & $10.06 \pm 0.23$ \\
CD14$^+$ monocytes & $1.78 \pm 0.22$ & $2.51 \pm 0.03$ & $2.42 \pm 0.08$ \\
CD4 T cells & $4.11 \pm 0.38$ & $6.62 \pm 0.00$ & $6.62 \pm 0.00$ \\
CD8 T cells & $19.85 \pm 3.59$ & $34.60 \pm 0.65$ & $34.01 \pm 0.83$ \\
Dendritic cells & $1.93 \pm 0.15$ & $3.07 \pm 0.07$ & $2.98 \pm 0.17$ \\
FCGR3A$^+$ monocytes & $3.53 \pm 0.43$ & $5.89 \pm 0.22$ & $5.52 \pm 0.72$ \\
NK cells & $5.84 \pm 0.33$ & $9.90 \pm 0.09$ & $9.33 \pm 0.25$ \\
\midrule
\multicolumn{4}{l}{\textit{F1-score}} \\
\midrule
B cells & $11.96 \pm 0.92$ & $18.67 \pm 0.00$ & $18.21 \pm 0.42$ \\
CD14$^+$ monocytes & $3.46 \pm 0.43$ & $4.89 \pm 0.05$ & $4.71 \pm 0.17$ \\
CD4 T cells & $7.69 \pm 0.72$ & $12.39 \pm 0.00$ & $12.39 \pm 0.00$ \\
CD8 T cells & $29.09 \pm 5.26$ & $50.70 \pm 0.96$ & $49.84 \pm 1.22$ \\
Dendritic cells & $3.74 \pm 0.28$ & $5.94 \pm 0.14$ & $5.77 \pm 0.33$ \\
FCGR3A$^+$ monocytes & $6.64 \pm 0.81$ & $11.10 \pm 0.42$ & $10.41 \pm 1.36$ \\
NK cells & $10.61 \pm 0.61$ & $17.97 \pm 0.16$ & $16.94 \pm 0.45$ \\
\midrule
\multicolumn{4}{l}{\textit{NDCG}} \\
\midrule
B cells & $50.37 \pm 5.57$ & $88.95 \pm 0.93$ & $87.63 \pm 0.86$ \\
CD14$^+$ monocytes & $47.75 \pm 8.26$ & $76.77 \pm 0.87$ & $73.66 \pm 3.80$ \\
CD4 T cells & $49.28 \pm 5.62$ & $88.08 \pm 0.44$ & $87.25 \pm 0.49$ \\
CD8 T cells & $49.77 \pm 6.69$ & $88.84 \pm 1.67$ & $88.82 \pm 1.46$ \\
Dendritic cells & $43.99 \pm 2.23$ & $89.36 \pm 1.69$ & $87.23 \pm 4.76$ \\
FCGR3A$^+$ monocytes & $43.06 \pm 5.69$ & $82.92 \pm 3.14$ & $78.24 \pm 10.25$ \\
NK cells & $50.69 \pm 4.57$ & $91.35 \pm 1.33$ & $87.85 \pm 2.83$ \\
\bottomrule
\end{tabular}
\end{table}

\begin{table}[H]
\centering
\caption{Recovery of pseudobulk DEGs per cell type at a top-$k$ cutoff of $k = 100$. Metrics and conventions are as in Table~\ref{tab:degs_cutoff10}.}
\label{tab:degs_cutoff100}
\small
\begin{tabular}{lccc}
\toprule
Cell type & CIG & DeepSHAP & Integrated gradients \\
\midrule
\multicolumn{4}{l}{\textit{Precision}} \\
\midrule
B cells & $64.00 \pm 1.87$ & $93.80 \pm 2.68$ & $90.40 \pm 2.97$ \\
CD14$^+$ monocytes & $74.60 \pm 6.23$ & $95.80 \pm 1.30$ & $92.60 \pm 4.39$ \\
CD4 T cells & $61.20 \pm 4.60$ & $91.00 \pm 0.71$ & $89.20 \pm 1.79$ \\
CD8 T cells & $57.00 \pm 3.61$ & $78.40 \pm 2.30$ & $75.40 \pm 1.82$ \\
Dendritic cells & $67.40 \pm 2.19$ & $92.40 \pm 3.65$ & $89.40 \pm 4.16$ \\
FCGR3A$^+$ monocytes & $64.40 \pm 7.23$ & $93.80 \pm 4.97$ & $87.60 \pm 9.76$ \\
NK cells & $56.60 \pm 3.21$ & $85.20 \pm 1.79$ & $80.40 \pm 1.95$ \\
\midrule
\multicolumn{4}{l}{\textit{Recall}} \\
\midrule
B cells & $13.47 \pm 0.39$ & $19.75 \pm 0.56$ & $19.03 \pm 0.62$ \\
CD14$^+$ monocytes & $3.78 \pm 0.32$ & $4.86 \pm 0.07$ & $4.70 \pm 0.22$ \\
CD4 T cells & $8.44 \pm 0.64$ & $12.55 \pm 0.10$ & $12.30 \pm 0.25$ \\
CD8 T cells & $41.61 \pm 2.63$ & $57.23 \pm 1.68$ & $55.04 \pm 1.33$ \\
Dendritic cells & $4.25 \pm 0.14$ & $5.83 \pm 0.23$ & $5.64 \pm 0.26$ \\
FCGR3A$^+$ monocytes & $7.94 \pm 0.89$ & $11.57 \pm 0.61$ & $10.80 \pm 1.20$ \\
NK cells & $11.48 \pm 0.65$ & $17.28 \pm 0.36$ & $16.31 \pm 0.40$ \\
\midrule
\multicolumn{4}{l}{\textit{F1-score}} \\
\midrule
B cells & $22.26 \pm 0.65$ & $32.63 \pm 0.93$ & $31.44 \pm 1.03$ \\
CD14$^+$ monocytes & $7.20 \pm 0.60$ & $9.25 \pm 0.13$ & $8.94 \pm 0.42$ \\
CD4 T cells & $14.84 \pm 1.12$ & $22.06 \pm 0.17$ & $21.62 \pm 0.43$ \\
CD8 T cells & $48.10 \pm 3.04$ & $66.16 \pm 1.94$ & $63.63 \pm 1.53$ \\
Dendritic cells & $8.00 \pm 0.26$ & $10.97 \pm 0.43$ & $10.61 \pm 0.49$ \\
FCGR3A$^+$ monocytes & $14.14 \pm 1.59$ & $20.59 \pm 1.09$ & $19.23 \pm 2.14$ \\
NK cells & $19.09 \pm 1.08$ & $28.74 \pm 0.60$ & $27.12 \pm 0.66$ \\
\midrule
\multicolumn{4}{l}{\textit{NDCG}} \\
\midrule
B cells & $59.58 \pm 4.15$ & $90.46 \pm 0.74$ & $89.41 \pm 0.72$ \\
CD14$^+$ monocytes & $56.36 \pm 6.72$ & $78.89 \pm 1.23$ & $77.29 \pm 3.39$ \\
CD4 T cells & $56.25 \pm 3.86$ & $89.83 \pm 0.30$ & $89.33 \pm 0.92$ \\
CD8 T cells & $58.47 \pm 3.97$ & $89.23 \pm 1.19$ & $88.31 \pm 1.07$ \\
Dendritic cells & $55.23 \pm 1.11$ & $88.55 \pm 1.66$ & $86.61 \pm 3.17$ \\
FCGR3A$^+$ monocytes & $54.48 \pm 5.25$ & $85.40 \pm 2.68$ & $80.86 \pm 7.76$ \\
NK cells & $60.68 \pm 4.01$ & $90.10 \pm 1.11$ & $87.18 \pm 2.34$ \\
\bottomrule
\end{tabular}
\end{table}

\begin{table}[H]
\centering
\caption{Recovery of pseudobulk DEGs per cell type at a top-$k$ cutoff of $k = 250$. Metrics and conventions are as in Table~\ref{tab:degs_cutoff10}.}
\label{tab:degs_cutoff250}
\small
\begin{tabular}{lccc}
\toprule
Cell type & CIG & DeepSHAP & Integrated gradients \\
\midrule
\multicolumn{4}{l}{\textit{Precision}} \\
\midrule
B cells & $56.48 \pm 1.43$ & $76.64 \pm 1.15$ & $73.28 \pm 1.37$ \\
CD14$^+$ monocytes & $71.28 \pm 4.21$ & $84.88 \pm 1.80$ & $81.60 \pm 3.97$ \\
CD4 T cells & $52.64 \pm 1.00$ & $66.72 \pm 1.58$ & $66.00 \pm 1.52$ \\
CD8 T cells & $37.60 \pm 1.96$ & $45.84 \pm 0.83$ & $45.28 \pm 0.87$ \\
Dendritic cells & $64.08 \pm 2.05$ & $79.28 \pm 1.18$ & $75.76 \pm 2.31$ \\
FCGR3A$^+$ monocytes & $63.12 \pm 6.52$ & $82.24 \pm 3.28$ & $75.60 \pm 6.80$ \\
NK cells & $47.36 \pm 1.00$ & $59.84 \pm 0.61$ & $57.52 \pm 1.11$ \\
\midrule
\multicolumn{4}{l}{\textit{Recall}} \\
\midrule
B cells & $29.73 \pm 0.75$ & $40.34 \pm 0.61$ & $38.57 \pm 0.72$ \\
CD14$^+$ monocytes & $9.04 \pm 0.53$ & $10.77 \pm 0.23$ & $10.35 \pm 0.50$ \\
CD4 T cells & $18.15 \pm 0.35$ & $23.01 \pm 0.55$ & $22.76 \pm 0.53$ \\
CD8 T cells & $68.61 \pm 3.58$ & $83.65 \pm 1.51$ & $82.63 \pm 1.58$ \\
Dendritic cells & $10.11 \pm 0.32$ & $12.50 \pm 0.19$ & $11.95 \pm 0.36$ \\
FCGR3A$^+$ monocytes & $19.46 \pm 2.01$ & $25.35 \pm 1.01$ & $23.30 \pm 2.10$ \\
NK cells & $24.02 \pm 0.51$ & $30.34 \pm 0.31$ & $29.17 \pm 0.56$ \\
\midrule
\multicolumn{4}{l}{\textit{F1-score}} \\
\midrule
B cells & $38.95 \pm 0.98$ & $52.86 \pm 0.79$ & $50.54 \pm 0.94$ \\
CD14$^+$ monocytes & $16.05 \pm 0.95$ & $19.11 \pm 0.40$ & $18.37 \pm 0.89$ \\
CD4 T cells & $26.99 \pm 0.51$ & $34.22 \pm 0.81$ & $33.85 \pm 0.78$ \\
CD8 T cells & $48.58 \pm 2.53$ & $59.22 \pm 1.07$ & $58.50 \pm 1.12$ \\
Dendritic cells & $17.46 \pm 0.56$ & $21.60 \pm 0.32$ & $20.64 \pm 0.63$ \\
FCGR3A$^+$ monocytes & $29.75 \pm 3.07$ & $38.76 \pm 1.55$ & $35.63 \pm 3.20$ \\
NK cells & $31.87 \pm 0.68$ & $40.27 \pm 0.41$ & $38.71 \pm 0.75$ \\
\midrule
\multicolumn{4}{l}{\textit{NDCG}} \\
\midrule
B cells & $66.91 \pm 3.33$ & $91.38 \pm 0.70$ & $89.66 \pm 0.85$ \\
CD14$^+$ monocytes & $63.63 \pm 4.69$ & $80.91 \pm 1.36$ & $78.33 \pm 3.34$ \\
CD4 T cells & $65.69 \pm 2.19$ & $89.48 \pm 0.47$ & $88.29 \pm 1.11$ \\
CD8 T cells & $63.11 \pm 3.55$ & $87.34 \pm 0.72$ & $86.84 \pm 1.32$ \\
Dendritic cells & $64.07 \pm 1.61$ & $86.08 \pm 0.57$ & $83.70 \pm 2.08$ \\
FCGR3A$^+$ monocytes & $64.74 \pm 4.81$ & $85.70 \pm 2.40$ & $81.60 \pm 6.17$ \\
NK cells & $66.59 \pm 3.00$ & $87.85 \pm 1.04$ & $85.28 \pm 1.88$ \\
\bottomrule
\end{tabular}
\end{table}

\subsection{Over-representation analysis of top-ranked genes} \label{appendix:go_enrichment}
\begin{table}[H]
\centering
\small
\setlength{\tabcolsep}{3pt}
\caption{Over-representation analysis of the top-100 genes unique to CIG in dendritic cells. Terms are gene sets from the MSigDB Hallmark 2020 collection, and the measured gene panel was used as the background set. The ten most significant terms are listed. $P$-values come from a hypergeometric test and are Benjamini--Hochberg adjusted across the collection; the combined score is $\ln(p)$ multiplied by the $z$-score of the deviation from the expected rank. The last column gives the genes shared between the top-100 set and the term. Adjusted $P$-values in bold are significant at $P < 0.01$.}
\label{tab:go_dc}
\begin{tabular}{>{\raggedright\arraybackslash}p{0.21\linewidth}rrr>{\raggedright\arraybackslash}p{0.26\linewidth}}
\toprule
Term  & Adjusted P-value  & Odds Ratio  & Combined Score  & Genes \\
\midrule
TNF-alpha Signaling via NF-kB & \textbf{0.0003} & 14.1062 & 156.8329 & JUN, KLF6, IRF1, MYC, FOS, NFE2L2 \\
Hypoxia & \textbf{0.0003} & 17.0927 & 178.5641 & JUN, KLF6, FOS, NR3C1, ETS1 \\
Interferon Gamma Response & \textbf{0.0003} & 12.1222 & 124.9307 & IRF4, IRF1, STAT2, IRF2, IRF8, HIF1A \\
G2-M Checkpoint & \textbf{0.0041} & 11.7922 & 86.3737 & MYC, CTCF, E2F4, HIF1A \\
IL-2/STAT5 Signaling & \textbf{0.0041} & 10.6567 & 74.2016 & KLF6, IRF4, MYC, IRF8 \\
Inflammatory Response & \textbf{0.0041} & 10.5641 & 73.2283 & KLF6, IRF1, MYC, HIF1A \\
UV Response Dn & \textbf{0.0041} & 17.5550 & 120.9222 & MYC, NR3C1, RUNX1 \\
Wnt-beta Catenin Signaling & \textbf{0.0047} & 44.2436 & 291.5354 & MYC, CTNNB1 \\
IL-6/JAK/STAT3 Signaling & \textbf{0.0047} & 15.1570 & 98.3114 & JUN, IRF1, STAT2 \\
Allograft Rejection & \textbf{0.0048} & 8.9954 & 57.2430 & IRF4, IRF8, ETS1, HIF1A \\
\bottomrule
\end{tabular}
\end{table}

\begin{table}[H]
\centering
\small
\setlength{\tabcolsep}{3pt}
\caption{Over-representation analysis of the top-100 genes unique to CIG in CD14$^+$ monocytes, against the MSigDB Hallmark 2020 gene sets with the measured gene panel as background. Columns are as in Table~\ref{tab:go_dc}. Adjusted $P$-values in bold are significant at $P < 0.01$.}
\label{tab:go_cd14}
\begin{tabular}{>{\raggedright\arraybackslash}p{0.21\linewidth}rrr>{\raggedright\arraybackslash}p{0.26\linewidth}}
\toprule
Term  & Adjusted P-value  & Odds Ratio  & Combined Score  & Genes \\
\midrule
TNF-alpha Signaling via NF-kB & \textbf{0.0001} & 17.6432 & 214.7754 & CEBPB, JUN, IRF1, MYC, FOS, NFE2L2 \\
Interferon Gamma Response & \textbf{0.0002} & 15.1617 & 172.0823 & STAT1, IRF1, STAT2, IRF2, IRF8, HIF1A \\
IL-6/JAK/STAT3 Signaling & \textbf{0.0003} & 26.2874 & 267.9484 & JUN, STAT1, IRF1, STAT2 \\
Hypoxia & \textbf{0.0016} & 15.7940 & 131.4392 & JUN, FOS, NR3C1, ETS1 \\
G2-M Checkpoint & \textbf{0.0018} & 14.4212 & 115.2408 & MYC, CTCF, E2F4, HIF1A \\
UV Response Dn & \textbf{0.0027} & 21.2632 & 157.1334 & MYC, NR3C1, RUNX1 \\
Allograft Rejection & \textbf{0.0033} & 11.0008 & 77.1754 & STAT1, IRF8, ETS1, HIF1A \\
Interferon Alpha Response & \textbf{0.0088} & 12.3993 & 73.2617 & IRF1, STAT2, IRF2 \\
Complement & 0.0162 & 9.1004 & 46.1886 & CEBPB, IRF1, IRF2 \\
Inflammatory Response & 0.0162 & 9.1004 & 46.1886 & IRF1, MYC, HIF1A \\
\bottomrule
\end{tabular}
\end{table}

\begin{table}[H]
\centering
\small
\setlength{\tabcolsep}{3pt}
\caption{Over-representation analysis of the top-100 genes unique to CIG in FCGR3A$^+$ monocytes, against the MSigDB Hallmark 2020 gene sets with the measured gene panel as background. Columns are as in Table~\ref{tab:go_dc}. Adjusted $P$-values in bold are significant at $P < 0.01$.}
\label{tab:go_fcgr3a}
\begin{tabular}{>{\raggedright\arraybackslash}p{0.21\linewidth}rrr>{\raggedright\arraybackslash}p{0.26\linewidth}}
\toprule
Term  & Adjusted P-value  & Odds Ratio  & Combined Score  & Genes \\
\midrule
IL-6/JAK/STAT3 Signaling & \textbf{0.0021} & 21.4953 & 204.3132 & JUN, STAT1, STAT2, STAT3 \\
TNF-alpha Signaling via NF-kB & \textbf{0.0028} & 11.1166 & 94.6276 & CEBPB, JUN, MYC, FOS, NFE2L2 \\
Interferon Gamma Response & \textbf{0.0036} & 9.5621 & 75.0479 & STAT1, STAT2, IRF2, STAT3, IRF8 \\
G2-M Checkpoint & \textbf{0.0046} & 11.7922 & 86.3737 & MYC, CTCF, E2F4, ATF5 \\
UV Response Dn & \textbf{0.0057} & 17.5550 & 120.9222 & MYC, NR3C1, RUNX1 \\
Wnt-beta Catenin Signaling & \textbf{0.0064} & 44.2436 & 291.5354 & MYC, CTNNB1 \\
Hypoxia & 0.0237 & 9.1681 & 47.0092 & JUN, FOS, NR3C1 \\
Cholesterol Homeostasis & 0.0321 & 15.4910 & 72.6594 & CTNNB1, ATF5 \\
Allograft Rejection & 0.0472 & 6.4055 & 26.8338 & SPI1, STAT1, IRF8 \\
Estrogen Response Early & 0.0585 & 9.8520 & 38.1050 & MYC, FOS \\
\bottomrule
\end{tabular}
\end{table}

\begin{table}[H]
\centering
\small
\setlength{\tabcolsep}{3pt}
\caption{Over-representation analysis of the top-100 genes unique to integrated gradients in dendritic cells, against the MSigDB Hallmark 2020 gene sets with the measured gene panel as background; the most significant terms are listed. Columns are as in Table~\ref{tab:go_dc}. Adjusted $P$-values in bold are significant at $P < 0.01$.}
\label{tab:go_ig_dc}
\begin{tabular}{>{\raggedright\arraybackslash}p{0.29\linewidth}rrr>{\raggedright\arraybackslash}p{0.22\linewidth}}
\toprule
Term & Adjusted P-value & Odds Ratio & Combined Score & Genes \\
\midrule
Allograft Rejection & \textbf{0.0049} & 97.8273 & 656.8822 & CD74, CCL22 \\
Interferon Alpha Response & 0.0685 & 38.4775 & 125.4228 & CD74 \\
Inflammatory Response & 0.0685 & 28.4083 & 84.3200 & CCL22 \\
Interferon Gamma Response & 0.0729 & 19.6686 & 51.4961 & CD74 \\
\bottomrule
\end{tabular}
\end{table}

\begin{table}[H]
\centering
\small
\setlength{\tabcolsep}{3pt}
\caption{Over-representation analysis of the top-100 genes unique to integrated gradients in CD14$^+$ monocytes, against the MSigDB Hallmark 2020 gene sets with the measured gene panel as background; the most significant terms are listed. Columns are as in Table~\ref{tab:go_dc}. Adjusted $P$-values in bold are significant at $P < 0.01$.}
\label{tab:go_ig_cd14}
\begin{tabular}{>{\raggedright\arraybackslash}p{0.29\linewidth}rrr>{\raggedright\arraybackslash}p{0.22\linewidth}}
\toprule
Term & Adjusted P-value & Odds Ratio & Combined Score & Genes \\
\midrule
Complement & 0.0587 & 22.9076 & 117.2823 & PLAUR, PFN1 \\
Allograft Rejection & 0.0587 & 19.5540 & 94.2961 & SOCS1, B2M \\
Interferon Gamma Response & 0.0587 & 15.8199 & 70.0418 & SOCS1, B2M \\
Apical Surface & 0.0587 & 96.1389 & 417.1983 & PLAUR \\
Cholesterol Homeostasis & 0.1113 & 30.2456 & 98.3633 & PLAUR \\
Epithelial Mesenchymal Transition & 0.1113 & 22.9467 & 68.5890 & PLAUR \\
Myogenesis & 0.1113 & 22.4935 & 66.8073 & PDE4DIP \\
Androgen Response & 0.1113 & 22.0577 & 65.1024 & B2M \\
IL-6/JAK/STAT3 Signaling & 0.1113 & 18.7787 & 52.5521 & SOCS1 \\
Apical Junction & 0.1113 & 18.4731 & 51.4093 & PFN1 \\
\bottomrule
\end{tabular}
\end{table}

\begin{table}[H]
\centering
\small
\setlength{\tabcolsep}{3pt}
\caption{Over-representation analysis of the top-100 genes unique to integrated gradients in FCGR3A$^+$ monocytes, against the MSigDB Hallmark 2020 gene sets with the measured gene panel as background; the most significant terms are listed. Columns are as in Table~\ref{tab:go_dc}. Adjusted $P$-values in bold are significant at $P < 0.01$.}
\label{tab:go_ig_fcgr3a}
\begin{tabular}{>{\raggedright\arraybackslash}p{0.29\linewidth}rrr>{\raggedright\arraybackslash}p{0.22\linewidth}}
\toprule
Term & Adjusted P-value & Odds Ratio & Combined Score & Genes \\
\midrule
Epithelial Mesenchymal Transition & 0.1406 & 27.5400 & 86.4653 & GLIPR1 \\
Interferon Alpha Response & 0.1406 & 15.3843 & 39.7786 & PARP9 \\
Apoptosis & 0.1406 & 12.7626 & 30.7555 & CD69 \\
Inflammatory Response & 0.1406 & 11.3583 & 26.1335 & CD69 \\
TNF-alpha Signaling via NF-kB & 0.1406 & 9.1087 & 19.0955 & CD69 \\
Interferon Gamma Response & 0.1406 & 7.8640 & 15.4290 & CD69 \\
\bottomrule
\end{tabular}
\end{table}

\begin{table}[H]
\centering
\small
\setlength{\tabcolsep}{3pt}
\caption{Over-representation analysis of the top-100 genes unique to DeepSHAP in dendritic cells, against the MSigDB Hallmark 2020 gene sets with the measured gene panel as background; the most significant terms are listed. Columns are as in Table~\ref{tab:go_dc}. Adjusted $P$-values in bold are significant at $P < 0.01$.}
\label{tab:go_ds_dc}
\begin{tabular}{>{\raggedright\arraybackslash}p{0.29\linewidth}rrr>{\raggedright\arraybackslash}p{0.22\linewidth}}
\toprule
Term & Adjusted P-value & Odds Ratio & Combined Score & Genes \\
\midrule
Interferon Alpha Response & \textbf{0.0000} & 236.2069 & 2761.8748 & UBA7, PSME2, NCOA7 \\
Interferon Gamma Response & \textbf{0.0090} & 39.5673 & 222.7963 & CXCL9, PSME2 \\
IL-6/JAK/STAT3 Signaling & 0.0588 & 37.5738 & 125.6833 & CXCL9 \\
Inflammatory Response & 0.0788 & 18.9361 & 50.9210 & CXCL9 \\
Allograft Rejection & 0.0788 & 16.1833 & 41.1132 & CXCL9 \\
\bottomrule
\end{tabular}
\end{table}

\begin{table}[H]
\centering
\small
\setlength{\tabcolsep}{3pt}
\caption{Over-representation analysis of the top-100 genes unique to DeepSHAP in CD14$^+$ monocytes, against the MSigDB Hallmark 2020 gene sets with the measured gene panel as background; the most significant terms are listed. Columns are as in Table~\ref{tab:go_dc}. Adjusted $P$-values in bold are significant at $P < 0.01$.}
\label{tab:go_ds_cd14}
\begin{tabular}{>{\raggedright\arraybackslash}p{0.29\linewidth}rrr>{\raggedright\arraybackslash}p{0.22\linewidth}}
\toprule
Term & Adjusted P-value & Odds Ratio & Combined Score & Genes \\
\midrule
Interferon Alpha Response & 0.0335 & 31.1182 & 177.3636 & DHX58, PARP12 \\
Interferon Gamma Response & 0.0597 & 15.8199 & 70.0418 & DHX58, PARP12 \\
Cholesterol Homeostasis & 0.0992 & 30.2456 & 98.3633 & CD9 \\
Coagulation & 0.0992 & 26.7093 & 83.6984 & CD9 \\
IL-6/JAK/STAT3 Signaling & 0.0992 & 18.7787 & 52.5521 & CD9 \\
Protein Secretion & 0.0992 & 16.8284 & 45.3449 & NAPA \\
Estrogen Response Late & 0.0992 & 16.3429 & 43.5836 & CD9 \\
Inflammatory Response & 0.1448 & 9.4639 & 20.3969 & OSM \\
mTORC1 Signaling & 0.1557 & 7.6950 & 15.1238 & CD9 \\
Myc Targets V1 & 0.1698 & 6.2182 & 11.0261 & RPL6 \\
\bottomrule
\end{tabular}
\end{table}

\begin{table}[H]
\centering
\small
\setlength{\tabcolsep}{3pt}
\caption{Over-representation analysis of the top-100 genes unique to DeepSHAP in FCGR3A$^+$ monocytes, against the MSigDB Hallmark 2020 gene sets with the measured gene panel as background; the most significant terms are listed. Columns are as in Table~\ref{tab:go_dc}. Adjusted $P$-values in bold are significant at $P < 0.01$.}
\label{tab:go_ds_fcgr3a}
\begin{tabular}{>{\raggedright\arraybackslash}p{0.29\linewidth}rrr>{\raggedright\arraybackslash}p{0.22\linewidth}}
\toprule
Term & Adjusted P-value & Odds Ratio & Combined Score & Genes \\
\midrule
Interferon Alpha Response & \textbf{0.0002} & 78.7126 & 795.1581 & BST2, IFITM2, IFI27 \\
Interferon Gamma Response & \textbf{0.0007} & 39.7941 & 324.4256 & BST2, IFITM2, IFI27 \\
Bile Acid Metabolism & 0.0554 & 36.3000 & 123.5496 & CH25H \\
Inflammatory Response & 0.1252 & 11.3583 & 26.1335 & BST2 \\
Myc Targets V1 & 0.1474 & 7.4630 & 14.2879 & RPL6 \\
\bottomrule
\end{tabular}
\end{table}

\section{Isolating the value of CIG's formulation beyond graph access} \label{appendix:naive_baseline_locality}

A possible concern is that CIG's advantage is simply a readout of having access to the causal graph, not of correctly using it. We address this on Dyngen with a test of instance-level specificity that a fixed, shared graph cannot explain on its own. Because the Dyngen causal graph is identical across cells, a graph-driven attribution method would be expected to produce similar TF importance across cells. In contrast, cells undergoing different state transitions should be explained by different sets of transition-specific TFs. For each transition, we therefore restrict the analysis to cells undergoing that transition and compare the attribution ranks of its known causal TFs against those of all other genes.

\begin{figure}[H]
    \centering
     \includegraphics[width=\textwidth]{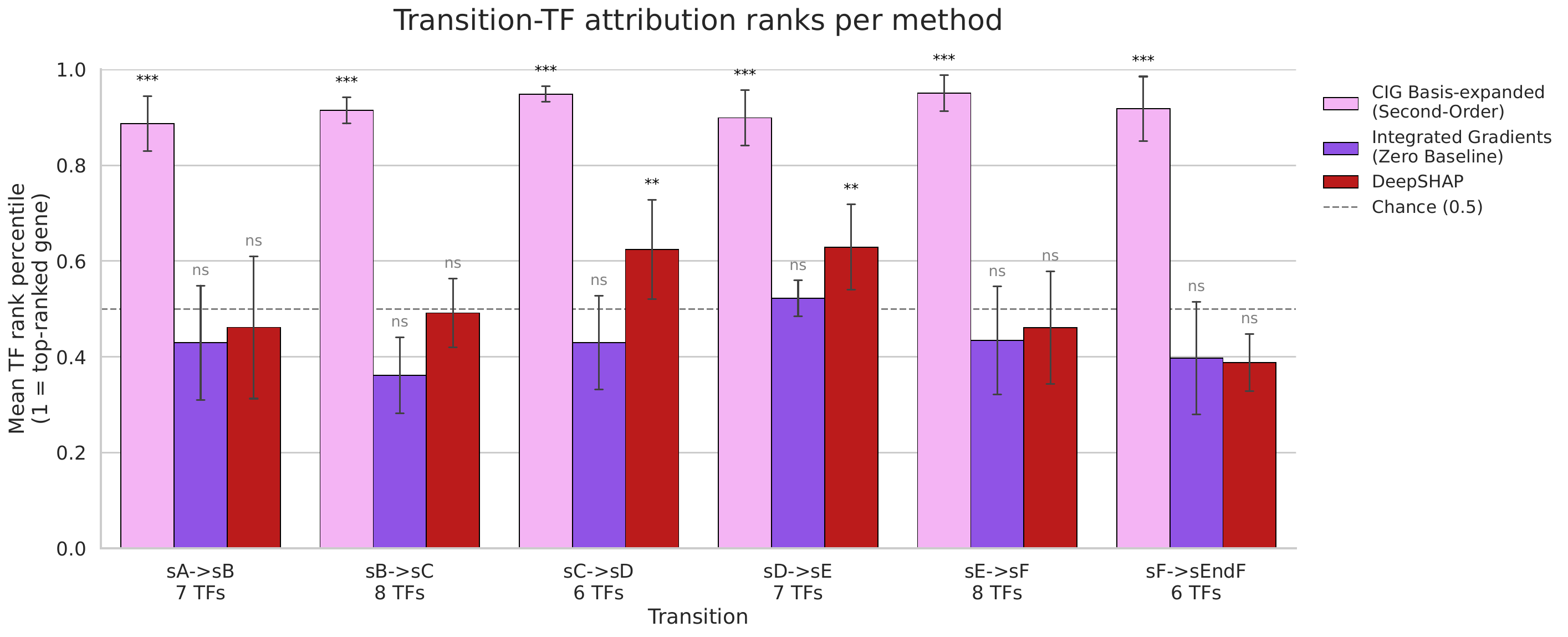}
\caption{Transition-TF attribution ranks by method. Bars show the mean TF rank across five seeds, with error bars indicating the standard deviation. Significance is assessed using a gene-label permutation test, Stouffer-combined across seeds and Benjamini--Hochberg adjusted. Significance levels: $\ast\ast\ast$: $q<0.001$; $\ast\ast$: $q<0.01$; $\ast$: $q<0.05$; ns, not significant.}
    \label{fig:locality_test_CIG}
\end{figure}

CIG consistently prioritizes the causal TFs for all six transitions, with significant enrichment in every case (Figure~\ref{fig:locality_test_CIG}). DeepSHAP shows significant prioritization in two of the six transitions, while integrated gradients does not show significant enrichment in any transition. The transition-specific differences in CIG attributions indicate that its explanations depend on the cells being explained, rather than simply reproducing the static causal graph.

\section{Robustness to causal graph misspecification} \label{appendix:graph_corruption}

We test how CIG's attribution quality degrades when the supplied causal graph is imperfect, using three corruption types applied at 25\%, 50\%, and 75\% of edges: removing true edges, adding spurious edges (cycle-safe, never removing a true edge), and reversing the direction of existing true edges. We apply this to both the synthetic SCM/CATE benchmark (Section~\ref{sec:causal_attributions_comparison}) and the Dyngen gene-regulatory network (Section~\ref{sec:dyngen_experiment}).

\begin{figure}[h]
    \centering
    \includegraphics[width=0.6\textwidth]{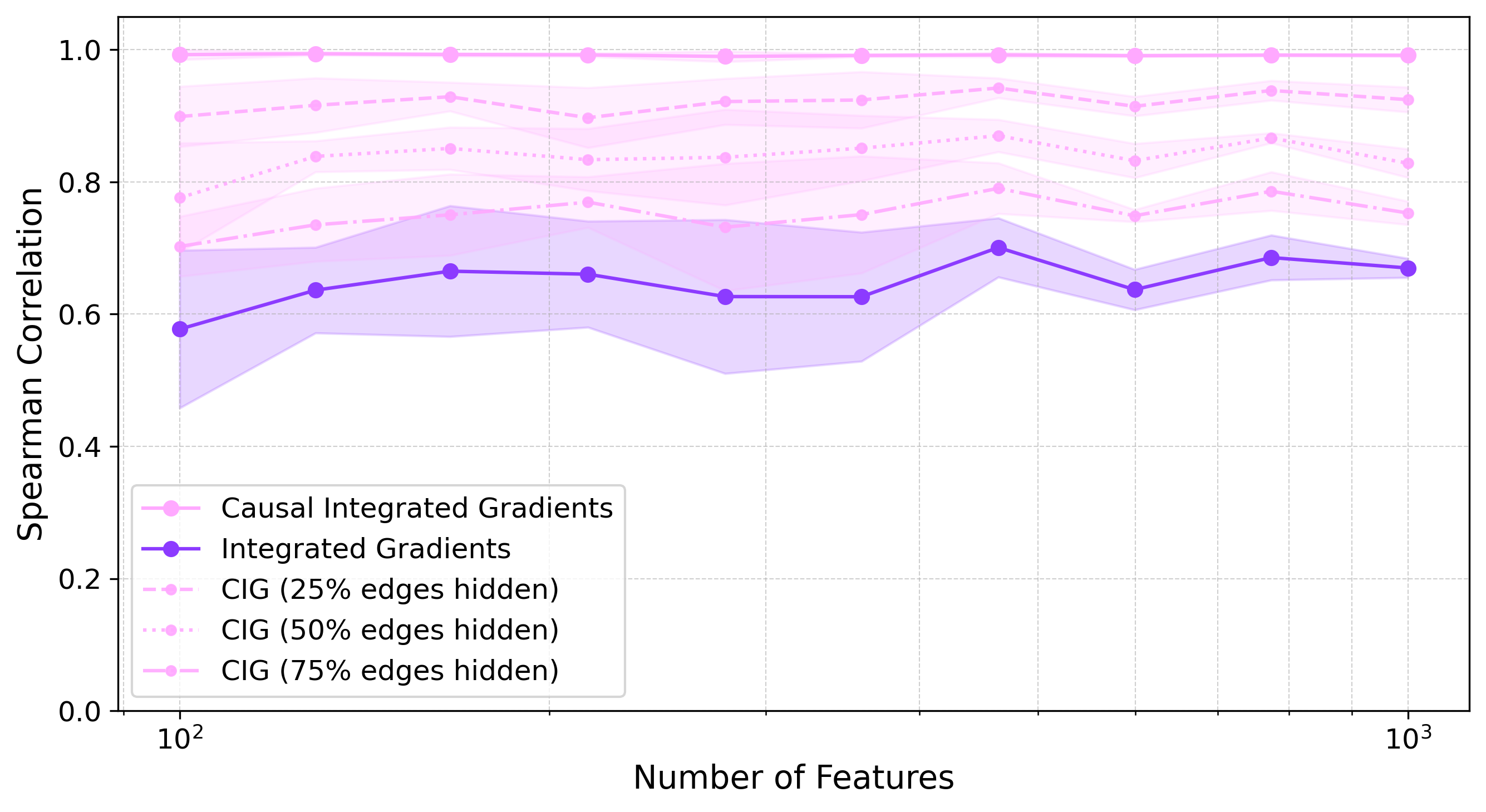}
    \caption{Spearman correlation with ground-truth CATE under edge removal (root-cause features), as a function of the number of features, at 25/50/75\% of edges removed. Shaded areas denote the standard deviation over 5 seeds.}
    \label{fig:graph_corruption_removed}
\end{figure}

\begin{figure}[h]
    \centering
    \includegraphics[width=0.6\textwidth]{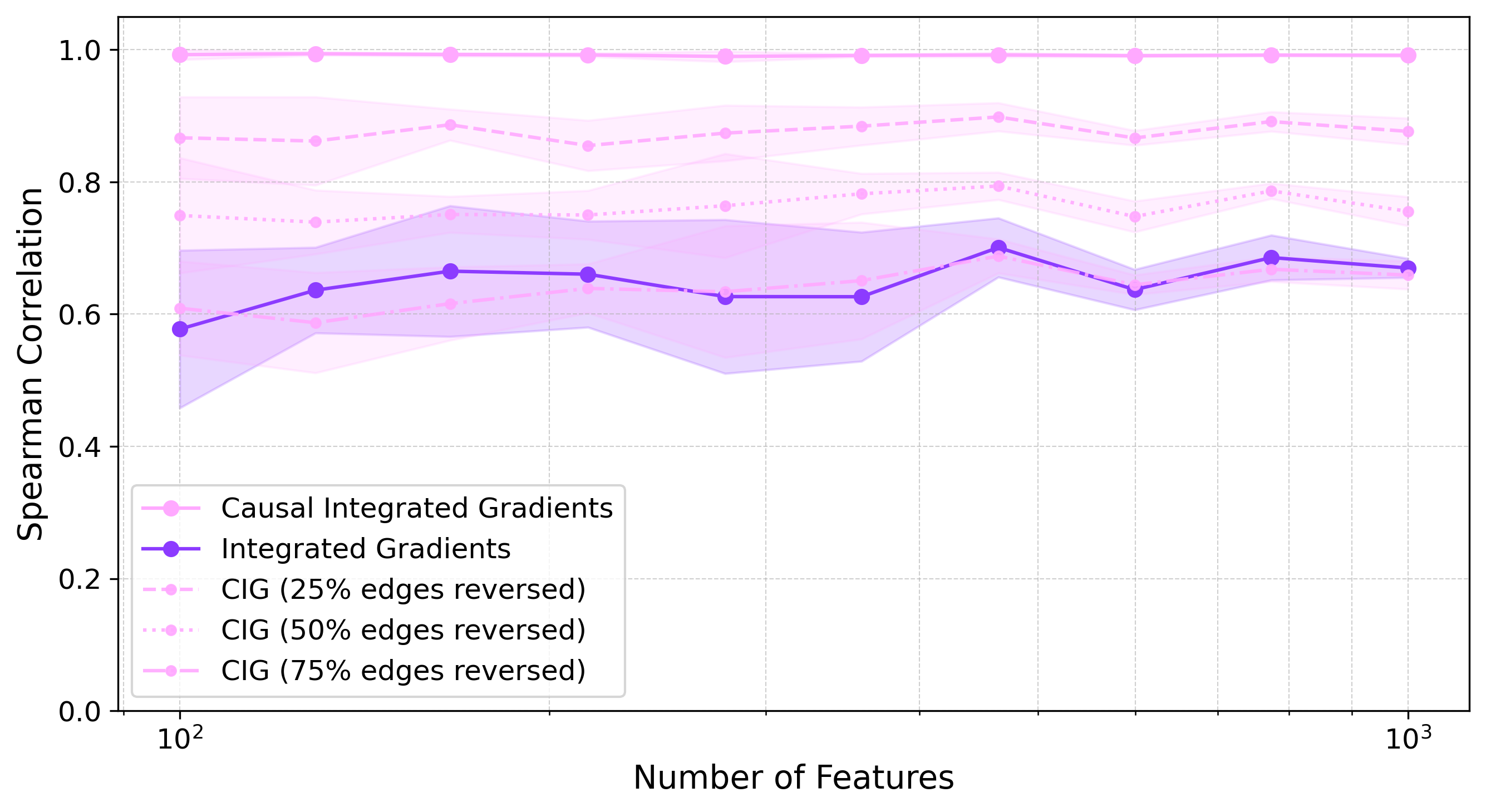}
    \caption{Spearman correlation with ground-truth CATE under edge reversal (root-cause features), as a function of the number of features, at 25/50/75\% of edges reversed. Shaded areas denote the standard deviation over 5 seeds.}
    \label{fig:graph_corruption_reversed}
\end{figure}

\begin{figure}[h]
    \centering
    \includegraphics[width=0.6\textwidth]{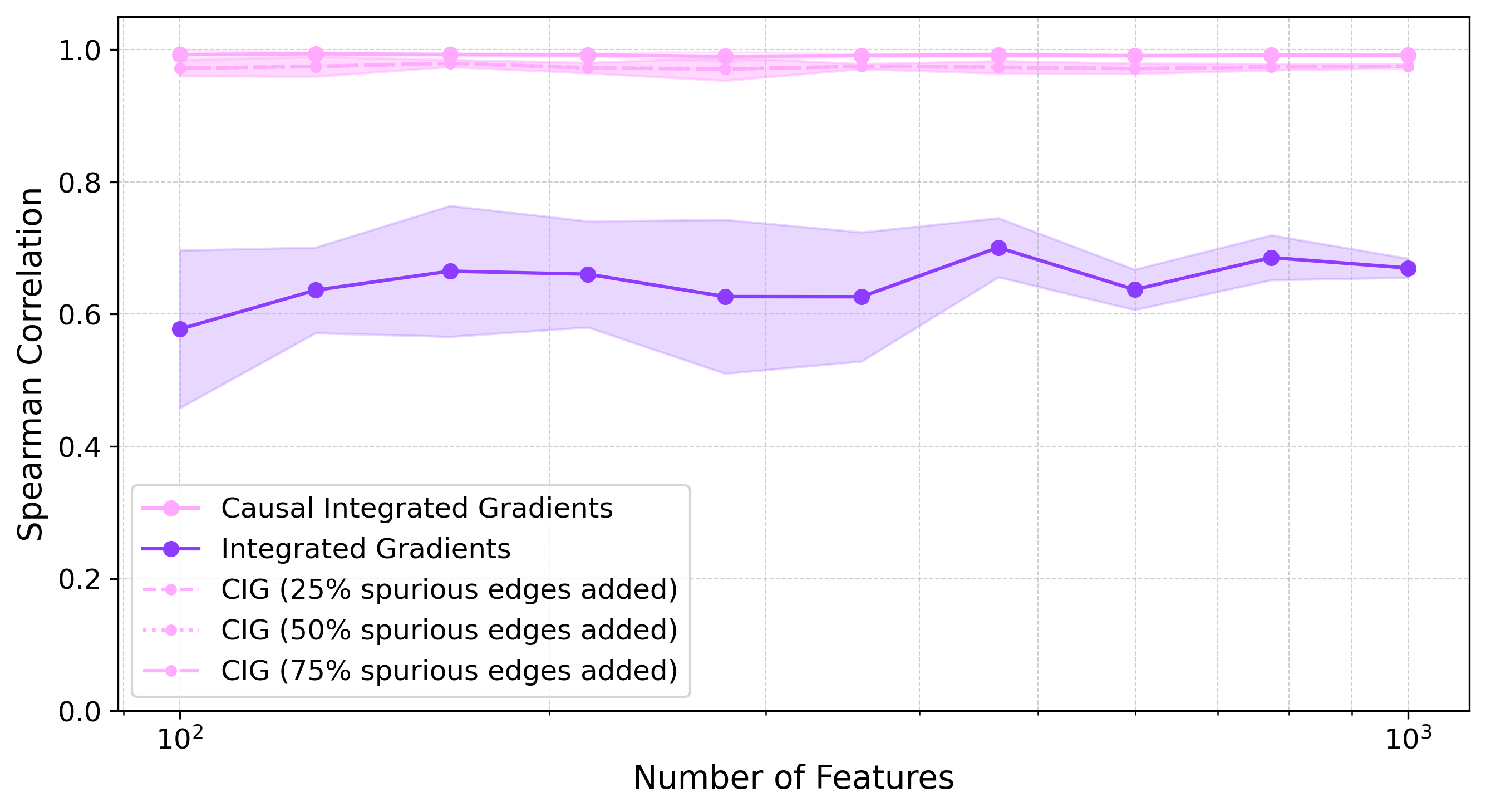}
    \caption{Spearman correlation with ground-truth CATE under spurious edge addition (root-cause features), as a function of the number of features, at 25/50/75\% of edges added. Shaded areas denote the standard deviation over 5 seeds.}
    \label{fig:graph_corruption_added}
\end{figure}

On the synthetic benchmark, we measure the Spearman correlation between CIG's attribution and the true CATE ranking (root-cause features), averaged over 5 seeds, across feature counts 100--1000. Adding spurious edges or reversing edge direction has no effect when the corrupted graph is fed to the oracle SCM mechanisms directly, since the oracle's forward pass only ever reads its hardcoded true-parent columns; for these two corruption types we instead re-fit a learned surrogate for each perturbed graph, matching how the method is used in practice. Edge removal is tested directly on the oracle mechanisms, as in the main benchmark. Figures~\ref{fig:graph_corruption_removed}--\ref{fig:graph_corruption_added} show results across the full feature sweep, one figure per corruption type.

On Dyngen, whose network separates \emph{backbone genes} (the core transcription-factor cascade driving the simulated differentiation trajectory) from \emph{housekeeping (HK) genes} (background genes uninvolved in any transition), we apply the same three corruption types to each community separately and report the normalized TF-recovery AUC metric used for the main Dyngen results (uncorrupted baseline: $97.76 \pm 3.18$). Table~\ref{tab:graph_corruption_dyngen} reports backbone/HK AUC pairs.
\begin{table}[H]
\centering
\caption{Normalized TF-recovery AUC under graph corruption on Dyngen, reported separately for the backbone and housekeeping genes (HK). Values are mean $\pm$ standard deviation over five seeds; uncorrupted baseline $97.76 \pm 3.18$.}
\label{tab:graph_corruption_dyngen}
\begin{tabular}{llccc}
\toprule
Corruption & Community & 25\% & 50\% & 75\% \\
\midrule
\multirow{2}{*}{Removed} & Backbone & $93.31 \pm 3.14$ & $89.50 \pm 2.44$ & $70.41 \pm 4.78$ \\
 & HK & $98.00 \pm 1.12$ & $97.93 \pm 1.01$ & $97.93 \pm 1.08$ \\
\midrule
\multirow{2}{*}{Reversed} & Backbone & $78.58 \pm 0.74$ & $61.80 \pm 2.73$ & $50.68 \pm 4.22$ \\
 & HK & $96.68 \pm 1.74$ & $96.38 \pm 1.91$ & $96.89 \pm 1.89$ \\
\midrule
\multirow{2}{*}{Added (spurious)} & Backbone & $96.38 \pm 1.54$ & $95.44 \pm 1.29$ & $93.85 \pm 2.74$ \\
 & HK & $97.99 \pm 1.10$ & $97.91 \pm 1.04$ & $97.81 \pm 1.09$ \\
\bottomrule
\end{tabular}
\end{table}

Across both datasets, the same pattern holds: removing or reversing edges degrades the performance of CIG substantially, while adding spurious edges barely decreases it, and in Dyngen, corrupting the housekeeping genes does not degrade the recovery of the causal TFs regardless of corruption type, while removing or reversing backbone edges degrades recovery increasingly with severity. Our hypothesis is that a spurious added edge does not correspond to any real dependency in the data-generating process; the surrogate has no true signal to associate with that edge, so its weight converges toward zero during training, making the gradient through it negligible even though the edge is present. Reversing or removing an edge, by contrast, is a genuine loss of information that the surrogate cannot recover, which is why both corruptions degrade recovery substantially. Together, both experiments show CIG's advantage is not solely inherited from a perfectly correct prior graph: it degrades under genuine information loss (removal, reversal) but tolerates redundant or causally irrelevant structural error (spurious edges; corruption confined to non-causal genes), performing about as well as with a perfect graph.

\section{Residual-dependence diagnostics under non-additive mechanisms} \label{appendix:residual_dependence}

Our surrogate models each feature's mechanism as additive across parents (Section~\ref{sec:surrogate}); when the true mechanism is non-additive, this can misallocate attribution between the direct and indirect terms (Section~\ref{sec:limitations}). We diagnose and quantify this bias directly, using the synthetic SCM/CATE benchmark (Section~\ref{sec:causal_attributions_comparison}), restricted to nodes whose mechanism is either the explicit two-parent interaction term $w_1x_1+w_2x_2+w_1w_2x_1x_2$ or a matched linear control (eligibility criteria below). Since ground truth is known for every node, we compare directly rather than inferring statistically.

A diagnostic based on the residual $h_j = x_j - \hat s_j(\text{parents})$ conflates the systematic mechanism misfit with injected exogenous noise, making it an inherently noisy signal of the bias itself. Instead, we isolate the same systematic misallocation without this contamination by comparing CIG's indirect-effect attribution computed with the oracle mechanism against the same attribution computed with the learned surrogate: both quantities are deterministic functions of the parents, so no noise term enters at all. We test two scenarios: the interacting variable present as a graph node (``in-graph''), and the interacting variable removed from the graph entirely (``omitted-variable'').

A node is eligible if its true mechanism matches the target type (interaction, or a matched linear control) and it is read directly by the downstream model, so that its attribution is nonzero; the linear control is further restricted to nodes with at least two parents (in-graph) or exactly two parents (omitted-variable), matching interaction's structural requirement of two active parents. The omitted-variable setting additionally restricts \emph{both} types to exactly two total parents, so that removing one parent's edge deletes an equal share of each node's explanatory input regardless of mechanism type. Eligible nodes are found by searching randomly generated graphs per setting, not fixed in advance, and pooled across 5 seeds (in-graph) and 15 seeds (omitted-variable).

For each eligible node, we compute the discrepancy between the oracle and surrogate indirect effects across sampled instances, and derive two statistics from that same discrepancy. The indirect effect is a path integral from a zero baseline, so along a linear path the source feature's own value enters via $dx_i/d\alpha = x_i$; the discrepancy is therefore expected to scale with the product of the two parents, $x_ix_k$, rather than either parent alone (confirmed empirically: correlating against the co-parent alone gives $r=-0.16$, $p=0.12$; against the product, $r=0.83$, $p\approx7\times10^{-27}$, on the same data). We report the Pearson correlation $r$ of the discrepancy against $x_ix_k$, and its relative error, $\mathrm{RMSE}(\text{discrepancy})/\mathrm{RMSE}(\text{oracle indirect effect})$. Because both statistics are computed from the same discrepancy per node, $n$ is identical for both columns within a row.\footnote{$p$-values are uncorrected across nodes and computed on non-independent samples (two per node); they characterize consistency across nodes, not a formal hypothesis test.} Table~\ref{tab:residual_dependence} reports the mean of $|r|$, the fraction of nodes with $p<0.05$, the median relative error, and the fraction of nodes with relative error exceeding 50\%.

\begin{table}[H]
\centering
\small
\caption{CIG's oracle-vs-surrogate indirect-effect discrepancy, by setting. Correlation is against the product of the node's two parents; the correlation and relative-error statistics are computed from the same per-node discrepancy, so $n$ is shared within each row.}
\label{tab:residual_dependence}
\begin{tabular}{llrrrrr}
\toprule
Setting & Type & $n$ & mean $|r|$ & \% sig. & Med.\ rel.\ err. & \% $>$50\% off \\
\midrule
In-graph & Interaction & 31 & 0.574 & 93.5\% & 0.286 & 6.5\% \\
In-graph & Linear (control) & 33 & 0.086 & 18.2\% & 0.288 & 0.0\% \\
Omitted-variable & Interaction & 42 & 0.264 & 73.8\% & 0.739 & 69.0\% \\
Omitted-variable & Linear (control) & 48 & 0.085 & 18.8\% & 0.789 & 81.3\% \\
\bottomrule
\end{tabular}
\end{table}

The discrepancy correlates far more strongly with the interaction term than with the linear control in both settings (3--7$\times$ the mean $|r|$, 4--5$\times$ the significance rate), confirming the bias is systematic rather than incidental: the indirect effect is predictably wrong wherever the mechanism is non-additive. The relative-error magnitude, by contrast, does not separate the two groups as cleanly: in the in-graph setting, interaction and linear are essentially identical (0.286 vs.\ 0.288), since overall surrogate fit quality there is dominated by factors common to both mechanism types (finite training, the downstream model's own regularization) rather than by the interaction term specifically; in the omitted-variable setting, both groups show substantial error because removing a parent's edge deletes its entire linear contribution regardless of mechanism type, which dominates the error budget for both. We therefore treat the correlation test as the primary evidence that the misallocation is systematic and tied to the interaction structure, and report relative error only to establish that indirect-effect errors of practically meaningful size do occur — not as an independent confirmation that interaction specifically causes larger errors than an equivalent linear mechanism. Despite these errors, the learned surrogate still achieves close to oracle-level Spearman correlation with the true CATE on the main benchmark (Section~\ref{sec:causal_attributions_comparison}).

\end{document}